\documentclass{amsart}

\usepackage{amsmath}
\usepackage{amssymb}

\usepackage{graphicx}
\usepackage{amsfonts,amsthm,amsbsy}
\usepackage{latexsym,mathrsfs}
\usepackage{color}
\usepackage{enumerate}
\usepackage{algorithm}
\usepackage{algorithmic}
\usepackage{stmaryrd}
\usepackage{skak}
\usepackage{multirow}
\usepackage{mathtools}
\usepackage{adjustbox}
\usepackage{float,lscape}

\DeclareMathOperator{\T}{T}
\DeclareMathOperator{\sT}{\scriptscriptstyle T}

\DeclareMathOperator{\alt}{alt}
\DeclareMathOperator{\dist}{dist}
\DeclareMathOperator{\grad}{grad}

\DeclareMathOperator{\sym}{sym}
\DeclareMathOperator{\tr}{tr}
\DeclareMathOperator{\UI}{ui}
\DeclareMathOperator{\diag}{Diag}

\DeclareMathOperator{\eig}{eig}

\DeclareMathOperator{\opt}{opt}
\DeclareMathOperator{\rank}{rank}
\DeclareMathOperator{\scf}{scf}

\makeatletter
\DeclareRobustCommand*\cal{\@fontswitch\relax\mathcal}
\makeatother

\def\cG{{\cal G}}
\def\wtd{\widetilde}
\def\what{\widehat}
\def\T{\mbox{\scriptsize \rm T}}

\def\bbO{\mathbb{O}}
\def\bbR{\mathbb{R}}

\def\bone{\pmb{1}}
\def\scrG{\mathscr{G}}

\def\cR{{\cal R}}

\def\cY{{\cal Y}}

\newtheorem{theorem}{Theorem}
\newtheorem{lemma}{Lemma}
\newtheorem{corollary}{Corollary}

\theoremstyle{definition}

\newtheorem{remark}{Remark}
\newtheorem{example}{Example}

\title[A SCF Iteration for Orthogonal Canonical Correlation Analysis]{A Self-consistent-field Iteration for \\Orthogonal Canonical Correlation Analysis}
\author[L. Zhang]{Leihong Zhang}
\author[L. Wang] {Li Wang}
\author[Z. Bai]{Zhaojun Bai}
\author[R. Li]{Ren-cang Li}
\thanks{Leihong Zhang is with School of Mathematical Sciences, Soochow University, Suzhou 215006, Jiangsu, China. Email: longzlh@163.com.}  
\thanks{Li Wang is with Department of Mathematics and Department of Computer Science and Engineering, University of Texas at Arlington, Arlington, TX 76019-0408, USA. Email: li.wang@uta.edu. }
\thanks{ Zhaojun Bai is with Department of Computer Science and Department of Mathematics, University of California, Davis, CA 95616, USA. Email: zbai@ucdavis.edu.}
\thanks{Ren-cang Li is with Department of Mathematics, University of Texas at Arlington, Arlington, TX 76019-0408, USA. Email: rcli@uta.edu.  }

\begin{document}
\maketitle

\begin{abstract}
We propose an efficient algorithm for solving orthogonal canonical correlation analysis (OCCA) in the form of
trace-fractional structure and orthogonal linear projections. Even though orthogonality has been widely used and
proved to be a useful criterion for pattern recognition and feature extraction, existing methods for solving OCCA
problem are either numerical unstable by relying on a deflation scheme, or less efficient by directly using
generic optimization methods. In this paper, we propose an alternating numerical scheme whose core is
the sub-maximization problem in the trace-fractional form with an orthogonal constraint.
A customized self-consistent-field (SCF) iteration for this sub-maximization problem is devised.
It is proved that the SCF iteration is globally convergent to a KKT point and that the alternating numerical scheme
always converges. We further formulate a new
trace-fractional maximization problem for orthogonal multiset CCA (OMCCA) and then propose an efficient algorithm with an either Jacobi-style or Gauss-Seidel-style updating scheme based on the same SCF iteration. Extensive
experiments are conducted to evaluate the proposed algorithms against existing methods including two
real world applications: multi-label classification and multi-view feature extraction. Experimental results
show that our methods not only perform competitively to or better than baselines but also are more efficient.
\end{abstract}

\section{Introduction}
Canoical Correlation analysis (CCA) \cite{cugh:2015,hote:1936} is a standard statistical technique and widely-used feature extraction paradigm for two sets of multidimensional variables. It finds basis vectors for the two sets of variables such that the correlations between the projections of the variables onto these basis vectors are mutually maximized.

During the last decade,  CCA  has received a renewed interest in the machine learning community.
Many CCA variants have been proposed. Kernel CCA \cite{akaho2006kernel,hardoon2004canonical}
and neural networks \cite{andrew2013deep} were introduced for exploring the nonlinear correlations.
To deal with the underdetermined settings, regularization was introduced into CCA.
Sparse CCA \cite{witten2009penalized} was further proposed to facilitate the interpretation of the results through
sparsity-inducing norms. Bayesian CCA \cite{klami2013bayesian} provided an alternative for CCA to handle samples of small sizes
and with noises. Semi-paired CCA \cite{zhang2014semi} is able to handle the case where some of the observations in either view
are missing. Multiset CCA \cite{kettenring1971canonical, nielsen2002multiset} extended CCA to find projections
for more than two views.

The applicability of CCA has been demonstrated in various fields \cite{uurtio2018tutorial}.
In machine learning community, CCA has been used for unsupervised data analysis when multiple views
are available \cite{dhillon2011multi,hardoon2007unsupervised}, data fusion of multiple modalities \cite{sargin2007audiovisual},
and reducing sample complexity of prediction problems using unlabeled data \cite{kakade2007multi}.
CCA becomes one popular tool for supervised learning, in which one view is derived from the data
and the other view is derived from the class labels, such as multi-label classification
\cite{rai2009multi,sun2010canonical,zhang2011multi}. More details about variants of CCA
and their applications can be found in the recent survey \cite{uurtio2018tutorial}.

In this paper, we are mainly interested in another variant of CCA, in which projections are linear
and constrained to be orthogonal. Orthogonality is a widely used and effective criterion for pattern recognition
and feature extraction. Orthogonality has been successfully adopted by various learning models.
Linear discriminant analysis (LDA) with an orthogonality constraint often leads to better performance
than conventional LDA since the orthogonality constraint to some extent can remove noise \cite{ye2005characterization}.
Orthogonal neighborhood preserving projections \cite{kokiopoulou2007orthogonal} is proposed to achieve good representation
of the global structure and be effective for data visualization. Orthogonal locality preserving indexing \cite{cai2005orthogonal}
shares the same locality preserving character as locality preserving indexing, but at the same time it requires
the basis functions to be orthogonal so that the metric structure of the document space is preserved. Orthogonality
has also been explored in CCA \cite{cugh:2015,shen2013orthogonal}. In addition to the above-mentioned properties brought
by the orthogonality such as less sensitive to noise, good for data visualization, and preserving metrics, OCCA specifically
preserves the covariance of the original data, while CCA does not \cite{cugh:2015}. OCCA is also extended for more than two
views \cite{shen2015orthogonal}.

Although OCCA brings additional advantages for data analysis, it no longer retains an analytic solution as CCA
in terms of singular value decomposition (SVD).
A common heuristic approach is to orthogonalize the basis vectors obtained by CCA. However, this produces a suboptimal solution for the OCCA problem. In \cite{shen2013orthogonal}, an incremental scheme is employed to produce current basis vectors with additional constraints to enforce the orthogonality with the previously computed basis vectors. We will point out in Section \ref{sec:related-work} that the incremental scheme relies on a numerically unstable generalized eigenvalue problem, so the set of basis vectors are not guaranteed to be mutually orthogonal. Often, the orthogonalization as a post-processing step is required to obtain a feasible solution. In \cite{cugh:2015}, generic optimization methods for minimizing a smooth function over the product of the Stiefel manifolds are used to solve the OCCA problem. These methods usually converge to a local minimizer but they do not take the special trace-fractional structure of OCCA into consideration. As a result, they are usually less efficient than  custom-made algorithms.

One of our goals of this paper is to design efficient algorithms for solving the OCCA problem with guaranteed theoretical convergence and numerical stabilizability. In order to fully explore the special trace-fractional structure of OCCA, we first uncover the connection of OCCA  with an eigenvector-dependent nonlinear eigenvalue problem (NEPv), and then naturally come up with a simple iterative method whose numerical efficiency is guaranteed by a succinct but structure-exploiting self-consistent-field (SCF) iteration. Global convergence and local convergence of this customized algorithm are established.

\vskip 1mm

\noindent\textbf{Contributions.}
The main contributions of this paper are summarized as follows:
\begin{itemize}
	
	\item We propose a novel algorithm OCCA-scf for solving OCCA in the form of a trace-fractional matrix optimization problem. The proposed algorithm is built upon an efficient and effective SCF iteration to solve a very special trace-ratio sub-maximization problem through taking the trace-fractional structure into account. It is proved that the SCF iteration is always convergent and, as a result, OCCA-scf is guaranteed to converge. It can also integrate the state-of-the-art eigensolvers within the iteration framework. Moreover, it guarantees the orthogonality of the computed basis vectors.
	%
	
	\item We present a new OMCCA model with integrated weights for each pair of views and trace-fractional objective for correlations between any two views. By leveraging the same customized SCF iteration, a novel range constrained OMCCA algorithm is proposed with an either Jacobi-style or Gauss-Seidel-style updating scheme.
	
	\item Extensive experiments are conducted for evaluating the proposed algorithms against existing methods in terms of various measurements, including sensitivity analysis, correlation analysis, computation analysis, and data visualization. We further apply our methods for two real world applications: multi-label classification and multi-view feature extraction. Experimental results show that our methods not only perform competitively to or better than baselines but also are more efficient.
\end{itemize}


\noindent\textbf{Paper organization.}
We first review several related CCA models and their orthogonal variants in Section \ref{sec:related-work}. In Section \ref{sec:occa-2v}, we propose a novel algorithm for solving OCCA problem. A customized SCF iteration is presented to solve the sub-maximization problem of the  algorithm in Section \ref{sec:SCFAlt}. In Section \ref{sec:MOCCA}, we develop a new algorithm for OMCCA by leveraging the same SCF iteration. Extensive experiments are conducted in
Section \ref{sec:experiments}. Finally, we conclude this work in Section \ref{sec:conclusion}. 

\vskip 1mm

\noindent\textbf{Notation.}
$\bbR^{m\times n}$ is the set of $m\times n$ real matrices and $\bbR^n=\bbR^{n\times 1}$. $I_n\in\bbR^{n\times n}$ is
the identity matrix, and $\bone_n\in\bbR^n$ is the vector of all ones. $\|\mathbf{x}\|_2$ is the 2-norm of vector $\mathbf{x}\in \bbR^n$. For $B\in\bbR^{m\times n}$,
$\cR(B)$ is the column subspace and its singular values are denoted by
$\sigma_i(B)$ for $i=1,\ldots,\min(m,n)$ arranged in the nonincreasing order.
The thin SVD of $B$ is the one $B=U\Sigma V^{\T}$ such that $\Sigma\succ 0$ and $\|B\|_{\tr}=\sum_{i=1}^{{\rm rank}(B)}\sigma_i(B)$ is its trace norm
(also known as the nuclear norm).
When $m=n$, $\sym(B)=(B+B^{\T})/2$;
if $B$ is also symmetric, then $\eig(B)=\{\lambda_i(B)\}_{i=1}^n$ denotes the set of its eigenvalues (counted by multiplicities)
arranged in the nonincreasing order, and
$B\succ 0 (\succeq 0)$ means that $B$ is also positive definite (semi-definite). 
The Stiefel manifold \cite{abms:2008}
$$\bbO^{n\times k}=\{X\in\bbR^{n\times k}\,:\, X^{\T}X=I_k\},$$
is an embedded submanifold of $\bbR^{n\times k}$ endowed with the standard inner product $\langle X,Y \rangle=\tr(X^{\T}Y)$ for $X,Y\in \bbR^{n\times k}$,
%
where $\tr(X^{\T}Y)$ is the trace of $X^{\T}Y$. Moreover, the tangent space ${\cal{T}}_X \bbO^{n\times k}$ of $\bbO^{n\times k}$ at $X\in \bbO^{n\times k}$ is given by (see e.g., \cite{abms:2008})

\begin{small}\vspace{-0.1in}
	\begin{align}\label{eq:tangentH}
	{\cal{T}}_X \bbO^{n\times k}&
	= \left\{\! H \in \bbR^{n\times k}\!\left|\!\!\!\begin{array}{l}
	H=XK+(I_n-XX^{\T})J \\
	\forall K=-K^{\T}\in \bbR^{k\times k},~J\in\bbR^{n\times k}
	\end{array}
	\right.
	\!\!\!\!\right\}.
	\end{align} \vspace{-0.1in}
\end{small}

\section{Related Work} \label{sec:related-work}
We review some representatives of CCA methods and their extension to multiple sets of variables.

\vspace{-0.1in}
\subsection{Canonical Correlation Analysis}\label{ssec:2v-cca}
Canonical Correlation Analysis (CCA) is a two-view multivariate statistical method \cite{hote:1936},
where the variables of  observations can be partitioned into two sets, i.e., the two views of the data.
Denote the data matrices $S_1 \in \bbR^{n \times q}$ and $S_2 \in \bbR^{m \times q}$ from view $1$
and  view $2$ with $n$ and $m$ features, respectively, where $q$ is the number of samples. Assume both $S_1$ and $S_2$ are centralized,
i.e., $ S_1 \mathbf{1}_q= 0$ and $ S_2 \mathbf{1}_q=0$; otherwise, we may preprocess $S_i$ as
$S_i\leftarrow S_i-\frac 1q(S_i\bone_q)\bone_q^{\T}$, $i=1,2$.  Let $\mathbf{x}_1 \in \bbR^{n}$
and $\mathbf{x}_2 \in \bbR^{m}$ be the canonical weight vectors. The canonical
variates are the linear transformations defined as $\mathbf{z}_1 = S_1^{\T} \mathbf{x}_1,~~  \mathbf{z}_2 = S_2^{\T} \mathbf{x}_2.$
The canonical correlation between the two canonical variates is defined as
$\rho(\mathbf{x}_1, \mathbf{x}_2) = \frac{\mathbf{z}_1^{\sT} \mathbf{z}_2}{||\mathbf{z}_1||_2 ||\mathbf{z}_2||_2}.$
CCA aims to find the pair of canonical weight vectors that maximize the canonical correlation:

\begin{small}\vspace{-0.1in}
	\begin{align}
	\max_{\mathbf{x}_1, \mathbf{x}_2} \rho(\mathbf{x}_1, \mathbf{x}_2). \label{op:cca-1d}
	\end{align}
\end{small}\noindent
It can also be interpreted as the problem of finding the best pair of canonical weight vectors
so that the cosine of the angle between two canonical variates is maximized, that is, the smallest angle in $[0, \frac{\pi}{2}]$.

This single-vector CCA (\ref{op:cca-1d}) has been extended to obtain the pair of canonical weight matrices,
namely, the pair of canonical weight matrices $X_1\in \bbR^{n \times k}$ and $X_2\in \bbR^{m \times k}$ by solving
the following optimization problem

\begin{small}\vspace{-0.1in}
	\begin{equation}\label{eq:CCA-multi}
	\max_{X_1, X_2}\, \tr(X_1^{\T} C_{1,2} X_2),\,\,
	\textrm{s.t.}\,\,  X_i^{\T} C_{i,i} X_i = I_k,\,i=1,2,
	\end{equation}
\end{small}\noindent
where $C_{i,j}=S_i S_j^{\T}$, $i, j=1,2$.
In general, the closed-form solution of (\ref{eq:CCA-multi}) can be obtained by SVD, and indeed,
it can be proved that there is an optimal solution  $(X_1,X_2)$  so that $X_1^{\T} C_{1,2} X_2\succeq 0$  \cite[Theorem III.2]{chlg:2013a}.

Traditional CCA is not suitable for settings
where an orthogonal projection is required such as for data visualization in an orthogonal coordinate system.
This is because optimal $X_1$ and $X_2$ in (\ref{eq:CCA-multi}) usually do not have orthonormal columns.
One can orthogonalize the columns of the optimal $X_1$ and $X_2$ to the problem \eqref{eq:CCA-multi}
as a post-processing step, but it is generally suboptimal. For that reason, orthogonal CCA (OCCA) \cite{cugh:2015}
is proposed to maximize the correlation

\begin{small}\vspace{-0.1in}
	$$
	f(X,Y)=\frac{\tr(X_1^{\T} C_{1,2} X_2)}{\sqrt{\tr(X_1^{\T} C_{1,1} X_1) \tr(X_2^{\T} C_{2,2} X_2)}}
	$$
\end{small}\noindent
directly over orthonormal matrices, i.e.,
\begin{small}\vspace{-0.1in}
	\begin{equation}\label{eq:OCCA}
	\max_{X_1\in \mathbb{O}^{n \times k}, X_2 \in \mathbb{O}^{m \times k}} \,
	f(X,Y). 
	\end{equation}
\end{small}\noindent
As pointed in \cite{cugh:2015}, OCCA is different from traditional CCA because OCCA preserves the covariance of the original
data $S_1$ and $S_2$ by finding orthonormal matrices where correlation is maximized, while traditional CCA whitens
each dataset and orthogonally projects them so that the correlation is maximized. Generic optimization methods for minimizing/maximizing a
smooth function over the product of the Stiefel manifolds are available; for example, classical optimization algorithms such as
the steepest descent gradient or the trust-region methods over the Euclidean space have been extended to the general
Riemannian manifolds in, for example, \cite{abms:2008}; however, besides only guaranteeing to converge to a local optimizer at best,
these generic algorithms do not make use of the special trace-fractional structure in \eqref{eq:OCCA},
and therefore, they usually are less efficient than custom-made algorithms for trace-ratio-related optimizations
(see \cite{zhli:2014a,zhli:2014b} for some numerical results of trace-ratio optimizations).

The motivation of imposing orthogonality constraints  was also explored in \cite{shen2013orthogonal}.
A greedy method (which we will call OCCA-SSY for short) is employed to find $k$ pairs of orthogonal vectors,
computed one pair at a time.
The initial step is the same as traditional CCA to find the pair of canonical weight vectors
$(\mathbf{x}_1^{(1)}, \mathbf{x}_2^{(1)})$ that solves (\ref{op:cca-1d}).
Given $\{ (\mathbf{x}_1^{(t)}, \mathbf{x}_2^{(t)}) \}_{t=1}^r$, the $(r+1)$st step is to solve the following problem

\begin{small}\vspace{-0.1in}
	\begin{align*}
	&(\mathbf{x}_1^{(r+1)}, \mathbf{x}_2^{(r+1)})=\arg \max_{\mathbf{x}_1, \mathbf{x}_2} ~ \rho(\mathbf{x}_1, \mathbf{x}_2) \\
	&\textrm{s.t.}\quad \mathbf{x}_j^{\T} C_{j,j} \mathbf{x}_j = 1, \,\,
	\mathbf{x}_j^{\T} \mathbf{x}_j^{(t)} = 0,\, j=1,2,\, t=1,\ldots,r.
	\end{align*}
\end{small}\noindent
Such an  approach relies on a deflation scheme, and the pair $(\mathbf{x}_1^{(r+1)}, \mathbf{x}_2^{(r+1)})$ is claimed to
correspond to the dominant eigenpair of a generalized eigenvalue problem, which, however, numerically may not have any real eigenpair
and thus is numerically unusable\footnote{Private communications with the authors of \cite{shen2013orthogonal}, 2019.}.

\subsection{Multiset Canonical Correlation Analysis}\label{ssec:MCCA}

Multiset CCA (MCCA) \cite{kettenring1971canonical,nielsen2002multiset} is proposed to analyze linear relationships
among more than two canonical variates. It is a generalization of traditional CCA \cite{hote:1936}.
Here, we briefly introduce one widely used model \cite{nielsen2002multiset}
by seeking projections to maximize the sum of the pairwise correlation between any two canonical variates.
Specifically, given $\ell$ datasets in the form of matrices

\begin{small}
	\begin{equation}\label{eq:data-Xi}
	S_i\in\bbR^{n_i\times q}\quad\mbox{for $i=1,2,\ldots,\ell$},
	\end{equation}
\end{small}\noindent
where $n_i$ is the number of features in the $i$th dataset, and $q$ is the number of data points in
each of the datasets. Without loss of generality, we may assume
that all $S_i$  are centered,
i.e., $S_i\bone_q=0$ for all $i$.


Let $C_{i,j}=S_iS_j^{\T}$ for $i,\,j=1,\ldots,\ell$. MCCA seeks to find the set of $\ell$
canonical weight vectors that solve

\begin{small}\vspace{-0.1in}
	\begin{align}\nonumber
	\max_{\mathbf{x}_1,\ldots, \mathbf{x}_\ell} &~ \sum_{i,\,j=1}^\ell  \mathbf{x}_i^{\T} C_{i,j} \mathbf{x}_j
	\end{align}
\end{small}\noindent
subject to 

\begin{small}\vspace{-0.1in}
	\begin{align}
	\mbox{either}\quad&	\sum_{i=1}^\ell \mathbf{x}_i^{\T} C_{i,i} \mathbf{x}_i = 1,  \label{eq:mset-CCA:contr-a} \\
	\mbox{or}\quad&	\mathbf{x}_i^{\T} C_{i,i} \mathbf{x}_i = 1, \, i=1,\ldots, \ell. \label{eq:mset-CCA:contr-b}
	\end{align}
\end{small}\noindent
KKT conditions for MCCA under either \eqref{eq:mset-CCA:contr-a} or \eqref{eq:mset-CCA:contr-b} can be found
in \cite{nielsen2002multiset}. In particular, under \eqref{eq:mset-CCA:contr-a}, the condition is a
generalized eigenvalue problem \cite[p.297]{nielsen2002multiset}, which can be solved by any mature eigensolver
\cite{bddrv:2000,govl:2013}.

%
In \cite{shen2015orthogonal}, a greedy OMCCA (which we will call OMCCA-SS for short) was proposed, similarly to
\cite{shen2013orthogonal}.
Recursively, it goes as follows.
Given $\{ \mathbf{x}_i^{(t)}, \forall i=1,\ldots,\ell, t=1,\ldots,r \}$,
OMCCA-SS solves the following problem

\begin{small}\vspace{-0.1in}
	\begin{align*}
	&\{\mathbf{x}^{(r+1)}_{1},\ldots, \mathbf{x}^{(r+1)}_{\ell}\}	
	= \arg  \max_{\mathbf{x}_1,\ldots, \mathbf{x}_\ell} ~ \sum_{i,\,j=1}^\ell  \mathbf{x}_i^{\T} C_{i,j} \mathbf{x}_j, \\
	&	\textrm{s.t.} \quad \mbox{\eqref{eq:mset-CCA:contr-a} and}\,\,
	\mathbf{x}_i^{\T} \mathbf{x}^{(t)}_{i} = 0, \, i=1,\ldots \ell, t=1,\ldots, r.
	\end{align*}
\end{small}\noindent
This OMCCA-SS method inherits the same issues as OCCA-SSY discussed in Subsection \ref{ssec:2v-cca}.

\section{Novel Algorithm for Orthogonal CCA} \label{sec:occa-2v}
In this section we propose a new optimization scheme for solving (\ref{eq:OCCA}), taking advantage of its special form.

\vspace{-0.1in}
\subsection{Problem Formulation}

Following the settings of all CCA methods, both views of the data $S_1$ and $S_2$ are centralized in advance. Define

\begin{small}\vspace{-0.1in}
	\begin{align*}
	A = S_1 S_1^{\T} \in\bbR^{n\times n},~ B = S_2 S_2^{\T} \in\bbR^{m\times m},~ C = S_1 S_2^{\T} \in\bbR^{n\times m},
	\end{align*}
\end{small}\noindent
and, let $X \in\bbO^{n\times k}$ and $Y \in\bbO^{m\times k}$ have orthonormal columns.
Given  an integer $1\le k<\min\{m,n\}$ (usually $k\ll \min\{m,n\}$),
we consider, instead of tackling problem (\ref{eq:OCCA}) directly,

\begin{small}\vspace{-0.1in}
	\begin{equation}\label{eq:CCAvar}
	\max_{X \in \mathbb{O}^{n \times k}, Y \in \mathbb{O}^{m \times k}}\left\{F(X,Y):=\frac {\tr^2(X^{\T}CY)}{\tr(X^{\T}AX)\tr(Y^{\T}BY)}\right\}.
	\end{equation}
\end{small}\noindent
%
Evidently, (\ref{eq:OCCA}) and (\ref{eq:CCAvar}) are equivalent. In the next subsection we will develop a new algorithm based on
(\ref{eq:CCAvar}). Our algorithm  can take the advantage of the specific structure of the problem
with  theoretical guarantees, and more importantly, can be extended easily
to handle the OMCCA model in Section \ref{sec:MOCCA}.

\vspace{-0.1in}
\subsection{The Proposed Algorithm}\label{ssec:AltOCCA}
We propose the numerical scheme as shown in Algorithm~\ref{alg:AltOCCA}. 
Although the framework of the proposed
numerical scheme is rather natural, i.e.,
maximizing $F(X,Y)$ (or equivalently $f(X,Y)$) alternatively with respective to $X$ and $Y$, the real novelty lies in the way how
its core sub-maximization problems \eqref{eq:AltXopt} and \eqref{eq:AltYopt} are solved, which
is the subject of Section~\ref{sec:SCFAlt} later.



\begin{algorithm}[h]
	\caption{An alternative optimization scheme for \eqref{eq:CCAvar}} \label{alg:AltOCCA}
	\begin{small}
		\begin{algorithmic}[1]
			\REQUIRE $\{X^{(0)},Y^{(0)}\}$ with $X^{(0)}\in \bbO^{n\times k},Y^{(0)} \in \bbO^{m\times k}$.
			\ENSURE  a solution $\{X^{(\nu)},Y^{(\nu)}\}$ to \eqref{eq:CCAvar}.
			\FOR{$\nu=1,2,\ldots$ until convergence}
			\STATE Solve
			\begin{equation}\label{eq:AltXopt}
			X^{(\nu)}\in\arg\max_{X\in \bbO^{n\times k}} F(X,Y^{(\nu-1)}).
			\end{equation}
			
			\STATE Solve
			\begin{equation}\label{eq:AltYopt}
			Y^{(\nu)}\in\arg\max_{Y\in \bbO^{m\times k}} F(X^{(\nu)},Y).
			\end{equation}
			
			\STATE Compute SVD of $(X^{(\nu)})^{\T}C Y^{(\nu)}=\wtd U\wtd\Sigma\wtd V^{\T}$ and set $X^{(\nu)}\leftarrow X^{(\nu)}\wtd U$ and $Y^{(\nu)}\leftarrow Y^{(\nu)}\wtd V$;
			\ENDFOR
			\RETURN $\{X^{(\nu)},Y^{(\nu)}\}$ as a numerical solution to \eqref{eq:CCAvar}.
		\end{algorithmic}
	\end{small}
\end{algorithm}

The role of line 4 in Algorithm~\ref{alg:AltOCCA} is to make sure
$X^{(\nu)}$ and $Y^{(\nu)}$ are always well aligned. It is particularly important when
$X^{(\nu)}$ and $Y^{(\nu)}$ are approximations that are not best possibly accurate in the working precision.
Its motivation is based on the structure of
the function $F(X,Y)$: Given a pair $(X^{(\nu)},Y^{(\nu)})$, the denominator is unchanged when this pair
is changed to $(X^{(\nu)} U,Y^{(\nu)} V)$ for any $U,V\in \bbO^{k\times k}$, while the numerator is
maximized by the particular pair $(U,V)=(\wtd U,\wtd V)$ given by (see e.g., \cite{demm:1997,govl:2013})

\begin{small}
	\begin{equation}\nonumber
	(\wtd U,\wtd V)=\arg\max_{U,V\in \bbO^{k\times k}}\tr\left(U^{\T}  (X^{(\nu)})^{\T}C Y^{(\nu)} V\right),
	\end{equation}
\end{small}\noindent
{whose solution, according to the following technical Lemma \ref{lm:maxtrace}, can be achieved by, e.g.,
	the SVD at line 4 of  Algorithm  \ref{alg:AltOCCA}, and the maximum is $\sum_{i=1}^k\sigma_i\left((X^{(\nu)})^{\T}C Y^{(\nu)}\right)$.
	\begin{lemma}\label{lm:maxtrace}
		Let $W\in\bbR^{k\times k}$. Then
		$
		|\tr(W)|\le\sum_{i=1}^k\sigma_i(W).
		$
		If, however,
		$
		|\tr(W)|=\sum_{i=1}^k\sigma_i(W),
		$
		then $W$ is symmetric and is either positive or negative semidefinite.
\end{lemma}}

We also notice by Lemma \ref{lm:maxtrace}, that such additional refinement step brings another nice property for the
sequence $\{X^{(\nu)},Y^{(\nu)}\}$,
that is, $(X^{(\nu)})^{\T} CY^{(\nu)}$ is symmetric and positive semidefinite, which is a necessary condition for
any global solution $\{X_{\opt},Y_{\opt}\}$ (see Theorem \ref{prop:symXCY}(i)).
Moreover, by using the effective solvers for \eqref{eq:AltXopt} and \eqref{eq:AltYopt}, Algorithm \ref{alg:AltOCCA}
always converges. We summarize these results in the following theorem:

\begin{theorem}\label{prop:symXCY}
	Let $\{X_{\opt},Y_{\opt}\}$ be the optimal solution to \eqref{eq:CCAvar} and $\{X^{(\nu)},Y^{(\nu)}\}$ be the $\nu$th approximation
	of Algorithm \ref{alg:AltOCCA}. 
	Then
	\begin{itemize}
		\item[(i)] 	$X_{\opt}^{\T}CY_{\opt}$ is symmetric  and positive/negative semidefinite.
		\item[(ii)]
		$(X^{(\nu)})^{\T}CY^{(\nu)}$ is symmetric and positive semidefinite for $\nu\ge 1$, and thus for any limit pair $\{X,Y\}$ of $\{X^{(\nu)},Y^{(\nu)}\}_{\nu=1}^\infty$, $X^{\T}CY$ is symmetric and positive semidefinite.
		\item[(iii)] The sequence $\{F(X^{(\nu)},Y^{(\nu)})\}_{\nu=1}^{\infty}$  is monotonically increasing and converges.
	\end{itemize}
\end{theorem}
%

The efficiency of  Algorithm \ref{alg:AltOCCA} relies heavily on solving
the sub-maximization problems \eqref{eq:AltXopt} and \eqref{eq:AltYopt}.
Abstractly, they are of the following type
\begin{small}\vspace{-0.1in}
	\begin{equation}\label{eq:subOCCA}
	\max_{G\in\bbO^{n\times k}}\,\eta(G)\quad\mbox{with}\,\, \eta(G):=\frac {\tr^2(G^{\T}D)}{\tr(G^{\T}AG)},
	\end{equation}
\end{small}\noindent
where $0\neq D\in \mathbb{R}^{n\times k}$ and $A\succ 0$. In Section \ref{sec:SCFAlt}, we present an SCF iteration that directly aims at solving \eqref{eq:subOCCA}.

\vspace{-0.1in}
\section{An SCF iteration for \eqref{eq:subOCCA}}\label{sec:SCFAlt}

It can be seen that the global maximum of \eqref{eq:subOCCA} is positive unless $D=0$.
Moreover, \eqref{eq:subOCCA} is very much like the trace ratio  (or trace quotient)
maximization, i.e., maximizing $\tr(G^{\T}A_1G)/\tr(G^{\T}A_2G)$ over $G\in\bbO^{n\times k}$
with given $A_1,A_2\succ 0$, for which an efficient SCF iteration is available
\cite{zhln:2010,lizh:2019,zhan:2011}. It has been proved that the SCF iteration is globally convergent and
the convergence is  locally quadratic.
We would like to mention that the SCF iteration
was commonly used to solve the Eigenvector-Dependent Nonlinear
Eigenvalue Problem (NEPv) \cite{cazb:2018}
from the {Kohn--Sham} density functional theory in
electronic structure calculations \cite{mart:2004,sacs:2010}. Lately,
it has been attracting a great deal attention in  data science
(e.g., \cite{balv:2018,cazb:2018,wara:2015,zhli:2014a,zhli:2014b}).

\vspace{-0.1in}
\subsection{A nonlinear eigenvalue problem}\label{ssec:NEPv}

We will first derive the formula for the partial derivative
$
{\partial \eta(G)}/{\partial G},
$
where all entries $G$  are treated as independent variables,
and then the formula for the gradient
$\grad\eta(G)$ at $G\in \bbO^{n\times k}$
on the Stiefel manifold $\bbO^{n\times k}$ is given by (see e.g., \cite{abms:2008})

\begin{small}
	\begin{equation}\label{eq:grad-formula}
	\grad\eta(G)=\Pi_G\left(\frac {\partial \eta(G)}{\partial G}\right)\in {\cal{T}}_G \bbO^{n\times k},
	\end{equation}
\end{small}\noindent
where
$
\Pi_G(Z)=Z-G\,\sym(G^{\T}Z)
$
for $Z\in\bbR^{n\times k}$. By calculations, we have

\begin{small}\vspace{-0.1in}
	\begin{align*}
	\frac {\partial \eta(G)}{\partial G}=\frac {2\phi_D}{\phi_A}\,D-\frac {2\phi_D^2}{\phi_A^2}\,AG,	
	\end{align*}
\end{small}\noindent
where, for convenience, we let $\phi_D\equiv \phi_D(G)=\tr(G^{\T}D)$ and
$\phi_A\equiv \phi_A(G)=\tr(G^{\T}AG)$. Finally, use \eqref{eq:grad-formula} to get

\begin{small}\vspace{-0.1in}
	\begin{equation}\nonumber
	-\frac{\xi^2(G)}{2}\grad\eta(G)=[AG-\xi(G)D]-GM(G)\in \bbR^{n\times k},
	\end{equation}
\end{small}\noindent
where

\begin{small}\vspace{-0.1in}
	\begin{equation}\label{eq:M}
	\xi(G)=\frac{\phi_A}{\phi_D},~~M(G)=\sym(G^{\T}AG-\xi(G)G^{\T}D)\in \bbR^{k\times k}.
	\end{equation}
\end{small}\noindent
The discussion so far partially proved the KKT condition for \eqref{eq:subOCCA} in
Lemma~\ref{lem:KKTsubOCCA}.

\begin{lemma}\label{lem:KKTsubOCCA}
	If $G$ is a maximizer of \eqref{eq:subOCCA}, then $G^{\T}D=D^{\T}G$ and
	
	\begin{small}\vspace{-0.1in}
		\begin{equation}\label{eq:KKTsubOCCA}
		AG-\xi(G)D=GM(G).
		\end{equation}
	\end{small}
\end{lemma}

Note that the condition \eqref{eq:KKTsubOCCA} is a type of nonlinear Sylvester equation
but with the orthogonality constraint $G^{\T}G=I_k$. To solve it, we will convert
it into an NEPv so that certain type of SCF is applicable. One straightforward way is to use the constraint $G^{\T}G=I_k$
and then rewrite \eqref{eq:KKTsubOCCA} equivalently as $[A-\xi(G)DG^{\T}]G=GM(G)$.
However, we notice that the matrix $A-\xi(G)DG^{\T}$ is not necessarily symmetric, even at a maximizer $G$.
This means that we cannot ensure $A-\xi(G)DG^{\T}$ has real eigenvalues at  $G\in \bbO^{n\times k}$.
To overcome that obstacle, we construct the following NEPv instead:

\begin{small}\vspace{-0.1in}
	\begin{equation}\label{eq:NepsubOCCA}
	E(G)G=G\what M(G),
	\end{equation}
\end{small}\noindent
where $\what M(G)=G^{\T}E(G)G\in\bbR^{k\times k}$ and

\begin{small}\vspace{-0.1in}
	$$ 
	E(G):=A-\xi(G)(DG^{\T}+GD^{\T}).
	$$ 
\end{small}\noindent
Evidently, $E(G)$ is always symmetric.
The following lemma establishes a relation between \eqref{eq:NepsubOCCA} and \eqref{eq:KKTsubOCCA}.

\begin{lemma}\label{lem:NepsubOCCA}
	Suppose $G\in\bbO^{n\times k}$. Then $G$ satisfies \eqref{eq:KKTsubOCCA} if and only if
	$G$ is an eigenbasis matrix of $E(G)$, i.e., $G$ satisfies \eqref{eq:NepsubOCCA}.
\end{lemma}

This lemma
characterizes any maximizer $G$ of \eqref{eq:subOCCA} as an orthonormal eigenbasis matrix of $E(G)$.
By  \eqref{eq:NepsubOCCA}, we find

\begin{small}\vspace{-0.1in}
	\begin{equation}\nonumber
	\eig(\what M(G))=\{\lambda_{\pi_1}(E(G)),\dots,\lambda_{\pi_k}(E(G))\}\subseteq \eig(E(G)),
	\end{equation}
\end{small}\noindent
where $\{\pi_1\le\dots\le\pi_k\}\subset\{1,2,\ldots,n\}$.
A followup question is where the eigenvalues $\lambda_{\pi_i}(E(G))$ are located within $\eig(E(G))$.
In the next two subsections, we will investigate this issue and
the investigation yields important necessary conditions for local and global maximizers
of \eqref{eq:subOCCA}.

\vspace{-0.1in}
\subsection{Eigenspace associated with a local maximizer}\label{subsec:eigenspace-local}
Even though our maximization problem \eqref{eq:subOCCA} is very much like the trace ratio problem
\cite{zhln:2010}, unfortunately, it does not enjoy some nice properties as the trace ratio problem
(for example, it is shown that any local maximizer of the trace ratio problem is also a global solution).
In  Example \ref{eg:Eg1} below, we will see numerically that  \eqref{eq:subOCCA}
may admit local but non-global maximizers.

\begin{example}\label{eg:Eg1}
	Consider the case with $n=5,~k=2$,
	{\small $$A=\left[\begin{array}{ccccc}4 & 0 & -5 & -5&1 \\ 0& 2 &1&-1&1  \\ -5& 1 &9&5&1\\-5&-1&5&18&4\\-1&1&1&4&2\end{array}\right] ~~\mbox{and}~~D=\left[\begin{array}{cc}-1 & 1 \\0 & 0 \\0 & 2\\0&0\\1&0\end{array}\right].$$}\noindent
	By calling the MATLAB function \texttt{fmincon}, we find two (numerical) local maximizers:
	
	\begin{small}\vspace{-0.1in}
		\begin{align}
		G_+&=\begin{bmatrix}
		-0.358041496119094 &  0.770164268103322\\
		-0.453284095949462 & -0.326431512218038\\
		-0.091335437376569 &   0.497561512998402\\
		-0.269574025133855 &  0.008593213179154\\
		0.765066989399257  & 0.229451880441015
		\end{bmatrix}, \nonumber \\
		G_*&=\begin{bmatrix}
		-0.506648923972689 &  0.664385053189626\\
		0.619602876311725 &  0.312889763321350\\
		-0.337893503149209 &  0.384494340924914\\
		0.103073503143856  & 0.210902556071053\\
		-0.484358314662567&  -0.518050876600301
		\end{bmatrix}. \nonumber
		\end{align}
	\end{small}\noindent
	It is computed that
	$\eta(G_+)\approx 2.303<\eta(G_*)\approx 10.16$. We argue that they are local maximizers.
	First the  norms of of the corresponding gradients at $G_+$ and $G_*$ are less than $10^{-6}$.
	Second, by sampling randomly $10^7$  tangent vectors $H$ {in the form of \eqref{eq:tangentH}},
	we found that for both $G_+$ and $G_*$, the following second order sufficient condition to be given in \eqref{eq:2ndOpt} hold,
	implying that both are local maximizers. This example numerically shows \eqref{eq:subOCCA}
	in general admits local but non-global maximizers.
\end{example}

The following lemma presents  second-order  necessary and sufficient conditions for
local maximizers \cite{nowr:2006,yazs:2014}.

\begin{lemma}\label{lem:2ndOpt}
	Let $G$ be a local maximizer of \eqref{eq:subOCCA}. Then for all nonzero $H\in \bbR^{n\times k}$ satisfying $H^{\T}G=-G^{\T}H$, it holds that
	
	\begin{small}\vspace{-0.1in} 	
		\begin{equation}\label{eq:2ndOpt}
		\tr^2(D^{\T}H)\le \eta(G)\left(\tr(H^{\T}AH)-\tr(HM(G)H^{\T})\right),
		\end{equation}
	\end{small}\noindent	
	where   $M(G)$ is  given in \eqref{eq:M}. Conversely, if the  inequality  \eqref{eq:2ndOpt}
	is strict, then $G$ is a  local maximizer.
\end{lemma}

\begin{corollary}\label{cor:2ndOpt}
	If $G$ is a local maximizer of \eqref{eq:subOCCA}, then  we have
	
	\begin{small}\vspace{-0.1in}
		\begin{equation}\label{eq:2ndOpta}
		\tr(K^{\T}(G^{\T}AG-M(G))K)\ge 0, \forall K=-K^{\T}\in \bbR^{k\times k},
		\end{equation}
	\end{small}\noindent
	and moreover, for all $J\in \bbR^{n\times k}$,
	
	\begin{small}\vspace{-0.1in}
		\begin{align}
		&\frac{\left(\tr(D^{\T}J)-\tr(G^{\T}DJ^{\T}G)\right)^2}{\eta(G)}
		\le \tr(J^{\T}E(G)J)\nonumber \\
		&\quad-\tr(J^{\T}GM(G)G^{\T}J)+\xi(G)\tr(J^{\T}GD^{\T}GG^{\T}J) \nonumber \\
		&\quad +\tr(G^{\T}JM(G)J^{\T}G)-\tr(JM(G)J^{\T}),
		\label{eq:2ndOptb}
		\end{align}
	\end{small}\noindent
	where   $\xi(G)$ is given in \eqref{eq:M}.
\end{corollary}
\begin{theorem}\label{thm:NepsubOCCALocal}
	Suppose $G$ is a local maximizer of \eqref{eq:subOCCA}. Then $\cR(G)$ is an eigenspace of $E(G)$  associated with
	eigenvalues $\lambda_{\pi_1}(E(G))\le \dots\le\lambda_{\pi_k}(E(G))$ satisfying $\pi_1\le k$.
\end{theorem}

Theorem \ref{thm:NepsubOCCALocal} indicates that for any local maximizer $G$, the smallest
eigenvalue associated with the eigenspace $\cR(G)$ must be smaller than $\lambda_k(E(G))$.
This offers a necessary condition for a KKT point to be a local maximizer.
As a much stronger version, we will show in the next subsection that any global maximizer $G$
must be an eigenbasis matrix  associated with the $k$ smallest eigenvalues of $E(G)$.

\vspace{-0.1in}
\subsection{Eigenspace associated with a global maximizer}\label{subsec:eigenspace-global}
The results in Lemma~\ref{lem:key} and Theorem~\ref{thm:NepsubOCCA} hold
keys to characterize the global maximizer and to establish an SCF iteration for \eqref{eq:subOCCA}.

\begin{lemma}\label{lem:key}
	For any $G,\widehat{G}\in \bbO^{n\times k}$, if $\widehat{G}^{\T}D=D^{\T}\widehat{G}\succeq 0$ and
	
	\begin{small}\vspace{-0.1in}
		\begin{equation}\label{eq:ineqeig}
		\tr(\widehat G^{\T}E(G)\widehat G)\le \tr(G^{\T}E(G)G),
		\end{equation}
	\end{small}\noindent
	then $\eta(\widehat {G})\ge \eta(G)$. Furthermore, if the inequality in \eqref{eq:ineqeig} is strict,
	then $\eta(\widehat {G})> \eta(G)$.
\end{lemma}
\begin{theorem}\label{thm:NepsubOCCA}
	Suppose $G_{\opt}$ is a global maximizer to \eqref{eq:subOCCA}.
	Then $G_{\opt}$ is an orthonormal eigenbasis matrix associated with the $k$ smallest eigenvalues of $E(G_{\opt})$.
	Moreover, the matrix $G_{\opt}^{\T}D$ is symmetric and  either positive or negative semidefinite.
\end{theorem}\marginpar{}
Returning to Example 1 with  the computed  $G_+$ and $G_*$, we observe that both $G_+$ and $G_*$
are (approximately) the orthonormal eigenbasis matrix of $E(G_+)$ and $E(G_*)$, respectively.

\vspace{-0.1in}
\subsection{A self-consistent-field (SCF) iteration}\label{subsec:SCF}
Equipped with the necessary condition in Theorem \ref{thm:NepsubOCCA},
we propose an SCF iteration as outlined in Algorithm~\ref{alg:SCFsubOCCA}
to solve NEPv  \eqref{eq:NepsubOCCA} for the purpose of solving
\eqref{eq:subOCCA}.

\begin{algorithm}[H]
	\caption{An SCF iteration for solving \eqref{eq:subOCCA}} \label{alg:SCFsubOCCA}
	\begin{small}
		\begin{algorithmic}[1]
			\REQUIRE $G_{(0)} \in \bbO^{n\times k}$;
			\ENSURE  approximate maximizer $G$ to \eqref{eq:subOCCA}.
			\FOR{$\nu=1,2,\ldots$ until convergence}
			\STATE construct $E_{(\nu)} = E(G_{(\nu-1)})$ as in \eqref{eq:NepsubOCCA};
			\STATE compute an orthonormal eigenbasis matrix $G_{(\nu)}$ associated with the $k$ smallest eigenvalues of  $E_{(\nu)}$;
			\STATE compute the SVD of $G_{(\nu)}^{\T}D=U \Sigma V^{\T}$ and update $G_{(\nu)}\leftarrow G_{(\nu)} UV^{\T}$;
			\ENDFOR
			\RETURN the last $G_{(\nu)}$ as a numerical maximizer of \eqref{eq:subOCCA}.
		\end{algorithmic}
	\end{small}
\end{algorithm}

\begin{remark}
	Line 3 in Algorithm \ref{alg:SCFsubOCCA} is justified by Lemma \ref{lem:key} because the chosen $G_{(\nu)}$
	satisfies
	\begin{small}
		\begin{align*}
		\tr(G_{(\nu)}^{\T}E(G_{(\nu-1)})G_{(\nu)})\le \tr(G_{(\nu-1)}^{\T}E(G_{(\nu-1)})G_{(\nu-1)})
		\end{align*}
	\end{small}\noindent
	and thus $\eta(G_{(\nu)})\ge\eta(G_{(\nu-1)})$.
	To understand line 4, we note that an eigenbasis matrix is not unique. In fact,
	$\what G_{(\nu)}=G_{(\nu)}P$ for any $P\in\bbO^{k\times k}$ is also one. Since $\tr(\what G_{(\nu)}^{\T}A\what G_{(\nu)})\equiv \tr( G_{(\nu)}^{\T}A G_{(\nu)})$
	but $\tr^2(\what G_{(\nu)}^{\T}D)\ne\tr^2(G_{(\nu)}^{\T}D)$ in general, it makes sense to update $G_{(\nu)}$ to $\what G_{(\nu)}$ so that
	$\tr^2(\what G_{(\nu)}^{\T}D)$ is maximized over $P\in\bbO^{k\times k}$. That is when Lemma \ref{lm:maxtrace} comes to help.
\end{remark}

Because an eigenbasis matrix is not unique, one may ask if $G_{(\nu)}$ at line 4 in Algorithm \ref{alg:SCFsubOCCA}
is well-defined. The next theorem addresses this issue.

\begin{theorem}\label{thm:unique}
	At line 3 of  Algorithm \ref{alg:SCFsubOCCA}, if the eigenvalue gap
	
	\begin{small}\vspace{-0.1in}
		\begin{equation}\nonumber
		\zeta_{\nu-1}=\lambda_{k+1}(E(G_{(\nu-1)}))-\lambda_{k}(E(G_{(\nu-1)}))>0,
		\end{equation}
	\end{small}\noindent
	then any two orthonormal eigenbasis matrices $\widehat G_{(\nu)}$ and $\widetilde G_{(\nu)}$ associated
	with  $k$ smallest eigenvalues of $E(G_{(\nu-1)})$ satisfy $\widetilde G_{(\nu)}=\widehat G_{(\nu)}Q$
	for some orthogonal matrix $Q\in\mathbb{O}^{k\times k}$.
	Furthermore, if additionally $\rank(D^{\T}{\widehat G_{(\nu)}})=k$, then the
	next approximation $G_{(\nu)}$ from line 4 of Algorithm \ref{alg:SCFsubOCCA} is uniquely determined.
\end{theorem}
 
\vspace{-0.1in}
\subsection{Convergence analysis}\label{sec:convergence}
We next provide some basic convergence properties of the simple SCF iteration (Algorithm \ref{alg:SCFsubOCCA})
for solving \eqref{eq:subOCCA}.

\begin{theorem}\label{thm:convg1}
	Let  the sequence $\{G_{(\nu)}\}$  be generated by the SCF iteration (Algorithm \ref{alg:SCFsubOCCA}). Then
	\begin{itemize}
		\item[(i)] For each $\nu\ge 1$, $D^{\T}G_{(\nu)}\succeq 0$ and $\tr(G_{(\nu)}^{\T}D)=\sum_{j=1}^k\sigma_j(G_{(\nu)}^{\T}D)$;
		\item[(ii)] The sequence $\{\eta(G_{(\nu)})\}$ is monotonically increasing and convergent;
		\item[(iii)]  If
		\begin{small}\vspace{-0.1in}
			\begin{align}\nonumber
			&\tr(G_{(\nu)}^{\T}E(G_{(\nu-1)}) G_{(\nu)})\\
			&\qquad< \tr(G_{(\nu-1)}^{\T}E(G_{(\nu-1)}) G_{(\nu-1)}), \label{eq:inexacteig}
			\end{align}
		\end{small}\noindent
		then $\eta(G_{(\nu-1)})<\eta(G_{(\nu)})$;
		\item[(iv)] $\{G_{(\nu)}\}$ has a convergent subsequence $\{G_{(\nu)}\}_{\nu\in \mathcal{I}}$;
		\item[(v)] Let  $\{G_{(\nu)}\}_{\nu\in \mathcal{I}}$  be any convergent subsequence of $\{G_{(\nu)}\}$ with the accumulation
		point $G_*$ satisfying
		\begin{equation}\label{eq:gapcondition}
		\zeta=\lambda_{k+1}(E(G_*))-\lambda_{k}(E(G_*))>0.
		\end{equation}
		Then $G_*$ satisfies the first order optimality condition in Lemma \ref{lem:KKTsubOCCA}
		and also the necessary condition  for a global minimizer in Theorem \ref{thm:NepsubOCCA}.
	\end{itemize}
\end{theorem}
\begin{remark}\label{rk:convg1}
	We have three remarks for Theorem \ref{thm:convg1}.
	\begin{itemize}
		\item[(a)]
		Item (iii) of Theorem \ref{thm:convg1} implies  that, to only guarantee   monotonicity of $\{\eta(G_{(\nu)})\}$,
		the partial eigen-decomposition in line 3 of Algorithm \ref{alg:SCFsubOCCA} can be inexact. In particular, in line 3,
		we can choose any approximation $G_{(\nu)}\in \bbO^{n\times k}$ satisfying \eqref{eq:inexacteig}, and then refine it by line 4
		to ensure $D^{\T} G_{(\nu)}\succeq 0$; by Lemma \ref{lem:key}, $\eta(G_{(\nu)})> \eta(G_{(\nu-1)})$ holds too.
		This facilitates us to employ certain sophisticated eigensolver for the computation task in line 3.
		\item[(b)] Item (iv) is rather obvious because $\{G_{(\nu)}\}$ is a bounded sequence in $\bbR^{n\times k}$. It is explicitly listed
		to substantiate part of the assumption in item (v). A stronger claim in Theorem~\ref{thm:convg2} later
		says the entire sequence $\{G_{(\nu)}\}$ converges under a mild condition.
		\item[(c)] Item  (v) shows one of advantages of our SCF iteration over the generic Riemannian optimization
		methods for solving the core subproblem \eqref{eq:subOCCA}. In particular, as our SCF iteration is
		built upon the necessary conditions of the global maximizer $G_{\opt}$, besides the general KKT conditions,
		the convergent point also fulfills certain necessary conditions for being a global maximizer.
	\end{itemize}
\end{remark}

For further analyzing convergence  of the sequence $\{G_{(\nu)}\}$, we now consider the sequence $\{\cR(G_{(\nu)})\}$ of subspaces.
For this purpose, we denote by
$\|\cdot\|_{\UI}$ any unitarily invariant norm, and
introduce the distance measure between two subspaces $\cG$ and $\cY$ of dimension $k$ \cite[p.95]{sun:1987}

\begin{small}\vspace{-0.1in}
	\begin{equation}\label{eq:sinTheta}
	\dist_{\UI}(\cG,\cY):=\|\sin\Theta(\cG,\cY)\|_{\UI}
	\end{equation}
\end{small}\noindent
in terms of the matrix of canonical angles between $\cG$ and $\cY$:

\begin{small}\vspace{-0.1in}
	$$
	\Theta(\cG,\cY)=\diag(\theta_1(\cG,\cY),\dots,\theta_k(\cG,\cY)).
	$$
\end{small}\noindent
Let
$\cG=\cR(G)$ and $\cY=\cR(Y)$, where $G,\,Y\in\bbR^{n\times k}$ with
$G^{\T}G=Y^{\T}Y=I_k$. The canonical angles
$\theta_1(\cG,\cY)\ge\cdots\ge\theta_1(\cG,\cY)$ is defined by

\begin{small}\vspace{-0.1in}
	\begin{align}
	0\le\theta_i(\cG,\cY):=\arccos \sigma_i(G^{\T}Y)\le\frac {\pi}2 \quad\mbox{for $1\le i\le k$}.\nonumber
	\end{align}
\end{small}\noindent
The collection of all $k$-dimensional subspaces in $\bbR^n$ is the so-called Grassmann manifold
$\scrG_k(\bbR^n)$,
and  \eqref{eq:sinTheta} is a unitarily invariant metric \cite[p.95]{sun:1987} on $\scrG_k(\bbR^n)$.
For the trace norm, also known as the nuclear norm, we have

\begin{small}\vspace{-0.1in}
	$$
	\dist_{\tr}(\cG,\cY)=\sum_{j=1}^k\sin\theta_j(\cG,\cY).
	$$
\end{small}\noindent
Using  the metric $\dist_{\tr}(\cG,\cY)$, we have the following convergence result for the sequence $\{G_{(\nu)}\}$ by the SCF iteration
in Algorithm \ref{alg:SCFsubOCCA}.

\begin{theorem}\label{thm:convg2}
	Let  the sequence $\{G_{(\nu)}\}$  be generated by the SCF iteration (Algorithm \ref{alg:SCFsubOCCA}),
	and let $G_*$ be an accumulation point of $\{G_{(\nu)}\}$.
	\begin{itemize}
		\item[(i)] If $\cR(G_*)$  is an isolated accumulation point
		(in the metric \eqref{eq:sinTheta}) of $\{\cR(G_{(\nu)})\}_{\nu=0}^{\infty}$, then $\{\cR(G_{(\nu)})\}_{\nu=0}^{\infty}$
		converges to $\cR(G_*)$.
		\item[(ii)] If also  $\rank(G_*^{\T}D)=k$, then $\{G_{(\nu)}\}_{\nu=0}^{\infty}$ converges to $G_*$
		(in the standard Euclidean metric), and for sufficiently large $\nu$,
		
		\begin{small}\vspace{-0.1in} 	
			\begin{align}\label{eq:ratea}
			\dist_{\tr}(\cR(G_*),\cR(G_{(\nu+1)}))\le c_0\|G_{(\nu)}-G_*\|_{\tr},
			\end{align}
		\end{small}\noindent	
		where
		
		\begin{small}\vspace{-0.1in}
			\begin{align*}
			c_0=\frac{3\|D\|_2}{\zeta}\left(\sqrt{\frac{\|A\|_{(k)}}{\eta(G_*)}}
			+2k\frac{\|A\|_2+\sqrt{\frac{\|A\|_{(k)}}{\eta(G_*)}}\|D\|_2}{\sqrt{\eta(G_*)\omega_k(A)}}\right),
			\end{align*}
		\end{small}\noindent
		with {\small $\|A\|_{(k)}=\sum_{j=1}^k\sigma_j(A)$, $\omega_k(A)=\sum_{j=1}^k\sigma_{n-j+1}(A)$}, and
		the eigenvalue gap $\zeta>0$ is given in \eqref{eq:gapcondition}.
	\end{itemize}
\end{theorem}

\section{Orthogonal Multiset CCA}\label{sec:MOCCA}

In this section, we propose to solve a new formulation of OMCCA based on the proposed methods in Sections \ref{sec:occa-2v}
and \ref{sec:SCFAlt}.

\subsection{Problem Formulation} \label{sec:MOCCA-weighting}
Let $S_i$ and $C_{i,j}$ be the ones defined in Subsection~\ref{ssec:MCCA}.
Related to notion of the set of canonical weight matrices for MCCA, similarly to \eqref{eq:CCA-multi},
a general model is to seek canonical weight matrices $X_i\in\bbR^{n_i\times k}$ that solve
\\	
\begin{small}\vspace{-0.1in}
	\begin{equation}\label{eq:mset-CCA-mtx}
	\max_{\{X_i\}}\, f(\{X_i\}),\quad \mbox{s.t.}\,\,X_i^{\T} C_{i,i} X_i=I_k,\,i=1,\ldots,\ell,
	\end{equation}
\end{small}\noindent
where $1\le k\le\min\{n_1,\ldots,n_{\ell},q\}$, and
\\
\begin{small}\vspace{-0.1in}
	\begin{equation}\label{eq:mset-CCA-f}
	f(\{X_i\})=\sum_{\begin{subarray}{l} i,\,j=1\\
		i\ne j
		\end{subarray} }^\ell
	{\rho}_{ij}\frac {\tr (X_i^{\T} C_{i,j} X_j)}
	{\sqrt{\tr(X_i^{\T} C_{i,i} X_i)}\sqrt{\tr(X_j^{\T} C_{j,j} X_j)}},
	\end{equation}
\end{small}\noindent	
with some weighting factors ${\rho}_{ij}\ge 0$ which will not only dictate the contribution of the correlation between $S_i$ and $S_j$ to the total $f(\{X_i\})$
but also, as we will see later, dramatically reduce the number terms in $f(\{X_i\})$ and thus speed up computations.
Most of all, we assume that judiciously chosen ${\rho}_{ij}$ with only a few of them nonzero
can in fact improve the performances of muti-view tasks, which will be verified by the experiments.
Analogously to \eqref{eq:OCCA}, OMCCA naturally arises:

\begin{small}\vspace{-0.1in}
	\begin{equation}\label{eq:OMCCA}
	\max_{\{X_i\in\bbO^{n_i\times k}\}}\,f(\{X_i\}).
	\end{equation}
\end{small}\noindent

Ideally, the optimal weights should be learned from data, but this is out of the scope for this paper. Hence, we take some heuristic weighting schemes.
To begin with, we define

\begin{small}\vspace{-0.05in}
	\begin{equation}
	\widehat{\rho}_{ij}=
	\frac{ \sum_{r=1}^{\rank(C_{i,j})} \sigma_r(C_{i,j})   }{\sqrt{\tr( C_{i,i} ) \tr(C_{j,j})}}, ~~\mbox{for}\,\, i, j=1,\ldots,\ell. \label{eq:weight}
	\end{equation}
\end{small}\noindent
It is known $0\le \widehat{\rho}_{ij}\le 1$ \cite[(3.5.22) on p.212]{hojo:1991}.
Envision a graph of $\ell$ nodes corresponding to dataset $X_i$, respectively, with every two nodes connected with an edge
whose weight is to be determined. We now explain our heuristic strategies to select the weights.
\begin{enumerate}
	\item uniform weighting: $\widehat{\rho}_{ij} = 1, \forall i, j=1,\ldots,\ell$.
	\item tree weighting: find the minimal spanning tree of the graph with the edge $(i,j)$ having weight $1-\widehat{\rho}_{ij}$, record
	the spanning tree with its edge weights reset back to $\widehat{\rho}_{ij}$ and weights $\widehat{\rho}_{ij}$ for all other edges not in the tree
	reset to $0$.
	\item top-$p$ weighting: find the $p$ largest weights among $\widehat{\rho}_{ij}$ for $i>j$, and reset all other weights $\widehat{\rho}_{ij}$
	to $0$.
\end{enumerate}
Next we  apply the soft-max function over those selected weights $\widehat{\rho}_{ij}$ with a bandwidth parameter
(e.g., $20$ used in our experiments) to yield ${\rho}_{ij}$ to use in \eqref{eq:mset-CCA-f}.
As a by-product, the sum of all ${\rho}_{ij}$ is $1$.

Unfortunately, there does not seem to have an efficient way to solve \eqref{eq:mset-CCA-mtx} or \eqref{eq:OMCCA}, beyond primitive
optimization techniques such as plain gradient ascent iteration or the trust-region methods. Next, we extend Algorithm \ref{alg:AltOCCA} for $\ell=2$ to solve \eqref{eq:OMCCA}.

\vspace{-0.1in}
\subsection{The Proposed Algorithm}

Problem (\ref{eq:OMCCA}) cannot be simply solved by maximizing
each individual term in $f(\{X_i\})$ of \eqref{eq:mset-CCA-f} separately; otherwise each dataset would have more than one projection matrix $X_i$.
What we plan to do, based on the machinery we have built in the previous sections, is to
optimize $f(\{X_i\})$ cyclically over each matrix variable $X_i$ in the styles similar to
either the Jacobi  or Gauss-Seidel iteration for linear systems \cite{demm:1997}.
Specifically, we establish an inner-outer iterative method  to solve \eqref{eq:OMCCA}. 
The most outer iteration -- each step called a cycle --  generates from the
current approximation $\{X_i^{(\nu)}\}_{i=1}^{\ell}$  to the next $\{X_i^{(\nu+1)}\}_{i=1}^{\ell}$
of the maximizer of \eqref{eq:OMCCA}; each cycle can be of an either Jacobi-style or
Gauss-Seidel-style updating scheme that relies on the proposed novel SCF iteration for solving
a series of subproblems in the form of (\ref{eq:subOCCA}).

%

Due to the possibility that $\rank(C_{i,i})\le\min\{n_i-1,q\}$ and possibly $q\ll n_i$, numerical
difficulties may arise and will arise when $q\ll n_i$. To circumvent them, we propose to add range constraints

\begin{small}\vspace{-0.1in}
	\begin{equation}\label{eq:PiXi}
	\cR(X_i)\subset\cR(S_i)\quad\mbox{for $i=1,2,\ldots,\ell$}.
	\end{equation}
\end{small}\noindent
In what follows, we describe an SVD-based implementation.
Let the SVDs of $S_i$ be

\begin{small}\vspace{-0.1in}
	\begin{equation}\label{eq:SVD4Xi}
	S_i=U_i\Sigma_iV_i^{\T},\,\, U_i\in\bbR^{n_i\times r_i},\,\,V_i\in\bbR^{q\times r_i},\,\,\Sigma_i\in\bbR^{r_i\times r_i},
	\end{equation}
\end{small}\noindent
where $r_i=\rank(S_i)$.
With the SVDs in \eqref{eq:SVD4Xi}, we have

\begin{small}\vspace{-0.1in}
	$$
	X_i^{\T}S_iS_j^{\T}X_j={X_i^{\T}U_i}\Sigma_i V_i^{\T}V_j\Sigma_j{U_j^{\T}X_j}
	=:\what X_i^{\T}\Sigma_i V_i^{\T}V_j\Sigma_j\what X_j,
	$$
\end{small}\noindent
where $\what X_i=U_i^{\T}X_i\in\bbR^{r_i\times k}$. Under \eqref{eq:PiXi}, we will have $X_i=U_i\what X_i.$
The function $f(\{X_i\})$ is then transformed into

\begin{small}\vspace{-0.1in}
	\begin{align}\nonumber
	\sum_{i\ne j}{\rho}_{ij}\frac {\tr(\what X_i^{\T}\Sigma_i V_i^{\T}V_j\Sigma_j\what X_j)}
	{\sqrt{\tr(\what X_i^{\T}\Sigma_i^2\what X_i)}\sqrt{\tr(\what X_j^{\T}\Sigma_j^2\what X_j)}}
	=:g(\{\what X_i\}),
	\end{align}
\end{small}\noindent
and

\begin{small}\vspace{-0.1in}
	\begin{align}\nonumber
	\max_{X_i\in\bbO^{n_i\times k},\,\cR(X_i)\subset\cR(S_i),~\forall i}f(\{X_i\})
	=\max_{\what X_i\in\bbO^{r_i\times k},\,\forall i}g(\{\what X_i\}).
	\end{align}
\end{small}\noindent
The key step to maximize $g(\{\what X_i\})$ by either the Jacobi- or Gauss-Seidel-style updating scheme
is to maximize it, for any $s\in\{1,\cdots,\ell\}$, over $\what X_{s}$ while keeping all other $\what X_j$ for $j\ne s$.
That is equivalent to
\begin{small}\vspace{-0.05in}
	\begin{equation}\label{eq:MOCCA-i1}
	\max_{\what X_{s}\in\bbO^{n_{s}\times k}}
	\frac {\tr(\what X_{s}^{\T}D_{s})}{\sqrt{\tr(\what X_{s}^{\T}\Sigma_{s}^2\what X_{s})}},
	\end{equation}
\end{small}\noindent
where

\begin{small}\vspace{-0.1in}
	\begin{equation}\label{eq:Ds}
	D_{s}(\{\what X_i\}_{i\ne s})=\Sigma_{s}V_{s}^{\T}\sum_{j\ne s} {\rho}_{sj}\frac {V_j\Sigma_j\what X_j}{\sqrt{\tr(\what X_j^{\T}\Sigma_j^2\what X_j)}}.
	\end{equation}
\end{small}\noindent
Problem \eqref{eq:MOCCA-i1} is equivalent to solving:
\begin{small}\vspace{-0.05in}
	\begin{equation}\label{eq:MOCCA-i0}
	\max_{\what X_{s}\in\bbO^{n_{s}\times k}}
	\frac {\tr^2(\what X_{s}^{\T}D_{s})}{ \tr(\what X_{s}^{\T}\Sigma_{s}^2\what X_{s})},
	\end{equation}
\end{small}\noindent
which takes the same form as \eqref{eq:subOCCA}, the subject of which has been studied in Section~\ref{sec:SCFAlt}.


Algorithm~\ref{alg:RCMOCCA} outlines the framework of two inner-outer numerical methods, where
each cycle (lines 6--9) follows either the Jacobi-style or Gauss-Seidel-style updating scheme.
A couple of comments are in order for efficiently implementing Algorithm~\ref{alg:RCMOCCA}.

First, evaluating
all $D_s$ as written in \eqref{eq:Ds} is rather costly when majority or all of ${\rho}_{ij}$ are nonzero.
But in the case when all ${\rho}_{ij}$ are the same, say $1$, there are many common terms among all $D_s$
for $s=1,2,\ldots, \ell$, and that should be taken advantage of. However, for two weighting strategies
-- tree weighting and top-$p$ weighting, we previously
discussed, most $D_s$ only has very few terms in its summation and evaluating $D_s$ is not a concern at all.

The second comment is that, for the Jacobi-style updating scheme, each subproblem \eqref{eq:MOCCA-i0}
is independent within one full cycle: line 6-9, and thus they can be solved in parallel.
As in the case for the linear system \cite{demm:1997}, often the Gauss-Seidel-style updating scheme
converges faster than the Jacobi-style updating scheme in terms of the cycle counts. However,
sometimes, the built-in parallelism  in the later may well compensate that disadvantage in the wall-clock.

%

\begin{algorithm}[t]
	\caption{RCOMCCA: Range Constrained OMCCA}\label{alg:RCMOCCA}
	\begin{small}
		\begin{algorithmic}[1]
			\REQUIRE $\{S_i\in\bbR^{n_i\times q}\}$ (each $S_i$ is centered),
			integer $1\le k\le\min\{\rank(S_1),\ldots,\rank(S_{\ell}),q\}$, and a tolerance $\epsilon$;
			
			\ENSURE  $\{X_i\in\bbO^{n_i\times k}\}$ that  maximizes
			$f(\{X_i\})$.
			
			\STATE compute the SVDs in \eqref{eq:SVD4Xi};
			
			\STATE pick an initial approximation $\what X_1^{(0)}$;
			
			\STATE $\nu=0$, $g=0$;
			
			\REPEAT
			\STATE $g_0=g$; $g=0$;
			\FOR{$s=1$ to $\ell$}
			\STATE compute the next $\{\what X_s^{(\nu+1)}\}$ by solving \eqref{eq:MOCCA-i0}, where either
			\begin{equation}\label{eq:Ds-alg-J}
			D_s=D_s(\{\what X_i^{(\nu)}\}_{i\neq s})
			\end{equation}
			for the Jacobi-style updating scheme,   or
			\begin{equation}\label{eq:Ds-alg-GS}
			D_s=D_s(\what X_1^{(\nu+1)},\ldots,\what X_{s-1}^{(\nu+1)},\what X_{s+1}^{(\nu)},\ldots,\what X_{\ell}^{(\nu)})
			\end{equation}
			for Gauss-Seidel-style updating scheme.
			\STATE $g=g+g_s$, where $g_s$ is the computed optimal objective value of \eqref{eq:MOCCA-i0}.		
			\ENDFOR
			\STATE $\nu=\nu+1$;
			\UNTIL{$|g-g_0|\le\epsilon g$;}
			\RETURN $X_i=U_i\what X_i^{(\nu)}$ for $1\le i\le\ell$.
		\end{algorithmic}
	\end{small}
\end{algorithm}

\vspace{-0.1in}
\section{Experiments} \label{sec:experiments}

\subsection{Implementation details}
For a practical implementation of the  SCF in Algorithm \ref{alg:SCFsubOCCA}, when the matrix size $n\le 500$,
we call the mex version \texttt{mexeig} of the LAPACK \cite{abbd:1999} eigen-decomposition subroutine
\texttt{dsyevd}\footnote{\texttt{mexeig} (available at: {\tt www.math.nus.edu.sg/$\sim$matsundf/}) is a
	MATLAB interface to call LAPACK eigen-decomposition subroutine \texttt{dsyevd} of a real symmetric matrix.}
to compute  $G_{(\nu)}$, while for matrix size $n>500$, we choose the locally optimal block preconditioned
conjugate gradient method\footnote{The MATLAB version of \textsf{lobpcg} is
	available at: http://cn.mathworks.com/matlabcentral/fileexchange/48-lobpcg-m.} (\textsf{lobpcg}) (see  \cite{knya:2001,knne:2003})
with the diagonal preconditioner $P=\diag(A)$ to compute an approximation orthonormal eigenbasis matrix $G_{(\nu)}$  in line 3.
As \textsf{lobpcg} searches an approximation eigenbasis matrix $G_{(\nu)}$ by optimizing
the Rayleigh quotient in a subspace containing $G_{(\nu-1)}$, the condition \eqref{eq:inexacteig} is always fulfilled,
meaning (by Theorem \ref{thm:convg1}) that the sequence $\{\eta(G_{(\nu)})\}$ is monotonically increasing and convergent.
The SCF iteration of Algorithm \ref{alg:SCFsubOCCA} terminates if $\nu>30$ or

\vspace{-0.1in}
{\small
	$$
	\frac{\|{\rm grad}(\eta(G_{(\nu)}))\|_1}{\xi^2(G_{(\nu)})(\|A\|_1+\|D\|_1)}\le \epsilon_{\scf}
	~~ {\rm or}~~
	\left|\frac{\eta(G_{(\nu)})-\eta(G_{(\nu-1)})}{\eta(G_{(\nu)})}\right|\le \epsilon_{\scf}^{3/2}
	$$
}\noindent
with $\epsilon_{\scf}=10^{-5}$ and $\xi(G_{(\nu)})$ is defined in (\ref{eq:M}).

\subsection{Comparisons with generic optimization methods}
We conduct extensive experiments to compare our  SCF-based alternative iteration (Algorithm \ref{alg:AltOCCA})
with three generic optimization methods over matrix manifolds implemented in LDR
toolbox\footnote{https://github.com/cunni/ldr} \cite{cugh:2015}. They are {\textsf{stiefel, stiefel\_trust, and stiefel\_trust\_prod}}
for solving problem \eqref{eq:OCCA} (i.e., $\ell=2$).
In particular, \textsf{stiefel, stiefel\_trust} are based similarly on our  scheme in Algorithm \ref{alg:AltOCCA} to update $X$ and $Y$ alternatively, but solve the subproblems \eqref{eq:subOCCA} by the generic Riemannian steepest gradient iteration and the Riemannian trust-region method \cite{abms:2008}, respectively; the third one \textsf{stiefel\_trust\_prod} just employs the plain Riemannian Trust-Region (RTR) of \cite{abms:2008} to solve \eqref{eq:OCCA} directly. We use the default settings of these three algorithms coded in the LDR toolbox, in which  \textsf{stiefel, stiefel\_trust} stop whenever the number of alternative steps $\nu>100$ or

\begin{small}\vspace{-0.1in}
	\begin{equation}\label{eq:tolrelf}
	\left|\frac{{f(X^{(\nu)}, Y^{(\nu)})}-{f(X^{(\nu-1)}, Y^{(\nu-1)})}}{{f(X^{(\nu)}, Y^{(\nu)})}}\right|\le \epsilon_{\alt},
	\end{equation}
\end{small}\noindent
with $\epsilon_{\alt}=10^{-8}$, whereas \textsf{stiefel\_trust\_prod} uses the default setting of  RTR \cite{abbg:2007} in the package \texttt{manopt}. For comparisons,  we terminate our  SCF-based Algorithm \ref{alg:AltOCCA} if $\nu>30$ or when the norm of Riemannian gradient ${\rm grad\_{norm}}=\|{\rm grad}F(X^{(\nu)}, Y^{(\nu)})\|_2\le \epsilon_{\alt}$ or \eqref{eq:tolrelf} holds with $F(X,Y)$ instead of $f(X,Y)$.

Our experiments are performed over a synthetic data with $m=n=1000$ by varying $k \in [3, 100]$, and the yeast data shown in Table \ref{tab:mlc} with $m=101, n=14$ by varying $k \in [2, 14]$.
Following  in \cite{cugh:2015}, we generate the synthetic data with $q=10^4$ and two views controlled by two sets of latent variables $W$ and $Z$ as follows:

\begin{small}\vspace{-0.1in}
	\begin{align*}
	&d_z = \left\lceil \frac{\max(m,n)}{2} \right\rceil, d_w = \left\lceil \frac{2 \max(m,n)}{5} \right\rceil,\\
	&S_X = P_X Z + Q_X W + \lambda E_X, \\
	&S_Y = P_Y Z + Q_Y W + \lambda E_Y,
	\end{align*}
\end{small}\noindent
where $Z \in \bbR^{d_z \times q}$, $W \in \bbR^{d_w \times q}$, $P_X \in \bbR^{m \times d_z}$, $Q_X \in \bbR^{m \times d_w}$, $P_Y \in \bbR^{n \times d_z}$, $Q_Y \in \bbR^{n \times d_w}$, $E_X \in \bbR^{m \times q}$, and $E_Y \in \bbR^{n \times q}$ are matrices where each entry is i.i.d. sampled from a normal distribution with zero mean and unit standard deviation, and $\lambda=2 \times 10^{-4}$.

The performance is evaluated in terms of the following criteria:
\begin{enumerate}
	\item Computational complexity measured by CPU time;
	\item The correlation difference. The differences of the objective values between the SCF-based iteration and each of the other three methods are reported. The larger the difference is, the better  the SCF-based iteration performs than the other;
	\item  The gradient norm ${\rm grad\_{norm}}$ at the computed solution.
\end{enumerate}

\begin{figure}[t]
	\begin{tabular}{@{}c@{}c@{}}
		\hline
		simulation data & yeast \\
		\hline
		\includegraphics[width=0.5\textwidth]{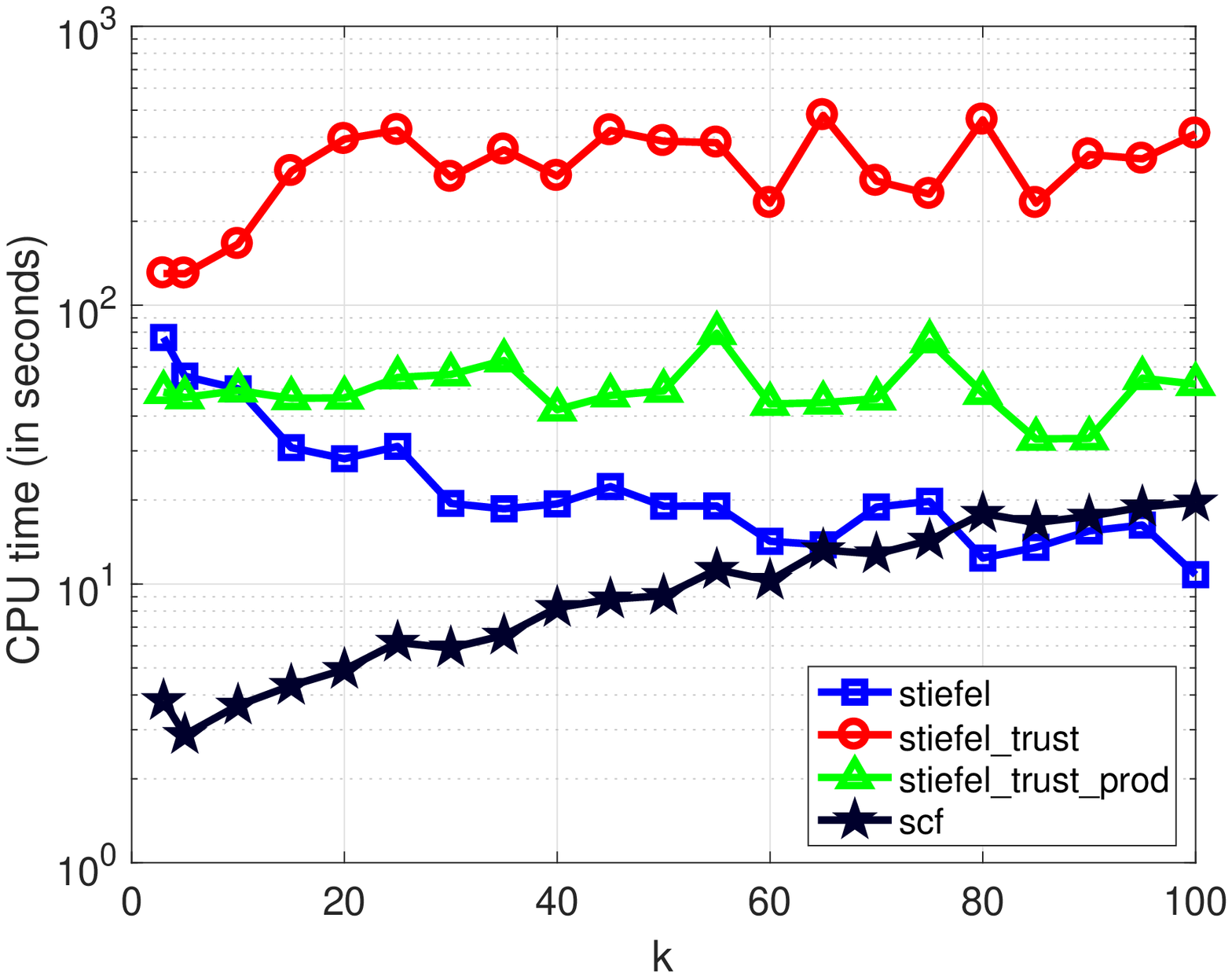} & \includegraphics[width=0.5\textwidth]{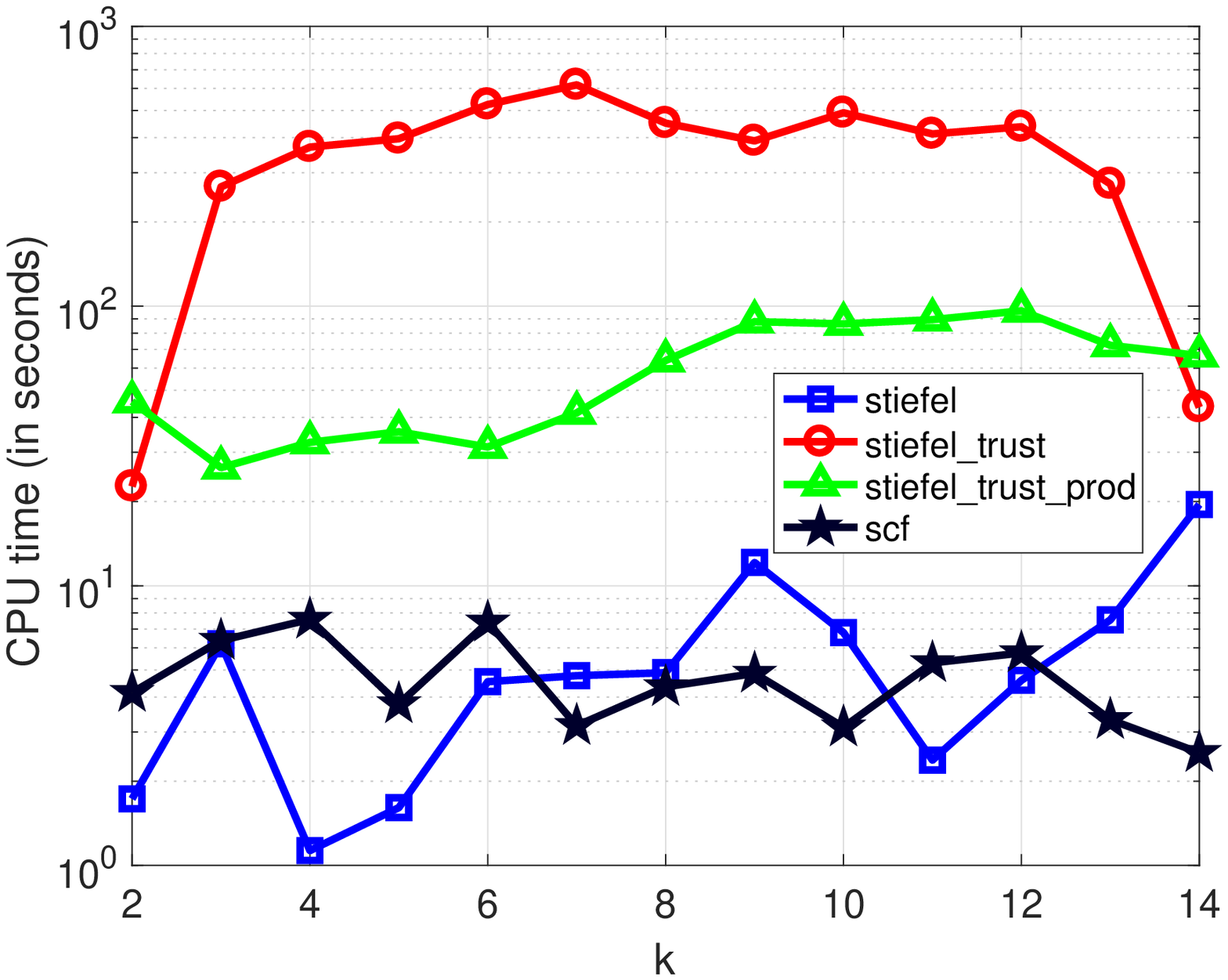}\\ \includegraphics[width=0.5\textwidth]{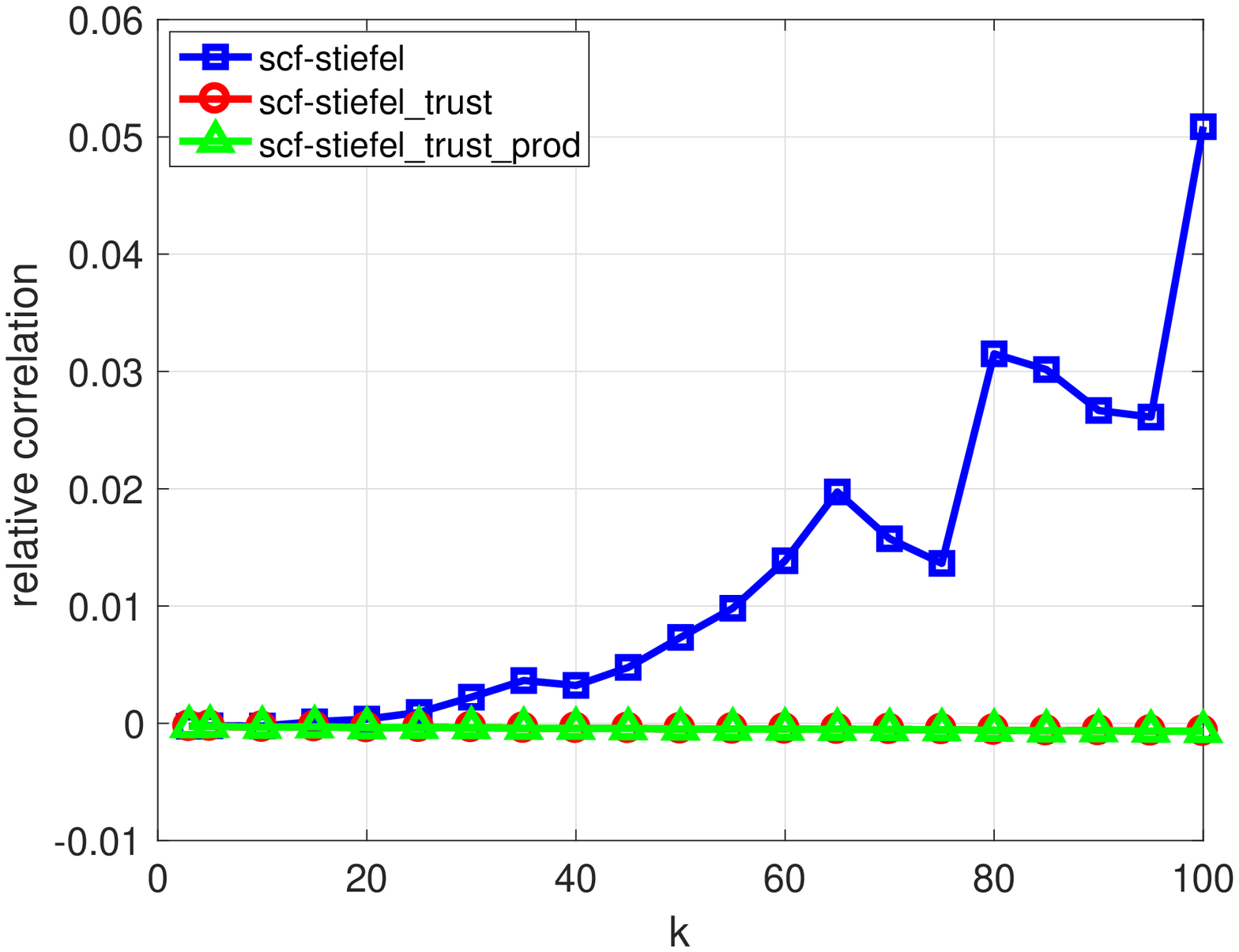} &
		\includegraphics[width=0.5\textwidth]{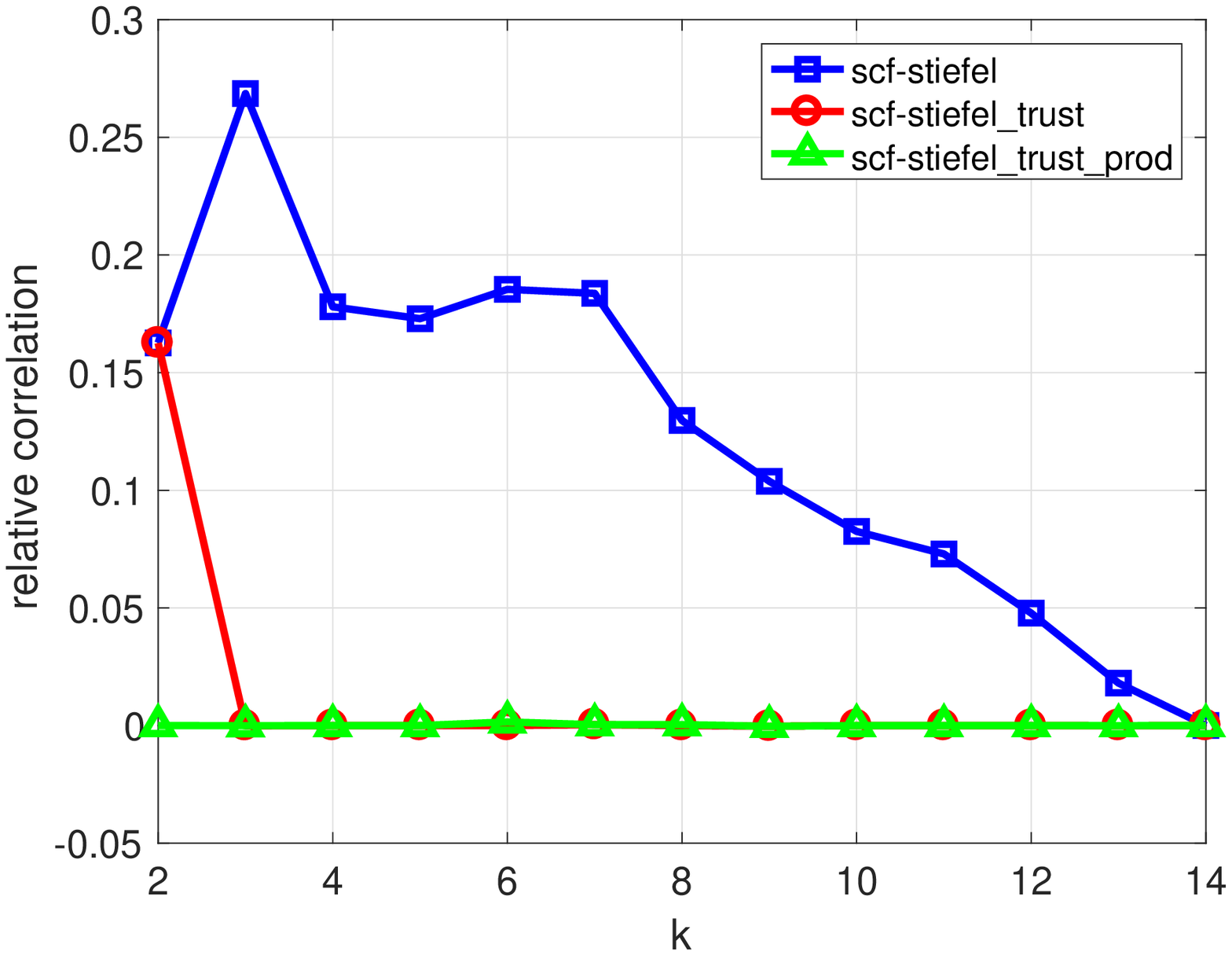} \\
		\includegraphics[width=0.5\textwidth]{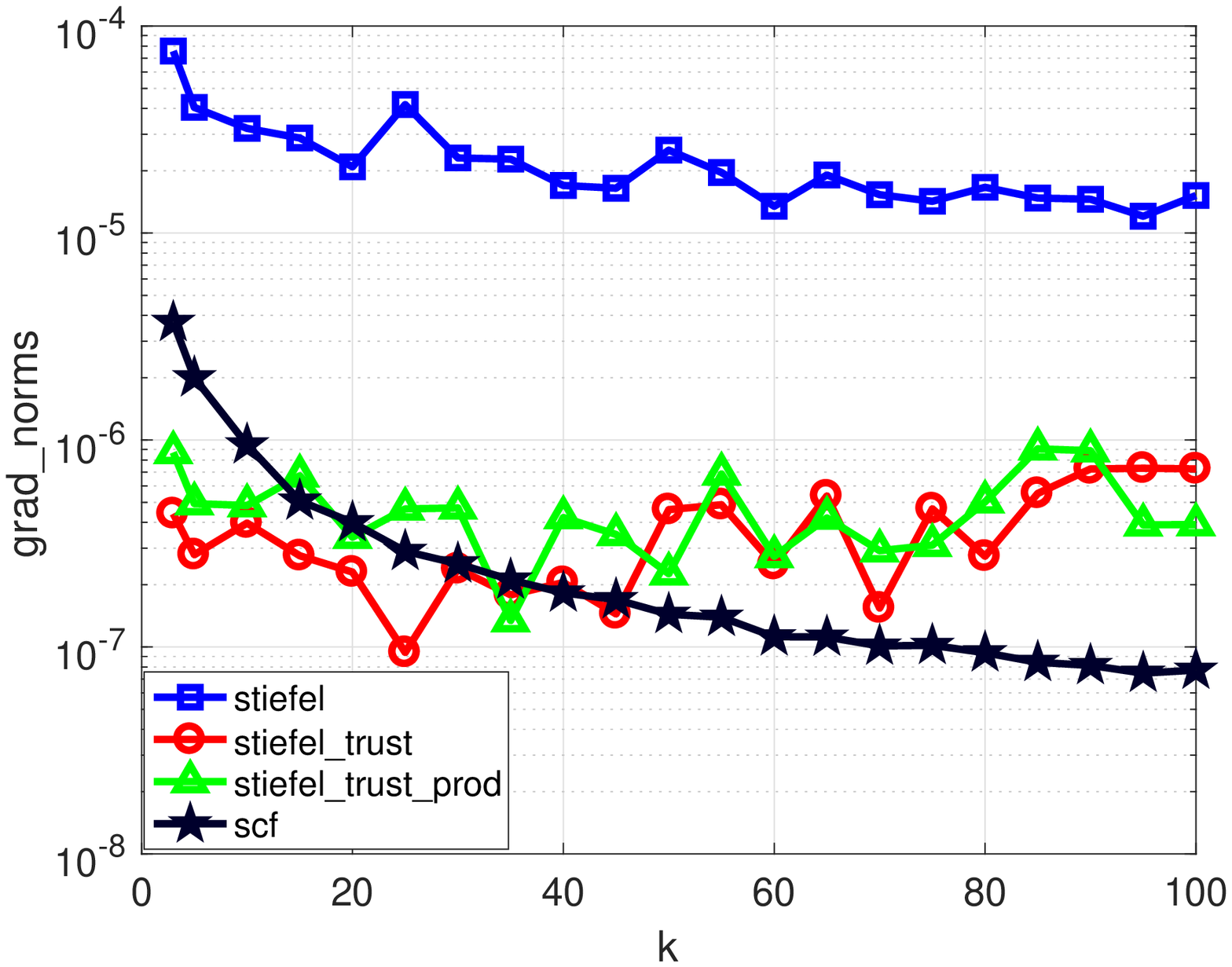} & \includegraphics[width=0.5\textwidth]{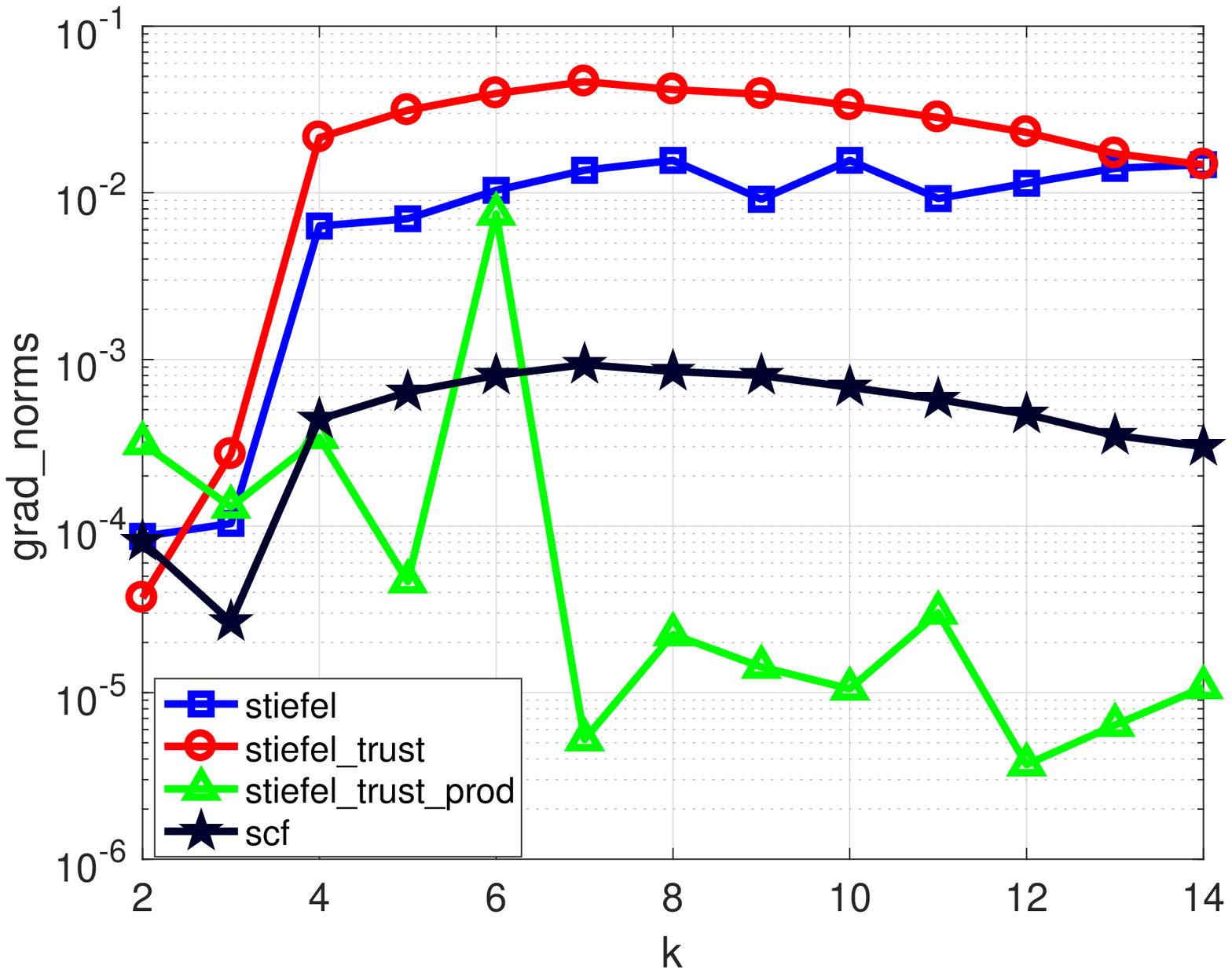}
	\end{tabular}
	\caption{The comparison of four optimization methods in terms of three different criteria evaluated on the synthetic data and real multi-label classification data yeast.} \label{fig:ldr}
\end{figure}

Fig. \ref{fig:ldr} shows the numerical results obtained by four different methods (all starting with the initial guess $\{X^{(0)},Y^{(0)}\}=\{I_{n,k}, I_{m,k}\}$ where $I_{n,k}$ contains the first $k$ columns of $I_n$) in terms of three performance measurements over the synthetic data and yeast for multi-label classification. We have the following observations:
\begin{enumerate}
	\item For small $k$, our SCF-based alternative iteration converges much faster than others.
	The time cost of SCF-based method is similar to \textsf{stiefel}, while \textsf{stiefel\_trust} is most expensive among all.
	
	\item The SCF-based iteration obtains the similar correlation values on both data with \textsf{stiefel\_trust} and \textsf{stiefel\_trust\_prod}. \textsf{stiefel} is the worst, and also the correlations over various $k$ values demonstrate opposite trends on the synthetic data and yeast. This implies that \textsf{stiefel} is quite sensitive to the value $k$ and the input data.
	
	\item Overall, our SCF-based iteration is an efficient way to solve the two-view OCCA problem \eqref{eq:OCCA}.
	Except for  consuming more CPU time, \textsf{stiefel\_trust\_prod} is also able to compute an accurate approximation;
	however, our SCF-based iteration can be easily extended to solve the more general MOCCA model \eqref{eq:OMCCA}
	in the framework of Algorithm \ref{alg:RCMOCCA}, and thus naturally, we choose it as a core engine for subproblems
	in line 7 of Algorithm \ref{alg:RCMOCCA} in our subsequent experiments.
	
\end{enumerate}

\begin{figure}
	\centering
	\begin{tabular}{@{}c@{}}
		\!\!\!\!\!\!\!\!\!\includegraphics[width=0.9\textwidth]{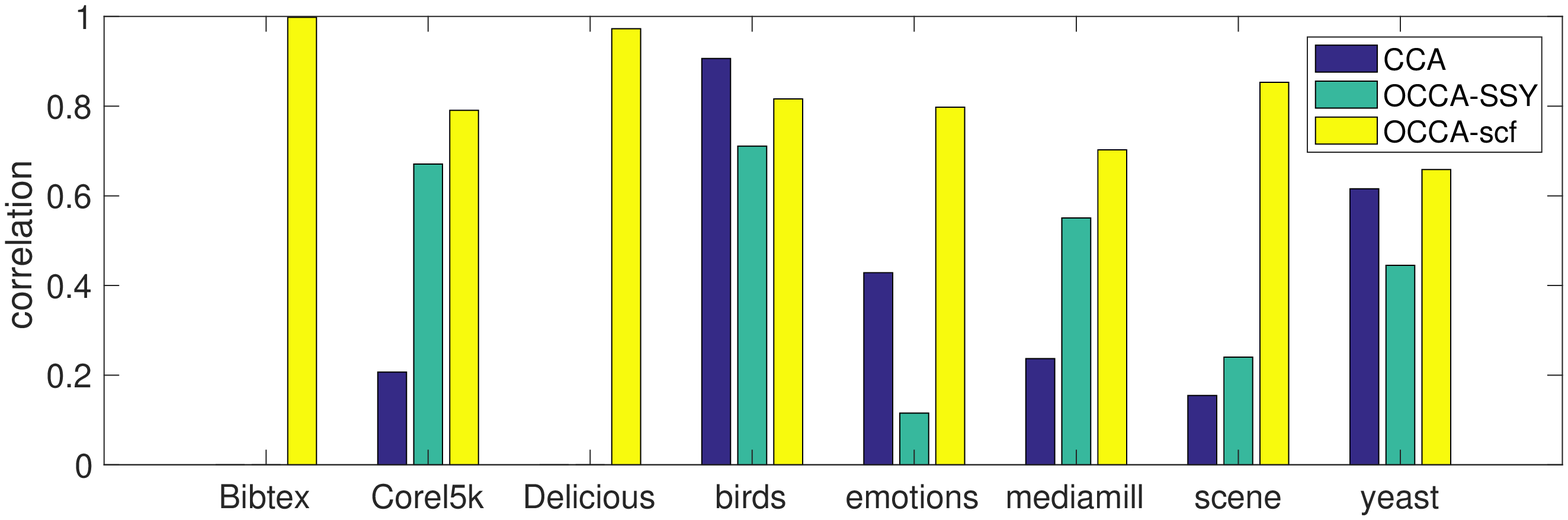} \\
		2-D orthogonal space \\
		\!\!\!\!\!\!\!\!\!\includegraphics[width=0.9\textwidth]{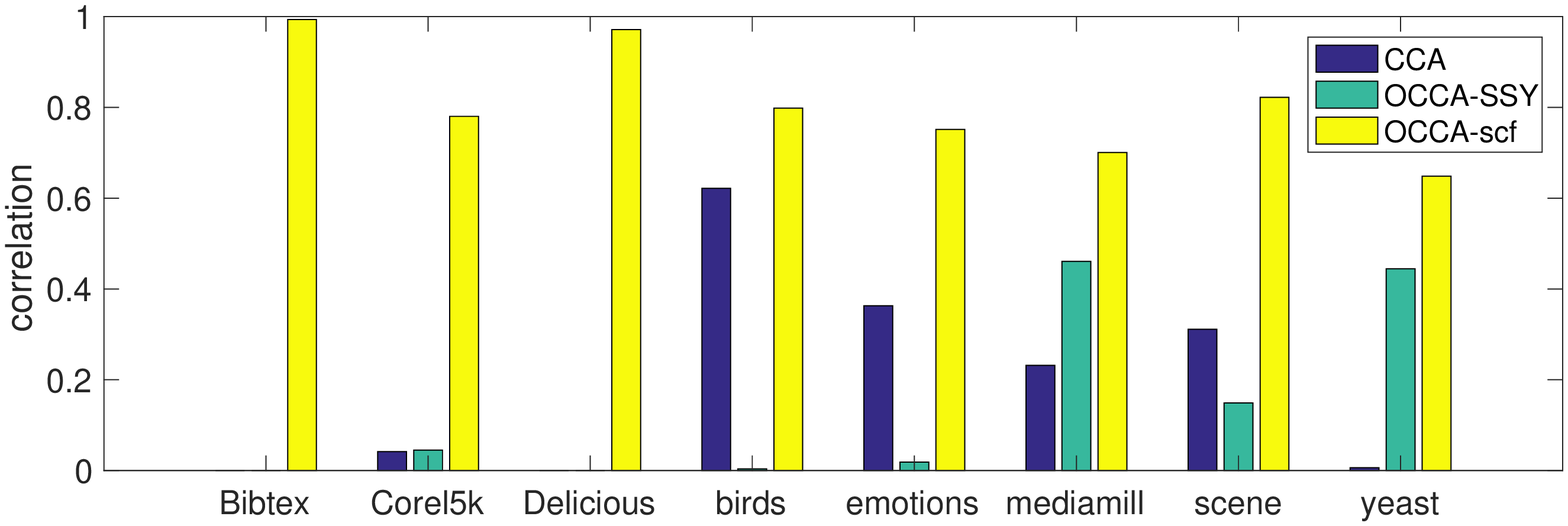}\\
		3-D orthogonal space\\
	\end{tabular}\vspace{-0.1in}
	\caption{The correlations obtained by three CCA methods in the 2-D and 3-D orthogonal spaces.} \label{fig:orth-corr}
\end{figure}

\vspace{-0.1in}
\subsection{Correlation Analysis and Data Visualization}
In this section, we first explore the embedded subspaces obtained by three different CCA methods:
classical CCA, OCCA-SSY and OCCA-scf. For data visualization, the orthogonal space in 2-D and 3-D are the main focus.
As aforementioned, the classical CCA method does not generate the orthonormal basis matrices for projection,
and OCCA-SSY method also does not guarantee to generate the orthonormal basis matrices either.
To obtain orthonormal basis matrices, we post-orthogonalize the matrices obtained by CCA and OCCA-SSY.
If the matrices is rank deficient, we set the correlation as zero since the number of orthogonal basis vectors is smaller
than requested. Fig. \ref{fig:orth-corr} shows the comparisons of three CCA methods in terms of the correlation score
over eight real data in Table \ref{tab:mlc} for multi-label classification problems (detailed description
is presented in Section \ref{sec:mlc}). It is obvious that our proposed OCCA-scf achieves the best performance among all.
More importantly, our method never encounters the matrix rank deficient issue,
while this happens for CCA and OCCA-SSY on some datasets, such as Bibtex and Delicious.

\begin{figure}[t]
	\begin{tabular}{@{}c@{}c@{}c@{}}
		\hline
		& 2-D & 3-D \\
		\hline
		\rotatebox[origin=c]{90}{CCA}&
		\raisebox{-.5\height}{\includegraphics[width=0.5\textwidth]{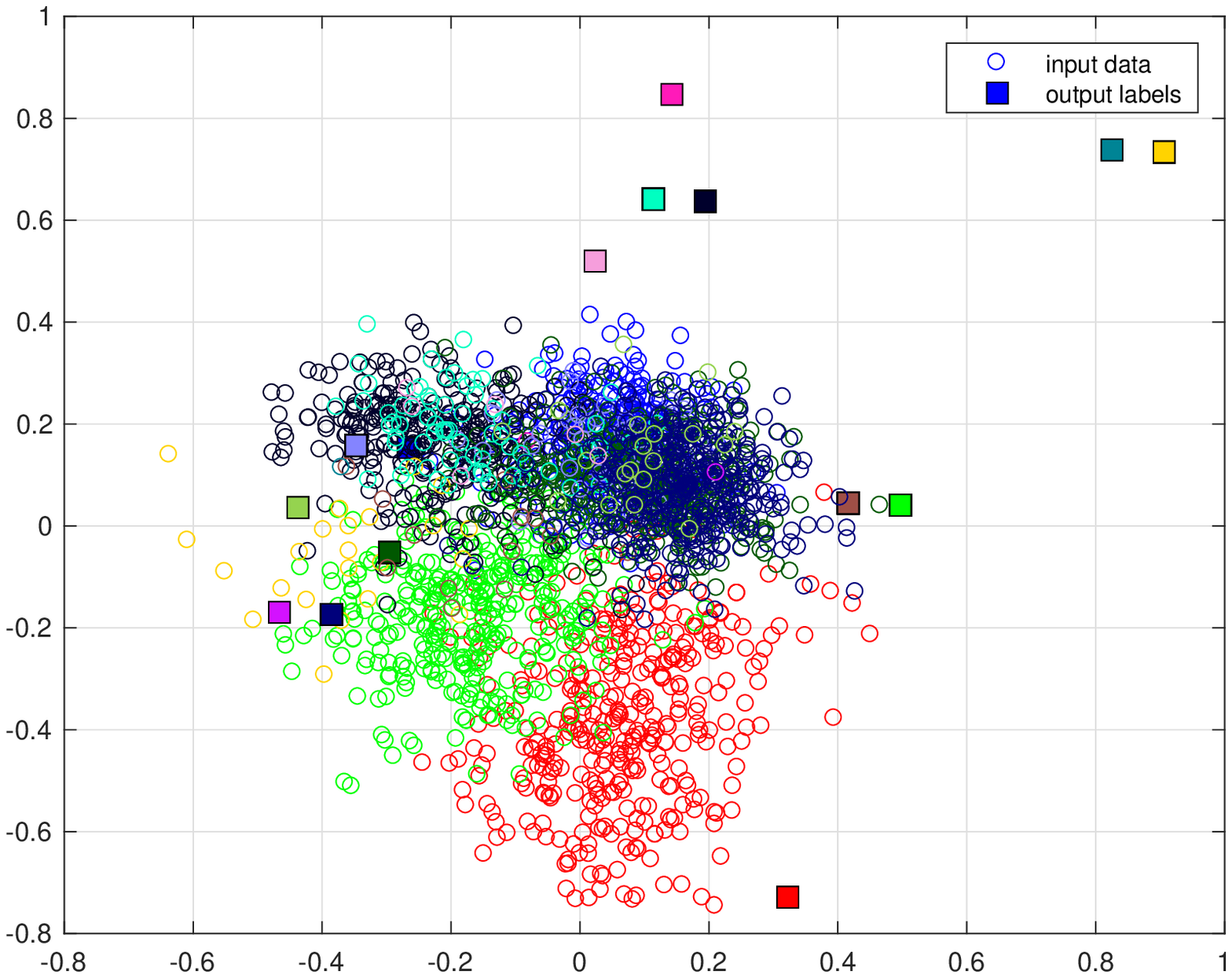}} & \raisebox{-.5\height}{\includegraphics[width=0.5\textwidth]{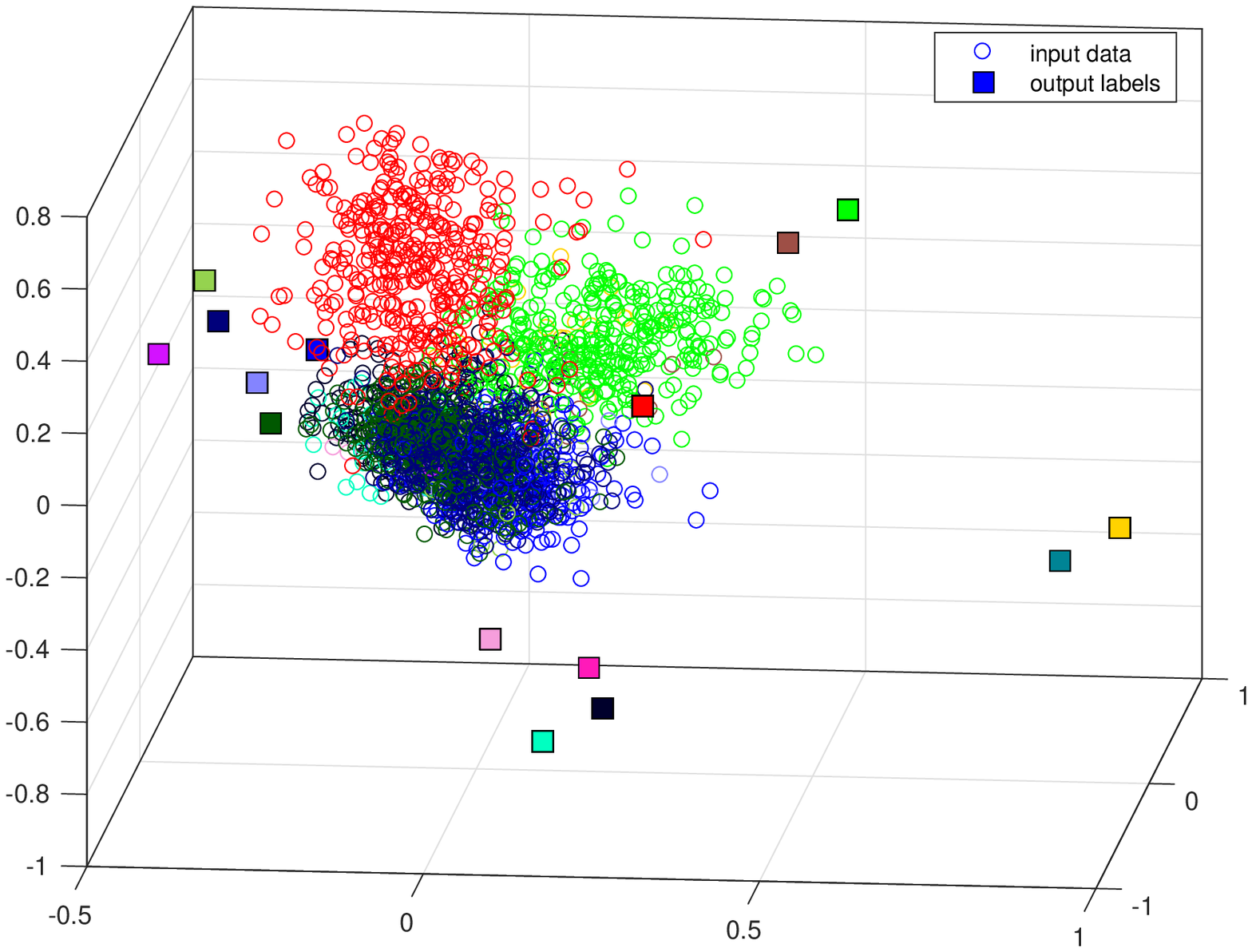}}\\
		\rotatebox[origin=c]{90}{OCCA-SSY}&
		\raisebox{-.5\height}{\includegraphics[width=0.5\textwidth]{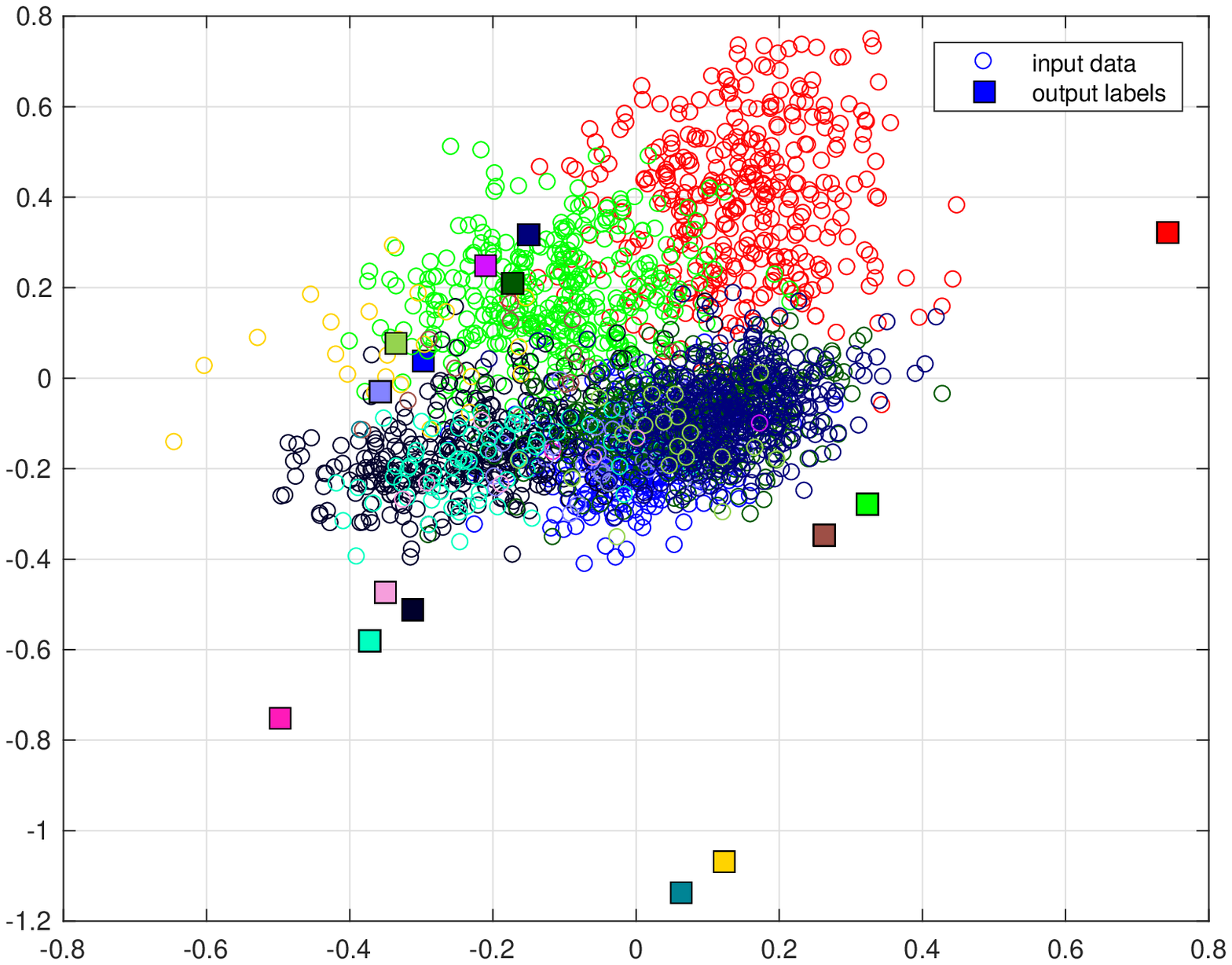}} & \raisebox{-.5\height}{\includegraphics[width=0.5\textwidth]{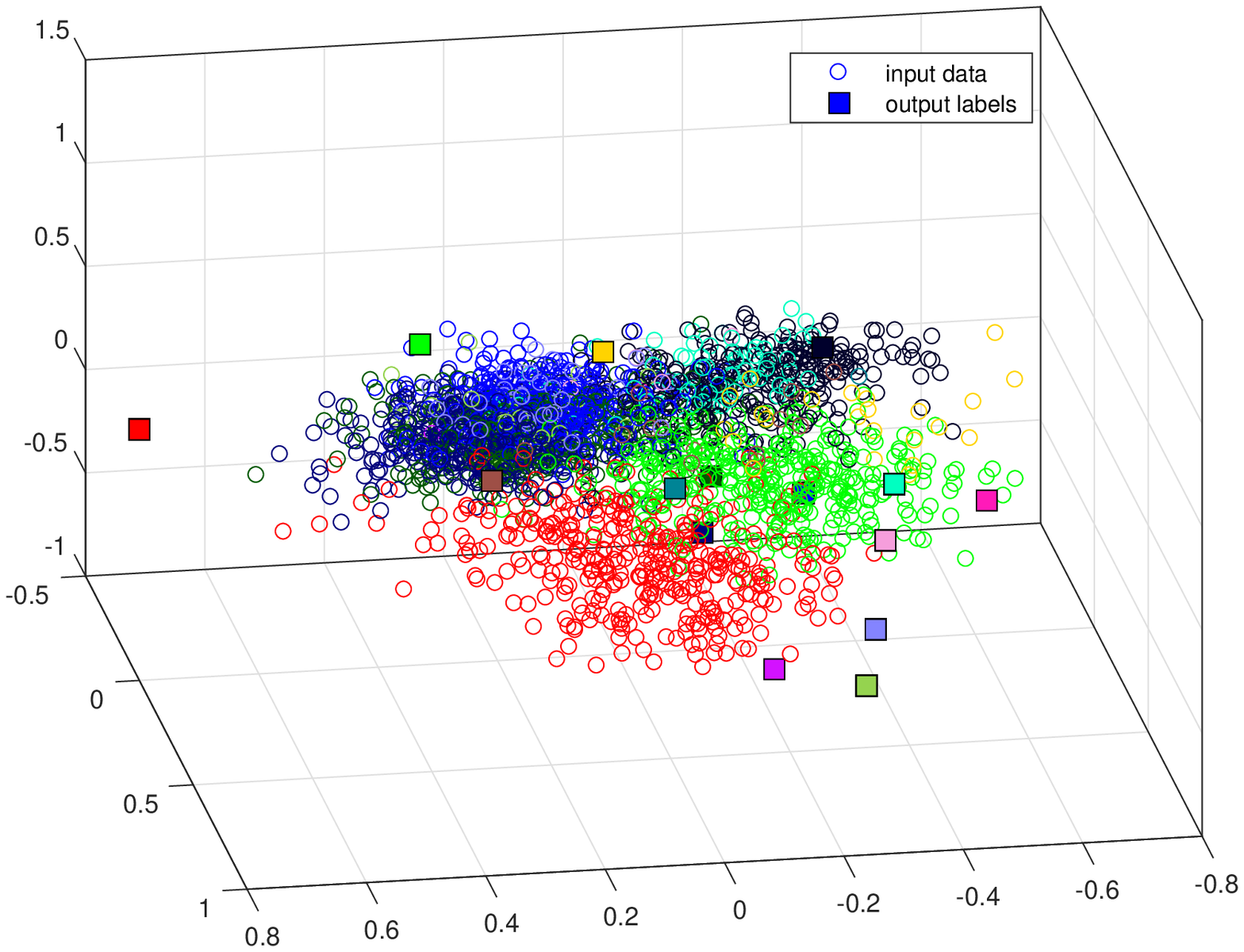}}\\
		\rotatebox[origin=c]{90}{OCCA-scf}&
		\raisebox{-.5\height}{\includegraphics[width=0.5\textwidth]{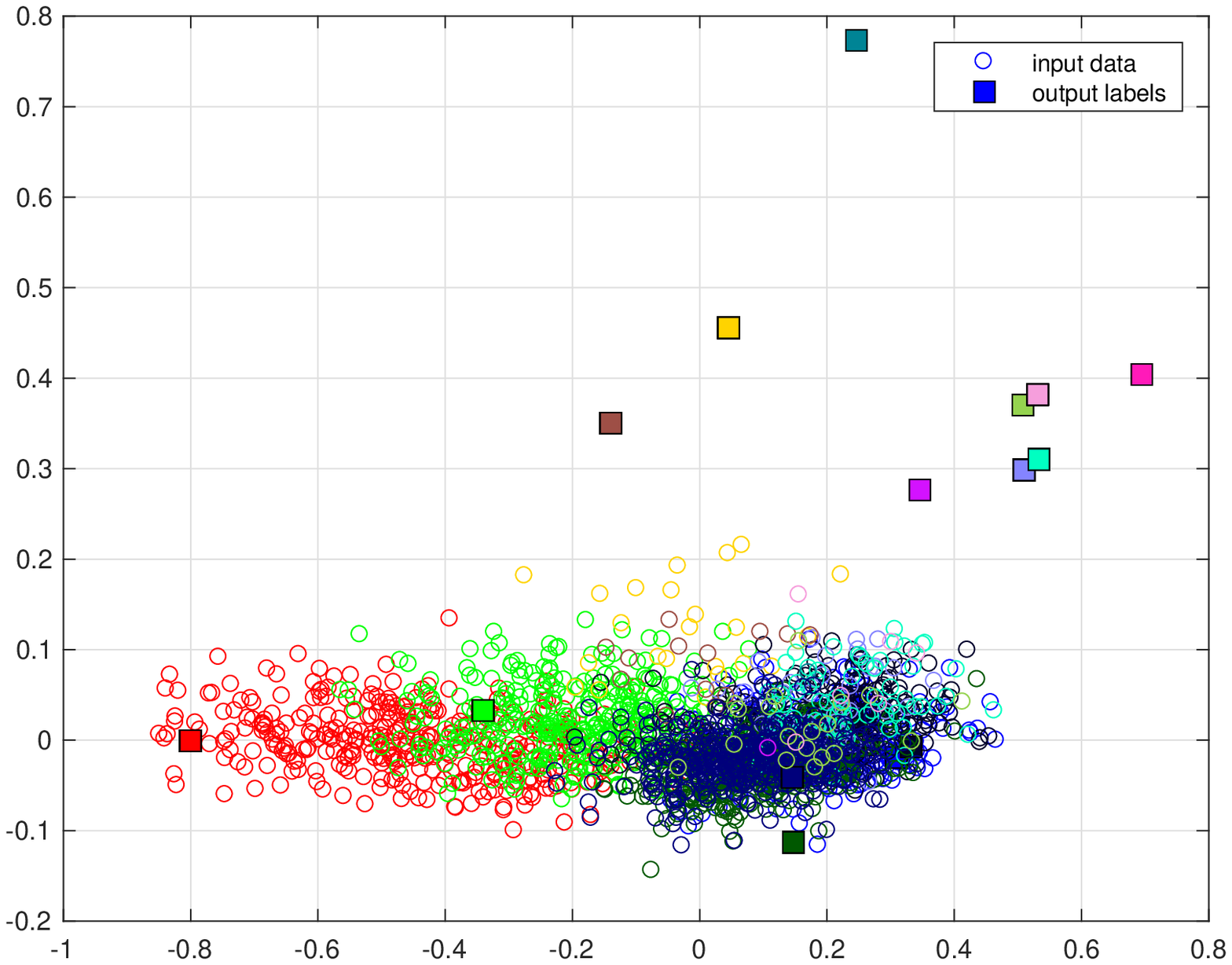}} &
		\raisebox{-.5\height}{\includegraphics[width=0.5\textwidth]{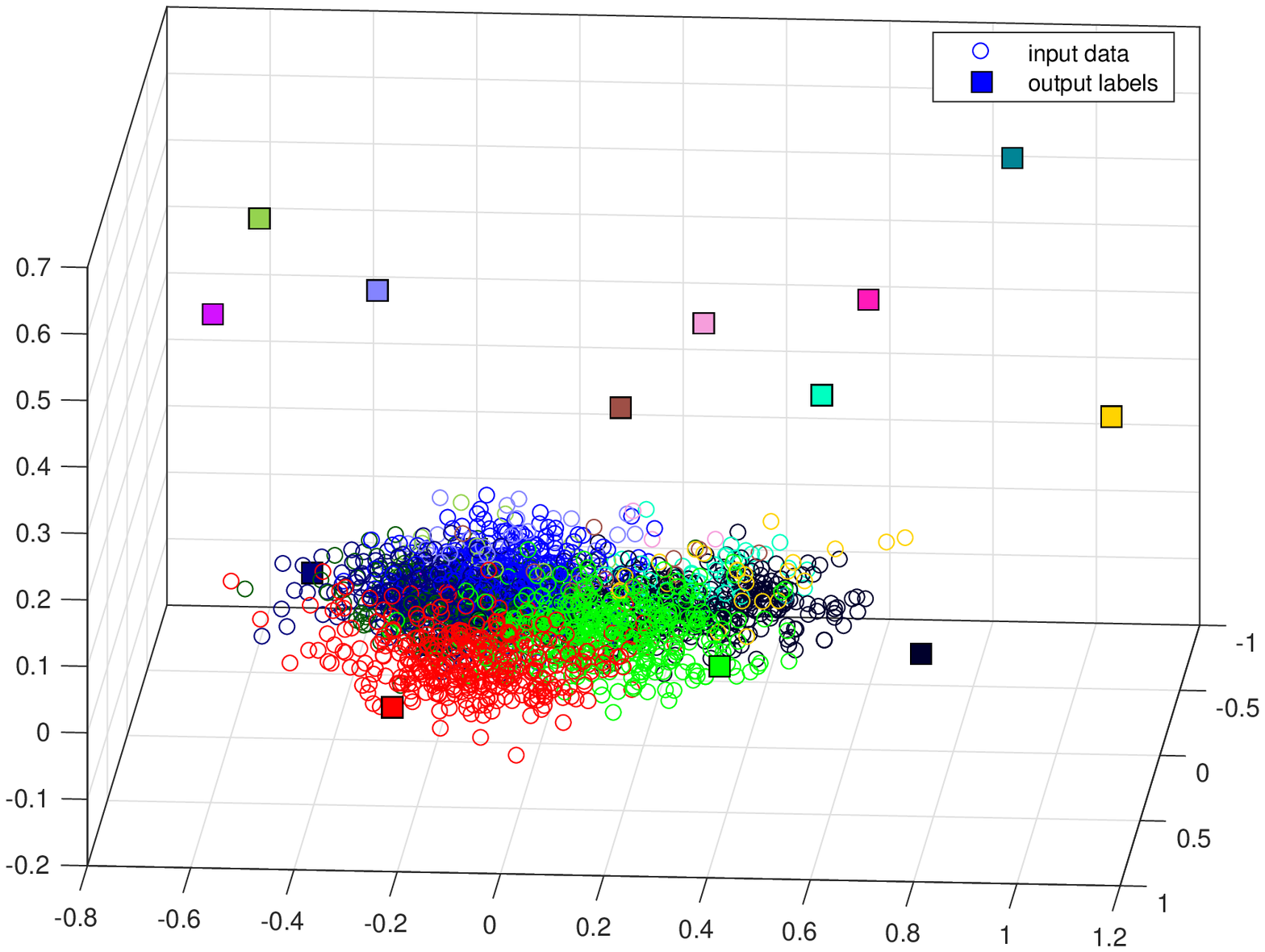}}
	\end{tabular}\vspace{-0.1in}
	\caption{The comparisons of three OCCA methods in terms of 2-D and 3-D embeddings evaluated on scene data. Colors represents classes. The markers circle and square represent the input data points and the output classes. We show one class for each input data and output label although there are $15$ classes extracted from $6$ multiple labels.} \label{fig:viz}
\end{figure}

We then explore the embeddings in 2-D and 3-D spaces and examine the correlations between the input data and its multi-label outputs.
Since each sample may have multiple labels, we transform the output labels into the multi-class classification problem using
the label powerset approach \cite{boutell2004learning} for the purpose of data visualization. The set of multiclass labels
consists of all unique label combinations found in the data. For example, the data scene has 6 labels, and there are $15$ unique label combinations
in total. Fig. \ref{fig:viz} shows the embeddings of both input and output in 2-D and 3-D spaces colored by unique classes.
Since multiple data points are assigned to the same unique class, there are only $15$ embedded output points.
In the cases of both 2-D and 3-D, our OCCA-scf method show the best alignments between input data and output labels,
for example, the majority of classes such as the red, green and blue ones show the better alignment in the reduced space with the input
data clouds than other two methods. In addition to correlation analysis and data visualization, we will evaluate the multi-label
classification problems in terms of various popularly used measurements in Section \ref{sec:mlc}.

\vspace{-0.1in}
\subsection{Applications}
We evaluate our proposed methods in terms of two real-world applications, multi-label classification and multi-view feature extraction,
where various CCA
methods have been explored in the literature \cite{sun2010canonical,shen2015orthogonal}. Since multi-label classification
inherently corresponds to the two-view problem, we evaluate the proposed OCCA-scf method in Algorithm \ref{alg:AltOCCA} against
baselines. For multi-view feature extraction, the proposed RCOMCCA in Algorithm \ref{alg:RCMOCCA} is evaluated over more than two views.

\vspace{-0.1in}
\subsubsection{Multi-label classification} \label{sec:mlc}

\begin{table}
	\caption{Multi-label classification datasets} \label{tab:mlc}
	\centering
	\begin{tabular}{cccc}
		\hline
		Dataset & Samples & Attributes & labels \\
		\hline
		birds & 645 & 260 & 19 \\
		Corel5k & 5000 & 499 & 374 \\
		emotions & 593 & 72 & 6 \\
		scene & 2407 & 294 & 6\\
		yeast & 2417 & 101 & 14 \\
		Bibtex & 7395 & 1836 & 159\\
		Mediamill & 43903 & 120 & 101 \\
		Delicious & 16105 & 500 & 983 \\
		\hline
	\end{tabular}
\end{table}

Multi-label classification \cite{tsoumakas2007multi} is a variant of the classification problem, where one instance
may have various number of labels from a set of predefined categories, i.e., a subset of labels.
It is different from multi-class classification problem, where each instance only has a single label.
In general, the output class labels of one instance are represented by the indicator vector of size $m$ where $m$
is the number of class labels. If the $c$th label is assigned to the instance, the $c$th element of the indicator vector is $1$,
and otherwise $0$. Let $X \in \bbR^{n \times q}$ be the $q$ instances of size $n$ and $Y \in \bbR^{m \times q}$
be comprised of the $q$ indicator vectors. The popular use of CCA for multi-label classification is to treat $X$ to
be one view and $Y$ as the other view \cite{sun2010canonical,zhang2011multi,rai2009multi}.

The multi-label classification datasets used in our experiments are shown in Table \ref{tab:mlc}.
All datasets are publicly available\footnote{http://mulan.sourceforge.net/datasets-mlc.html}.
Following \cite{sun2010canonical}, we take CCA as a supervised dimensionality reduction step for multi-label classification
so that the embeddings obtained by CCA methods can encode certain correlations among input data and labels.
Hence, it expects to have better performance for multi-label classification. Since some datasets have a small
number of output labels, the reduced dimension is up-bounded by the number of output labels due to the inherent property of CCA.
To alleviate the limitation from CCA and improve the classification performance for general datasets,
we propose to augment the learned embeddings using the original input data through a simple concatenation over two sets of features.

In this paper, we choose to use ML-kNN\footnote{http://lamda.nju.edu.cn/files/MLkNN.rar}
as our backend multi-label classifier \cite{zhang2007ml}, which has demonstrated good performance over various datasets.
We compare our OCCA-scf with other CCA methods including OCCA-SSY \cite{shen2013orthogonal}, LS-CCA \cite{sun2010canonical}
and classical CCA. All CCA-based methods take ML-kNN as the base classifier and corresponding augmentation approach for each CCA method
is indicated by the suffix ``-aug''. We randomly split the data into 40\% for training and 60\% for testing and
tune the neighborhood parameter within the set $\{ 1, 3, 5, 7, 9, 13, 15\}$ for ML-kNN.
Following \cite{zhang2007ml}, we report the best results and their standard deviations over $10$ random train/test splits
in terms of the following five measurements:
\begin{itemize}
	\item HammingLoss: the average number of times an instance-label pair is misclassified.
	
	\item RankingLoss: the average fraction of label pairs that are reversely ordered for the instance.
	
	\item OneError: the average number of times the top-ranked label is not in the set of proper labels of the instance
	
	\item Coverage: on the average how far we need to go down the list of labels in order to cover all the proper labels of the instance.
	
	\item Average\_Precision: the average precision of labels ranked above a particular label in the same label set.
\end{itemize}
Except Average\_Precision, the first four measurements show good performances of multi-label classification
if the measurement value is small.

\begin{table*}
	\centering	
	\caption{Results on five datasets in terms of the $5$ measurements(40\% for training and $60\%$ for testing over $10$ random splits). Best results are in bold.} \label{tab:mlc-results}
	\begin{tiny}
	\begin{tabular}{c|c|ccccc}
		\hline
		dataset & method & HammingLoss & RankingLoss & OneError & Coverage & Average\_Precision\\
		\hline
		\multirow{9}{*}{birds} &OCCA-scf& \textbf{0.0503 $\pm$ 0.0035}& \textbf{0.2173 $\pm$ 0.0062}& \textbf{0.4964 $\pm$ 0.0201}& \textbf{2.8866 $\pm$ 0.1580}& \textbf{0.5452 $\pm$ 0.0118}\\
		&	OCCA-scf-aug& 0.0545 $\pm$ 0.0026& 0.3045 $\pm$ 0.0047& 0.7101 $\pm$ 0.0136& 3.8597 $\pm$ 0.1754& 0.3942 $\pm$ 0.0107\\
		&	CCA& 0.1167 $\pm$ 0.0095& 0.3509 $\pm$ 0.0197& 0.8110 $\pm$ 0.0302& 4.2028 $\pm$ 0.2954& 0.3087 $\pm$ 0.0192\\
		&	CCA-aug& 0.0545 $\pm$ 0.0026& 0.3046 $\pm$ 0.0046& 0.7101 $\pm$ 0.0136& 3.8602 $\pm$ 0.1745& 0.3942 $\pm$ 0.0107\\
		&	LS-CCA& 0.1167 $\pm$ 0.0095& 0.3509 $\pm$ 0.0197& 0.8110 $\pm$ 0.0302& 4.2028 $\pm$ 0.2954& 0.3084 $\pm$ 0.0191\\
		&	LS-CCA-aug& 0.0545 $\pm$ 0.0026& 0.3046 $\pm$ 0.0046& 0.7101 $\pm$ 0.0136& 3.8602 $\pm$ 0.1745& 0.3942 $\pm$ 0.0107\\
		&	OCCA-SSY& 0.0618 $\pm$ 0.0049& 0.2669 $\pm$ 0.0150& 0.5978 $\pm$ 0.0269& 3.4499 $\pm$ 0.2146& 0.4722 $\pm$ 0.0182\\
		&	OCCA-SSY-aug& 0.0545 $\pm$ 0.0026& 0.3046 $\pm$ 0.0046& 0.7101 $\pm$ 0.0136& 3.8607 $\pm$ 0.1752& 0.3942 $\pm$ 0.0108\\
		&	ML-kNN& 0.0545 $\pm$ 0.0026& 0.3046 $\pm$ 0.0046& 0.7101 $\pm$ 0.0136& 3.8607 $\pm$ 0.1752& 0.3942 $\pm$ 0.0108\\
		\hline
		\multirow{9}{*}{emotions} 	&OCCA-scf& \textbf{0.2283 $\pm$ 0.0064}& \textbf{0.2016 $\pm$ 0.0091}& \textbf{0.3258 $\pm$ 0.0201}& \textbf{1.9643 $\pm$ 0.0456}& \textbf{0.7640 $\pm$ 0.0118}\\
		&OCCA-scf-aug& 0.2716 $\pm$ 0.0057& 0.2799 $\pm$ 0.0098& 0.3989 $\pm$ 0.0171& 2.3862 $\pm$ 0.0573& 0.6959 $\pm$ 0.0085\\
		&CCA& 0.2395 $\pm$ 0.0090& 0.2204 $\pm$ 0.0138& 0.3497 $\pm$ 0.0169& 2.0736 $\pm$ 0.0730& 0.7443 $\pm$ 0.0126\\
		&CCA-aug& 0.2718 $\pm$ 0.0059& 0.2801 $\pm$ 0.0094& 0.3986 $\pm$ 0.0168& 2.3860 $\pm$ 0.0595& 0.6960 $\pm$ 0.0082\\
		&LS-CCA& 0.2346 $\pm$ 0.0084& 0.2088 $\pm$ 0.0149& 0.3385 $\pm$ 0.0182& 2.0096 $\pm$ 0.0930& 0.7553 $\pm$ 0.0154\\
		&LS-CCA-aug& 0.2719 $\pm$ 0.0056& 0.2795 $\pm$ 0.0099& 0.3983 $\pm$ 0.0172& 2.3848 $\pm$ 0.0572& 0.6964 $\pm$ 0.0085\\
		&OCCA-SSY& 0.2577 $\pm$ 0.0141& 0.2543 $\pm$ 0.0198& 0.3860 $\pm$ 0.0274& 2.2309 $\pm$ 0.0916& 0.7190 $\pm$ 0.0172\\
		&OCCA-SSY-aug& 0.2719 $\pm$ 0.0057& 0.2800 $\pm$ 0.0095& 0.3986 $\pm$ 0.0170& 2.3862 $\pm$ 0.0589& 0.6958 $\pm$ 0.0084\\
		&ML-kNN& 0.2720 $\pm$ 0.0057& 0.2798 $\pm$ 0.0097& 0.3983 $\pm$ 0.0169& 2.3862 $\pm$ 0.0589& 0.6960 $\pm$ 0.0085\\
		\hline
		\multirow{9}{*}{Scene} 	&OCCA-scf& 0.1214 $\pm$ 0.0024& 0.1375 $\pm$ 0.0070& 0.3329 $\pm$ 0.0073& 0.7772 $\pm$ 0.0360& 0.7902 $\pm$ 0.0060\\
		&OCCA-scf-aug& \textbf{0.0941 $\pm$ 0.0016}& \textbf{0.0817 $\pm$ 0.0028}& \textbf{0.2428 $\pm$ 0.0081}& \textbf{0.4981 $\pm$ 0.0154}& 0.8557 $\pm$ 0.0041\\
		&CCA& 0.1267 $\pm$ 0.0032& 0.1448 $\pm$ 0.0068& 0.3451 $\pm$ 0.0087& 0.8153 $\pm$ 0.0369& 0.7810 $\pm$ 0.0065\\
		&CCA-aug& 0.0949 $\pm$ 0.0020& 0.0820 $\pm$ 0.0035& 0.2440 $\pm$ 0.0080& 0.4999 $\pm$ 0.0194& 0.8555 $\pm$ 0.0044\\
		&LS-CCA& 0.1228 $\pm$ 0.0028& 0.1401 $\pm$ 0.0059& 0.3361 $\pm$ 0.0085& 0.7909 $\pm$ 0.0326& 0.7873 $\pm$ 0.0058\\
		&LS-CCA-aug& 0.0948 $\pm$ 0.0021& 0.0821 $\pm$ 0.0034& 0.2440 $\pm$ 0.0082& 0.5003 $\pm$ 0.0187& 0.8553 $\pm$ 0.0044\\
		&OCCA-SSY& 0.1183 $\pm$ 0.0030& 0.1302 $\pm$ 0.0055& 0.3226 $\pm$ 0.0086& 0.7405 $\pm$ 0.0314& 0.7979 $\pm$ 0.0063\\
		&OCCA-SSY-aug& 0.0943 $\pm$ 0.0020& 0.0818 $\pm$ 0.0025& 0.2431 $\pm$ 0.0090& 0.4981 $\pm$ 0.0137& \textbf{0.8558 $\pm$ 0.0042}\\
		&ML-kNN& 0.0949 $\pm$ 0.0020& 0.0823 $\pm$ 0.0033& 0.2442 $\pm$ 0.0085& 0.5009 $\pm$ 0.0188& 0.8554 $\pm$ 0.0042\\
		\hline
		\multirow{9}{*}{Corel5k} 	&OCCA-scf& \textbf{0.0094 $\pm$ 0.0000}& \textbf{0.1365 $\pm$ 0.0015}& \textbf{0.7252 $\pm$ 0.0055}& \textbf{116.3179 $\pm$ 1.2727}& \textbf{0.2529 $\pm$ 0.0034}\\
		&OCCA-scf-aug& \textbf{0.0094 $\pm$ 0.0000}& 0.1373 $\pm$ 0.0016& 0.7309 $\pm$ 0.0067& 116.9343 $\pm$ 1.3411& 0.2474 $\pm$ 0.0031\\
		&CCA& \textbf{0.0094 $\pm$ 0.0000}& 0.1396 $\pm$ 0.0012& 0.7519 $\pm$ 0.0079& 117.9406 $\pm$ 1.1541& 0.2339 $\pm$ 0.0035\\
		&CCA-aug& \textbf{0.0094 $\pm$ 0.0000}& 0.1381 $\pm$ 0.0016& 0.7327 $\pm$ 0.0056& 117.3671 $\pm$ 1.2944& 0.2436 $\pm$ 0.0031\\
		&LS-CCA& 0.0095 $\pm$ 0.0000& 0.1379 $\pm$ 0.0015& 0.7432 $\pm$ 0.0082& 116.8526 $\pm$ 1.3332& 0.2463 $\pm$ 0.0027\\
		&LS-CCA-aug& \textbf{0.0094 $\pm$ 0.0000}& 0.1376 $\pm$ 0.0015& 0.7323 $\pm$ 0.0082& 117.1294 $\pm$ 1.2682& 0.2459 $\pm$ 0.0039\\
		&OCCA-SSY& \textbf{0.0094 $\pm$ 0.0000}& \textbf{0.1365 $\pm$ 0.0015}& 0.7263 $\pm$ 0.0085& 116.4133 $\pm$ 1.3187& 0.2522 $\pm$ 0.0038\\
		&OCCA-SSY-aug& \textbf{0.0094 $\pm$ 0.0000}& 0.1371 $\pm$ 0.0016& 0.7304 $\pm$ 0.0054& 116.8367 $\pm$ 1.2580& 0.2481 $\pm$ 0.0032\\
		&ML-kNN& \textbf{0.0094 $\pm$ 0.0000}& 0.1381 $\pm$ 0.0019& 0.7323 $\pm$ 0.0062& 117.5434 $\pm$ 1.4205& 0.2434 $\pm$ 0.0035\\
		\hline
		\multirow{9}{*}{yeast} 	&OCCA-scf& 0.2080 $\pm$ 0.0021& 0.1838 $\pm$ 0.0039& 0.2538 $\pm$ 0.0066& 6.5615 $\pm$ 0.0736& 0.7445 $\pm$ 0.0049\\
		&OCCA-scf-aug& \textbf{0.1997 $\pm$ 0.0033}& \textbf{0.1735 $\pm$ 0.0034}& \textbf{0.2356 $\pm$ 0.0075}& \textbf{6.3870 $\pm$ 0.0859}& \textbf{0.7556 $\pm$ 0.0041}\\
		&CCA& 0.2108 $\pm$ 0.0031& 0.1894 $\pm$ 0.0056& 0.2539 $\pm$ 0.0069& 6.6438 $\pm$ 0.0969& 0.7364 $\pm$ 0.0054\\
		&CCA-aug& 0.2011 $\pm$ 0.0026& 0.1762 $\pm$ 0.0035& 0.2398 $\pm$ 0.0068& 6.4152 $\pm$ 0.0827& 0.7512 $\pm$ 0.0040\\
		&LS-CCA& 0.2077 $\pm$ 0.0038& 0.1855 $\pm$ 0.0044& 0.2518 $\pm$ 0.0092& 6.5928 $\pm$ 0.1040& 0.7436 $\pm$ 0.0054\\
		&LS-CCA-aug& 0.2012 $\pm$ 0.0028& 0.1760 $\pm$ 0.0036& 0.2405 $\pm$ 0.0060& 6.4096 $\pm$ 0.0864& 0.7511 $\pm$ 0.0038\\
		&OCCA-SSY& 0.2097 $\pm$ 0.0035& 0.1871 $\pm$ 0.0052& 0.2526 $\pm$ 0.0058& 6.6265 $\pm$ 0.1023& 0.7403 $\pm$ 0.0055\\
		&OCCA-SSY-aug& 0.1997 $\pm$ 0.0033& 0.1740 $\pm$ 0.0033& 0.2356 $\pm$ 0.0074& 6.3907 $\pm$ 0.0984& 0.7554 $\pm$ 0.0036\\
		&ML-kNN& 0.2017 $\pm$ 0.0029& 0.1759 $\pm$ 0.0036& 0.2397 $\pm$ 0.0067& 6.4075 $\pm$ 0.0831& 0.7512 $\pm$ 0.0036\\
		\hline
	\end{tabular}
\end{tiny}
\end{table*}

Table \ref{tab:mlc-results} shows the results obtained by compared methods over $5$ datasets in terms of the $5$ measurements.
We observed that our OCCA-scf and OCCA-scf-aug show better results on almost all the five measurements
except Average\_Precision on Scene by OCCA-SSY-aug. For datasets scene and yeast, OCCA-scf-aug shows better results than OCCA-scf.
This implies that our augmentation approach is effective when the features obtained by dimensionality reduction method
such as CCA somehow lose the information that is also useful for multi-label classification although the correlations  remain.
It is worth noting that our methods outperform ML-kNN over all the experimented datasets.
These observations imply that CCA with orthogonality constraints improves ML-kNN for multi-label classification
and our proposed OCCA-scf methods outperform other CCA methods.

\begin{table}[h]
	\caption{Multi-view datasets} \label{tab:mv}
	\centering
	\begin{tabular}{cccc}
		\hline
		Dataset & Samples & Multiple views & classes \\
		\hline
		mfeat & 2000 & 216;76;64;6;240;47 & 10 \\
		Caltech101-7 & 1474 & 254;512;1180;1008;64;1000 & 7 \\
		Caltech101-20 & 2386 & 254;512;1180;1008;64;1000 & 20 \\
		Scene15 & 4310 & 254;512;531;360;64;1000 & 15\\
		yeast\_ribosomal & 1040 & 3735;4901;441 & 2 \\
		\hline
	\end{tabular}
\end{table}

\subsubsection{Multi-view feature extraction}
Previous experiments focus on the problems with only two views. Here, we aim to explore our proposed RCOMCCA for multiset
canonical correlation analysis in terms of multi-view feature extraction \cite{shen2015orthogonal,shen2013orthogonal}.
Following \cite{shen2013orthogonal}, we employ the serial fusion strategy to concatenate embeddings
from all views for classification based on 1-nearest neighbor classifier. Since LDR-based CCA and LS-CCA are not
easy to be extended for learning with multiple views, we compare our proposed RCOMCCA with
MCCA \cite{kettenring1971canonical,nielsen2002multiset} and OMCCA-SS \cite{shen2015orthogonal}.
It is worth noting that RCOMCCA allows two updating scheme and is capable of integrating various weighting scheme. Hence, we name the variants of RCOMCCA by augmenting ``-G'' for Gauss-Seidel-style and ``-J'' for the Jacobi-style,
together with three weighting schemes shown in bracket. As a result, there are six variants of RCOMCCA in total.
For the top-$p$ weighting scheme, $p \in \{1, 3, 6\}$ is used, except that $p \in \{1,3\}$ is used for yeast\_ribosomal.

The datasets with relevant statistics used for multi-view feature extraction are shown in Table \ref{tab:mv}.
For image datasets such as Caltech101\footnote{http://www.vision.caltech.edu/Image\_Datasets/Caltech101/}\cite{fei2007learning}
and Scene15\footnote{https://figshare.com/articles/15-Scene\_Image\_Dataset/7007177} \cite{lazebnik2006beyond},
we apply various feature descriptors to extract features of views including CENTRIST \cite{wu2008place}, GIST \cite{oliva2001modeling},
LBP \cite{ojala2002multiresolution}, histogram of oriented gradient (HOG), color histogram (CH),
and SIFT-SPM \cite{lazebnik2006beyond}. Note that we drop  CH for Scene15 due to the gray-level images.
mfeat is handwritten numeral
data\footnote{https://archive.ics.uci.edu/ml/datasets/Multiple+Features}   \cite{Dua:2019} with $6$ views
including 76 Fourier coefficients of the character shapes, 216 profile correlations, 64 Karhunen-Love coefficients,
240 pixel averages in $2 \times 3$ windows, 47 Zernike moments, and 6 morphological features.
The Berkeley genomic dataset yeast\_ribosomal
\footnote{https://noble.gs.washington.edu/proj/sdp-svm/} is used where three aspects of the
protein are considered as the views including Pfam HMM, Hydrophobicity FFT and Gene expression for binary classification problem,
e.g., ribosomal vs. non-ribosomal.

\begin{landscape}
\begin{table}[htbp]
	\caption{Means and standard deviations of accuracy obtained by 1-nearest neighbor classifier on each view
		and embeddings obtained by three CCA methods over $10$ random drawn from each dataset (30\% training and 70\% testing).
		Parameter $k$ used by CCA methods to achieve the best accuracy is shown in the bracket. The symbol ``-'' for the non-existence of the view.}
	\centering \vspace{-0.15in}
\begin{scriptsize}	
	\begin{tabular}{lccccc}\hline
		& mfeat& Caltech101-7& Caltech101-20& Scene15& yeast\_ribosomal\\\hline
		view1 & 0.9513 $\pm$ 0.0053 & 0.9259 $\pm$ 0.0049 & 0.7659 $\pm$ 0.0046 & 0.5766 $\pm$ 0.0091 & 0.8553 $\pm$ 0.0472\\
		view2 & 0.7604 $\pm$ 0.0104 & 0.9443 $\pm$ 0.0051 & 0.8257 $\pm$ 0.0064 & 0.5269 $\pm$ 0.0070 & 0.8831 $\pm$ 0.0072\\
		view3 & 0.9293 $\pm$ 0.0043 & 0.9415 $\pm$ 0.0070 & 0.8226 $\pm$ 0.0106 & 0.5528 $\pm$ 0.0081 & \textbf{0.9856 $\pm$ 0.0046}\\
		view4 & 0.6780 $\pm$ 0.0064 & 0.9287 $\pm$ 0.0105 & 0.7968 $\pm$ 0.0118 & 0.4609 $\pm$ 0.0079 & -\\
		view5 & 0.9630 $\pm$ 0.0025 & 0.7759 $\pm$ 0.0133 & 0.6042 $\pm$ 0.0122 & 0.6946 $\pm$ 0.0130 & -\\
		view6 & 0.7814 $\pm$ 0.0077 & 0.9152 $\pm$ 0.0059 & 0.7645 $\pm$ 0.0128 & - & -\\\hline
		MCCA & 0.8679 $\pm$ 0.0073 (6) & 0.8865 $\pm$ 0.0072 (15) & 0.8620 $\pm$ 0.0072 (40) & 0.6851 $\pm$ 0.0043 (35) & 0.8155 $\pm$ 0.0139 (3)\\
		OMCCA-SS & 0.8298 $\pm$ 0.0089 (6) & 0.9493 $\pm$ 0.0024 (45) & 0.8527 $\pm$ 0.0057 (50) & 0.7030 $\pm$ 0.0081 (50) & 0.8379 $\pm$ 0.0110 (5)\\
		RCOMCCA-G (uniform) & 0.7634 $\pm$ 0.0134 (5) & 0.8880 $\pm$ 0.0052 (50) & 0.7150 $\pm$ 0.0075 (45) & 0.4866 $\pm$ 0.0044 (50) & 0.8639 $\pm$ 0.0291 (40)\\
		RCOMCCA-G (top-$p$) & \textbf{0.9696 $\pm$ 0.0035} (5) & \textbf{0.9664 $\pm$ 0.0060} (35) & 0.8887 $\pm$ 0.0077 (25) & 0.7542 $\pm$ 0.0054 (30) & 0.8756 $\pm$ 0.0095 (45)\\
		RCOMCCA-G (tree) & 0.9566 $\pm$ 0.0031 (6) & 0.9392 $\pm$ 0.0043 (45) & 0.7882 $\pm$ 0.0078 (50) & 0.4004 $\pm$ 0.0063 (30) & 0.8678 $\pm$ 0.0161 (45)\\
		RCOMCCA-J (uniform) & 0.7540 $\pm$ 0.0121 (5) & 0.8868 $\pm$ 0.0068 (30) & 0.7350 $\pm$ 0.0091 (50) & 0.4995 $\pm$ 0.0059 (50) & 0.8492 $\pm$ 0.0201 (35)\\
		RCOMCCA-J (top-$p$) & 0.9692 $\pm$ 0.0038 (5) & 0.9649 $\pm$ 0.0029 (15) & \textbf{0.8893 $\pm$ 0.0074} (25) & \textbf{0.7574 $\pm$ 0.0077} (30) & 0.8782 $\pm$ 0.0071 (35)\\
		RCOMCCA-J (tree) & 0.9581 $\pm$ 0.0055 (6) & 0.9474 $\pm$ 0.0041 (45) & 0.7799 $\pm$ 0.0084 (50) & 0.4188 $\pm$ 0.0123 (35) & 0.8678 $\pm$ 0.0099 (25)\\
		
		\hline
	\end{tabular}
\end{scriptsize}
\end{table}
\end{landscape}

\begin{figure*}[t]
	\centering
	\begin{tabular}{@{}c@{}c@{}c@{}c@{}c@{}}
		\hline
		& Caltech101-7 & Caltech101-20 & Scene15 & yeast\_ribosomal \\\hline
		\rotatebox[origin=c]{90}{Accuracy}& \raisebox{-.5\height}{\includegraphics[width=0.235\textwidth]{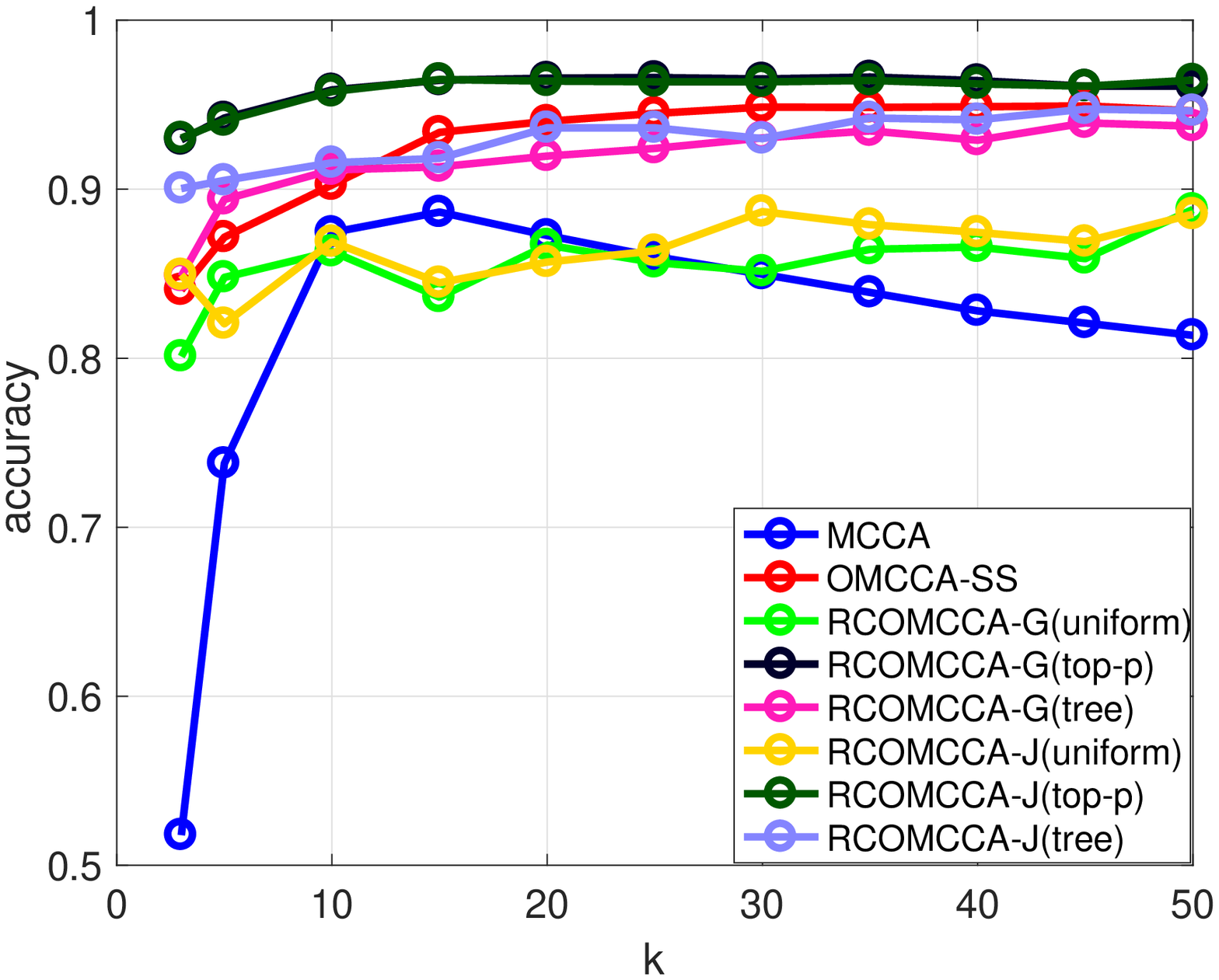}} &
		\raisebox{-.5\height}{\includegraphics[width=0.235\textwidth]{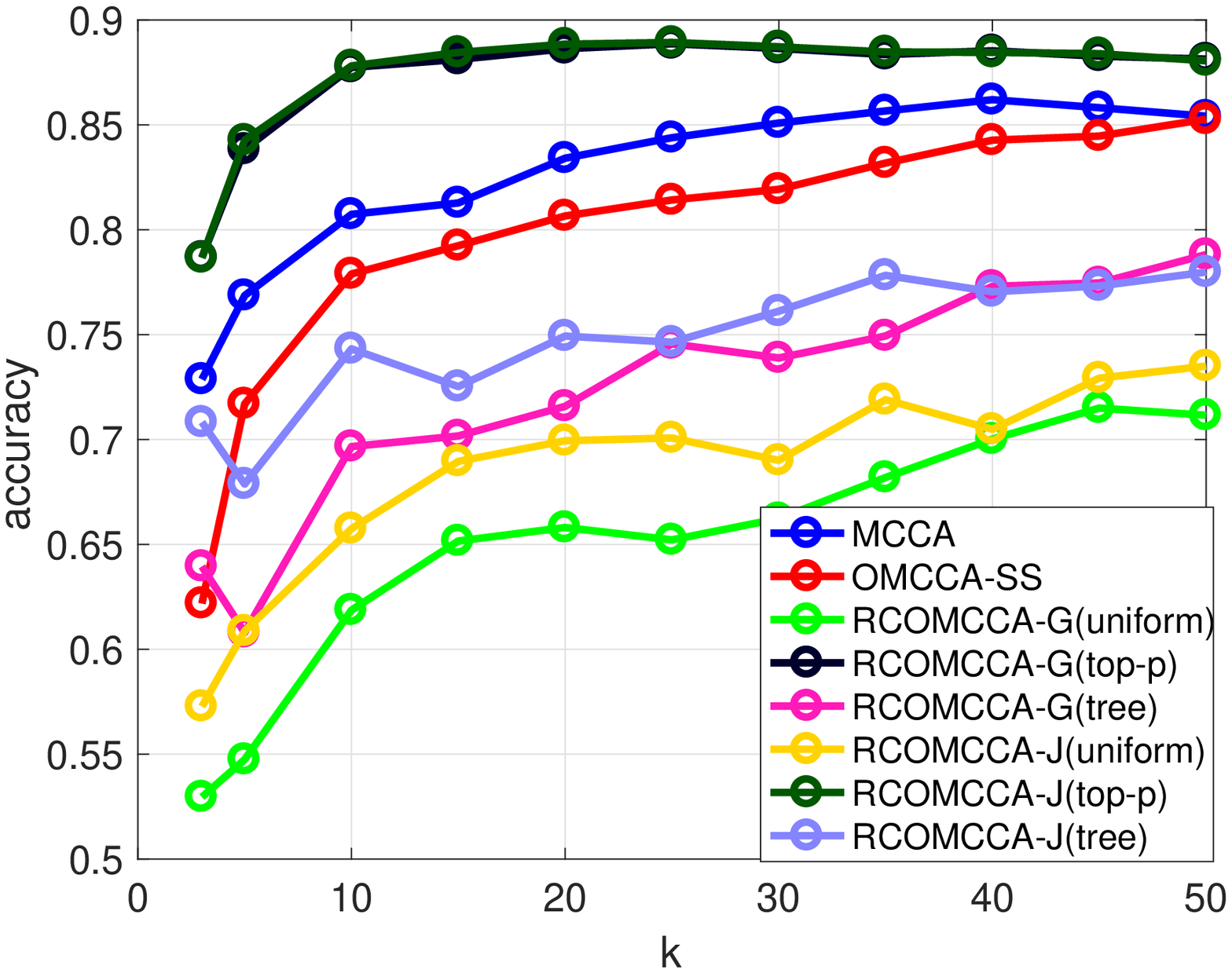}} &
		\raisebox{-.5\height}{\includegraphics[width=0.235\textwidth]{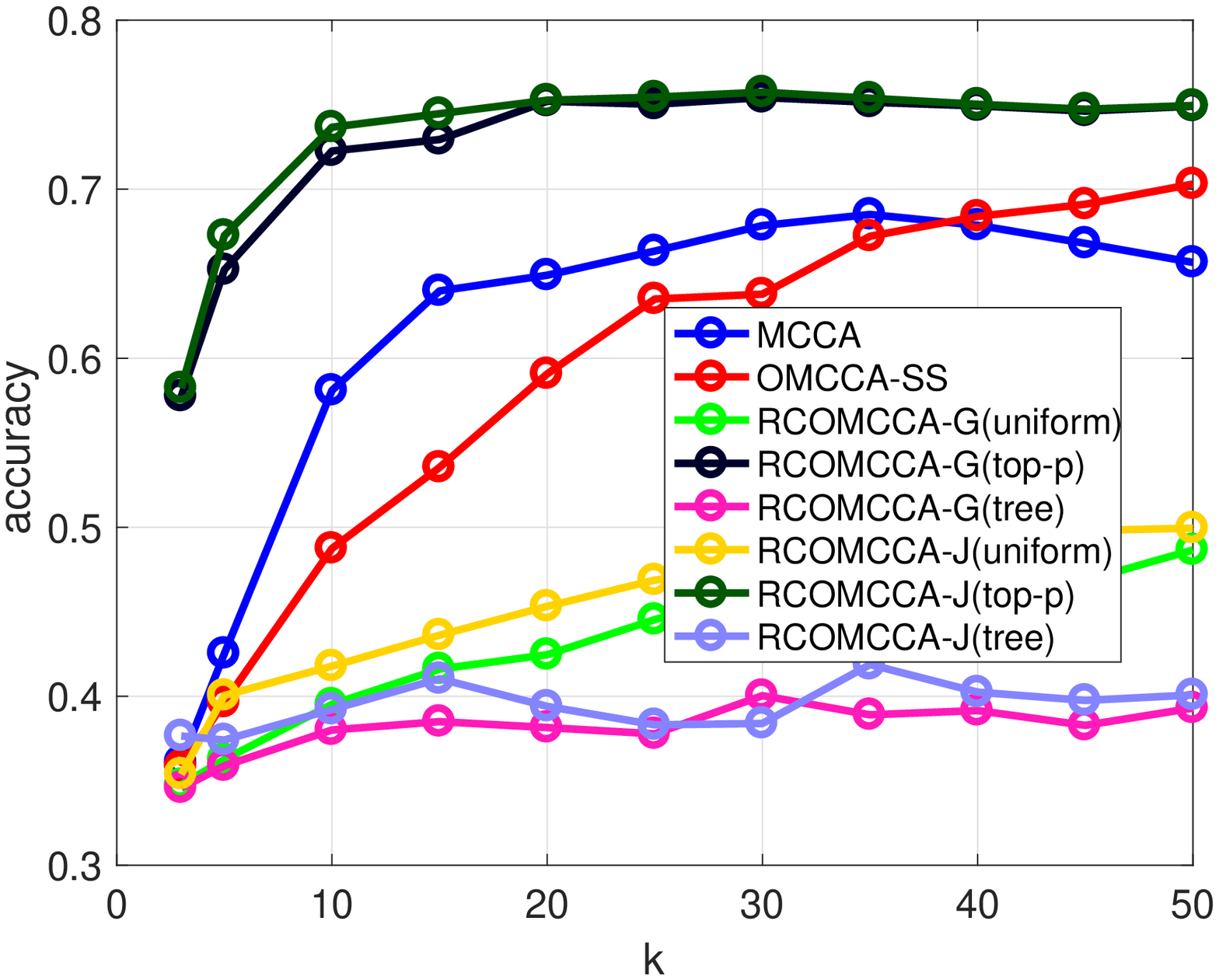}} &
		\raisebox{-.5\height}{\includegraphics[width=0.235\textwidth]{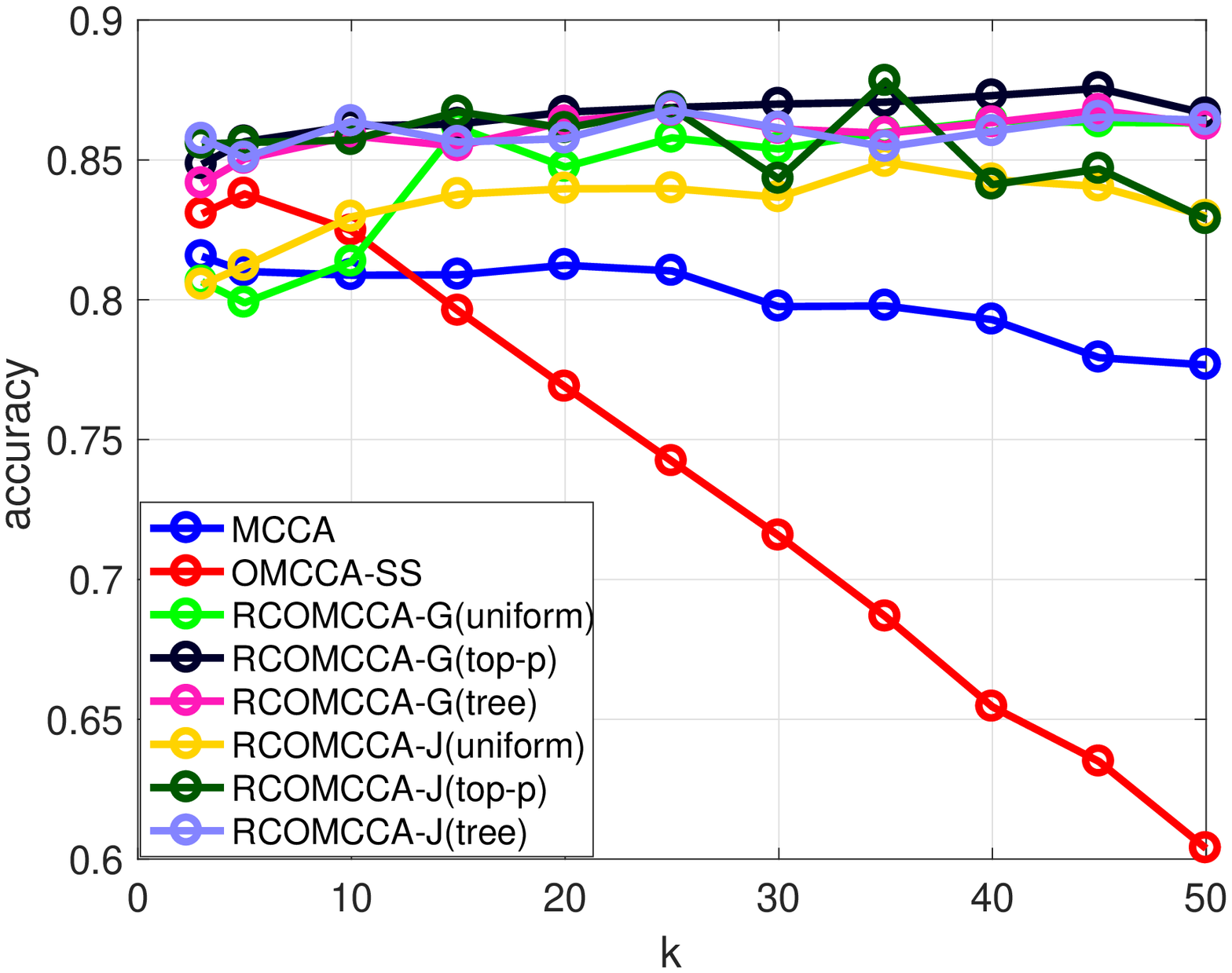}}
		\\
		\rotatebox[origin=c]{90}{CPU time}&
		\raisebox{-.5\height}{\includegraphics[width=0.235\textwidth]{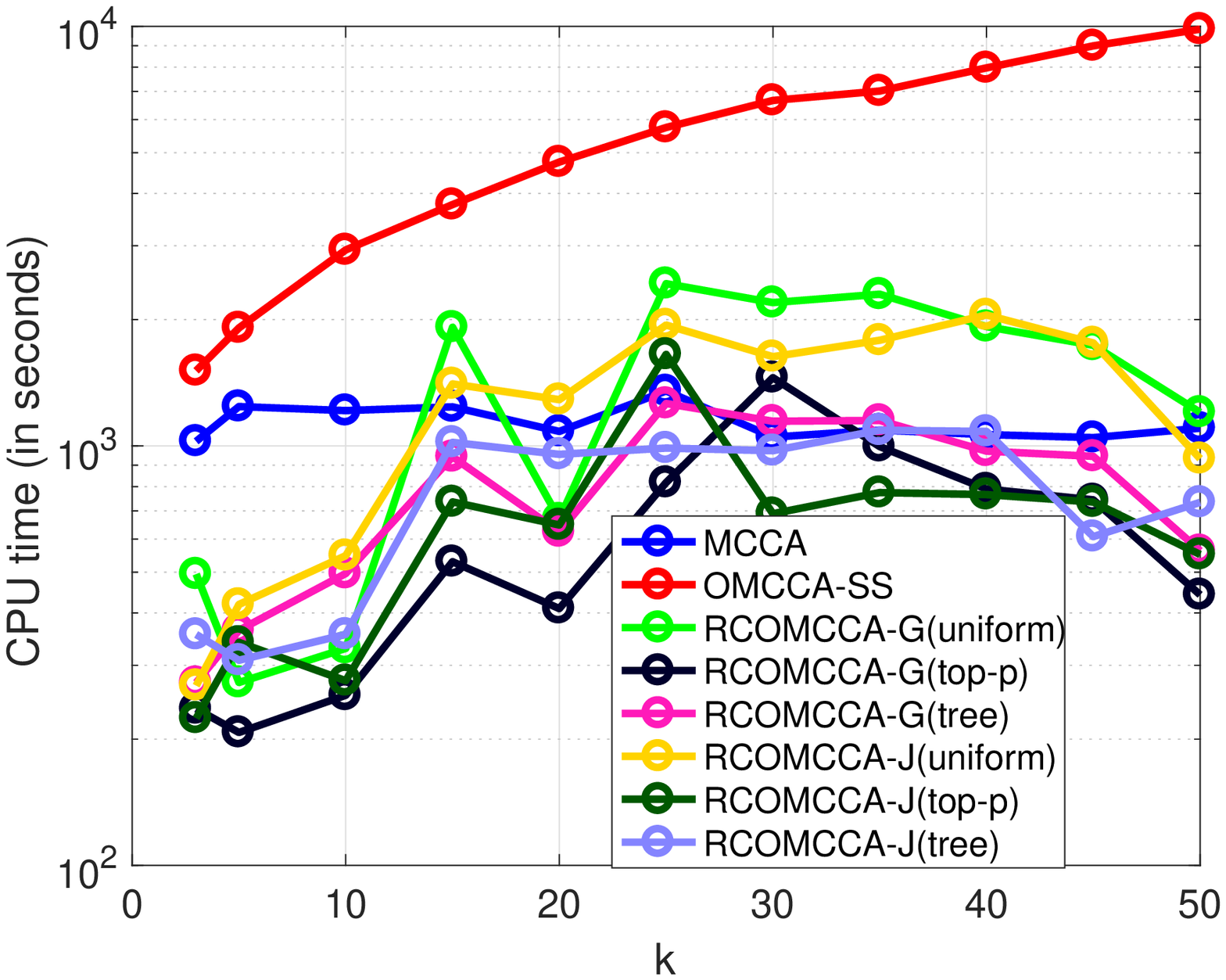}} &
		\raisebox{-.5\height}{\includegraphics[width=0.235\textwidth]{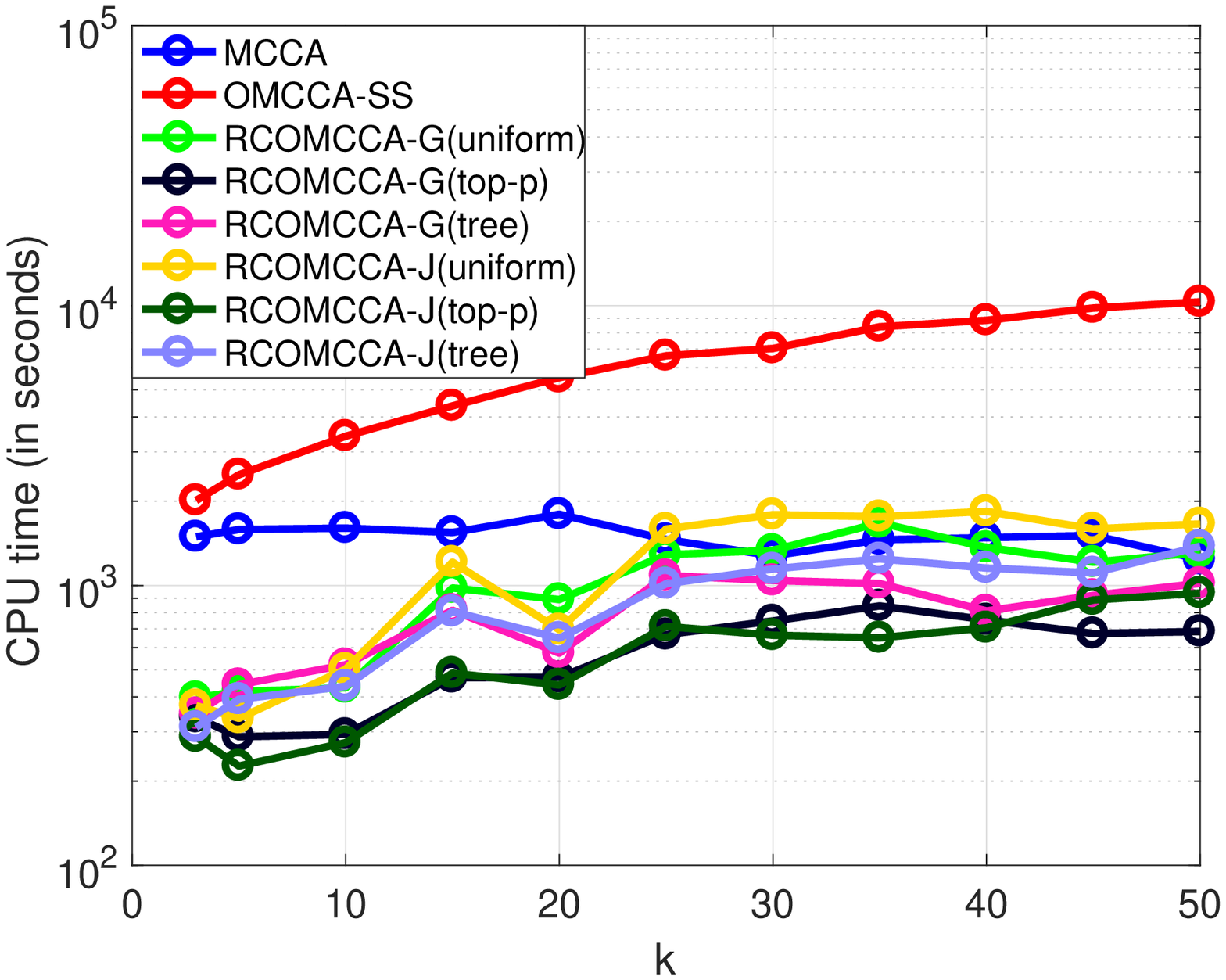}} &
		\raisebox{-.5\height}{\includegraphics[width=0.235\textwidth]{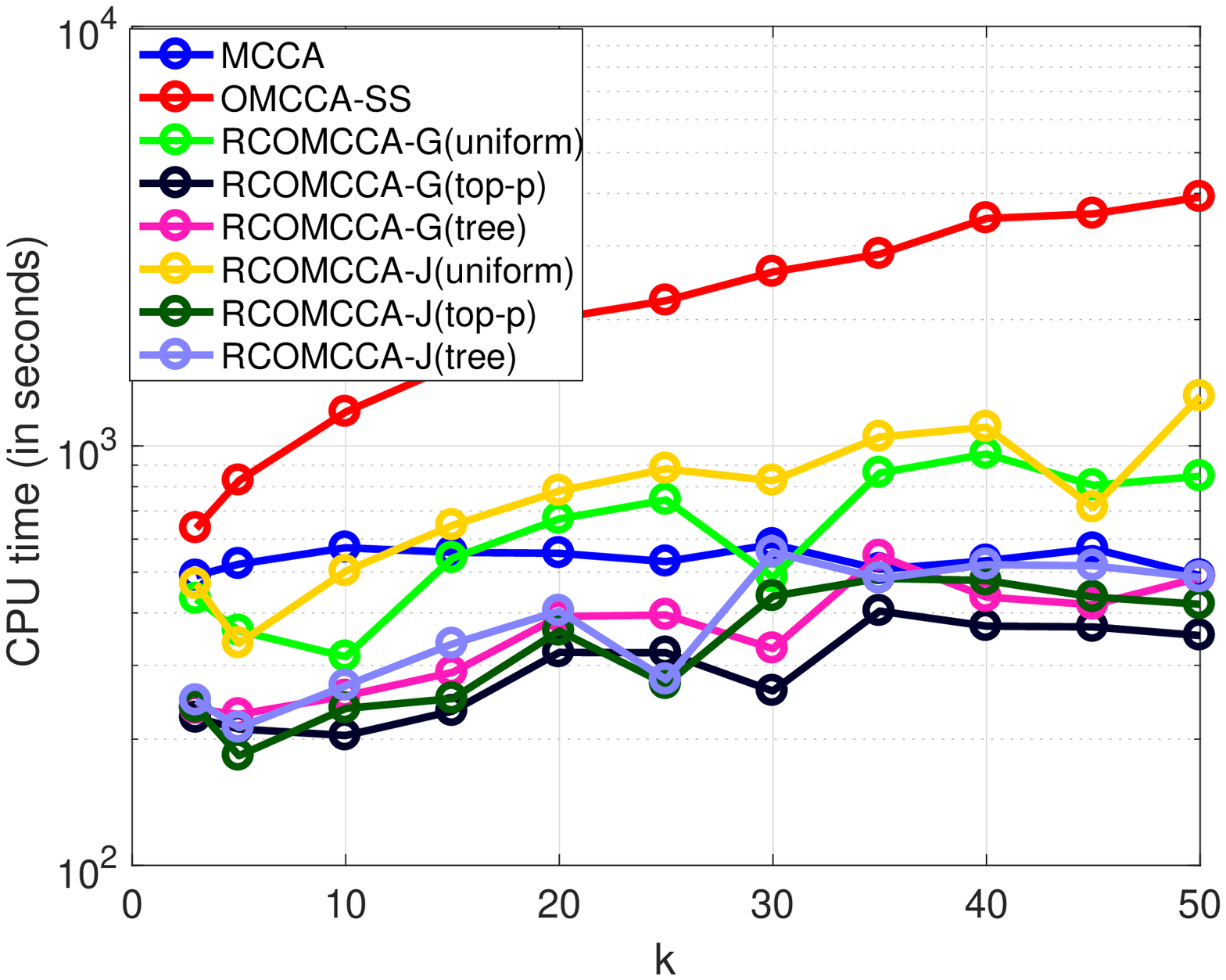}} &
		\raisebox{-.5\height}{\includegraphics[width=0.235\textwidth]{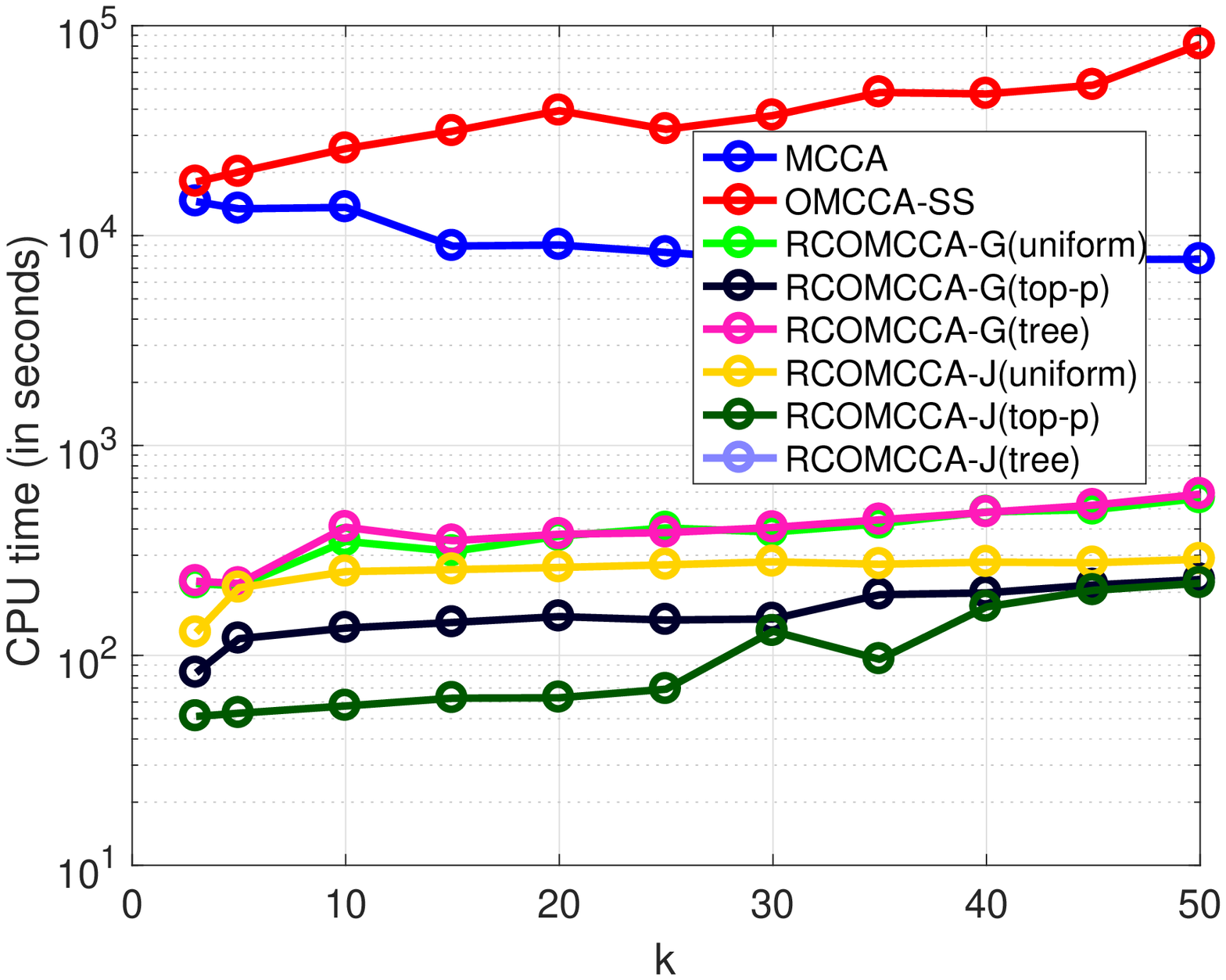}} \\
		\rotatebox[origin=c]{90}{Training ratio}&
		\raisebox{-.5\height}{\includegraphics[width=0.235\textwidth]{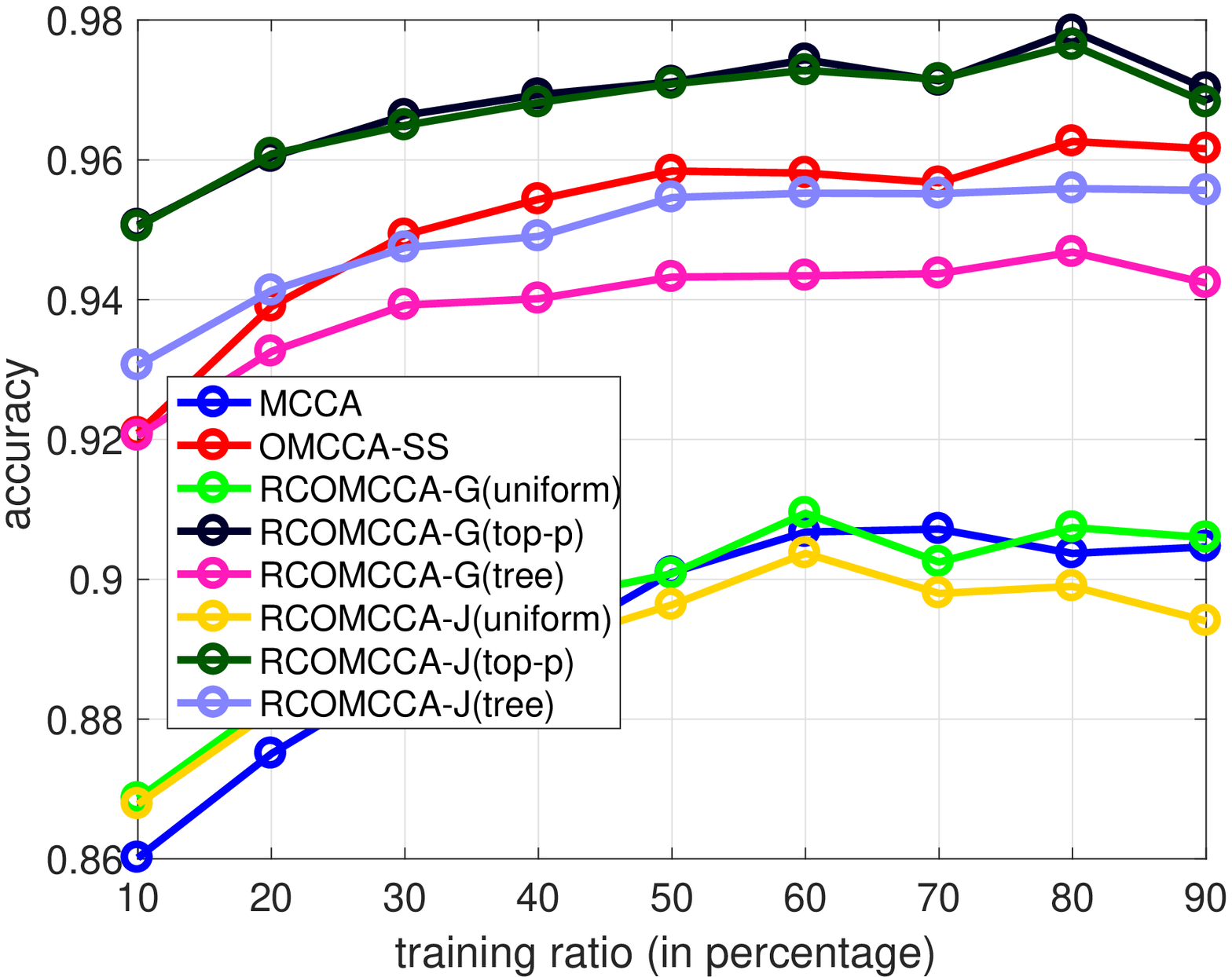}} &
		\raisebox{-.5\height}{\includegraphics[width=0.235\textwidth]{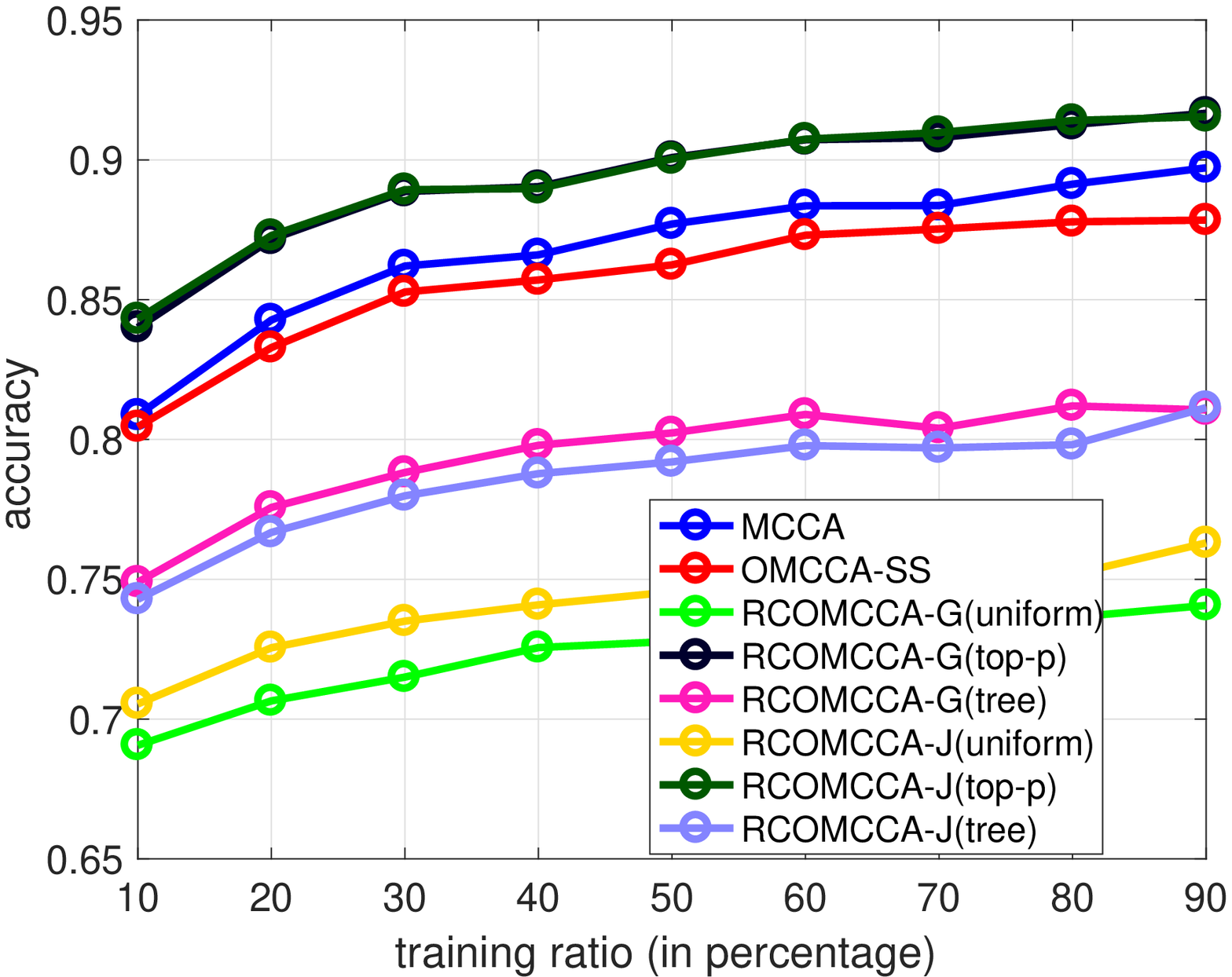}} &
		\raisebox{-.5\height}{\includegraphics[width=0.235\textwidth]{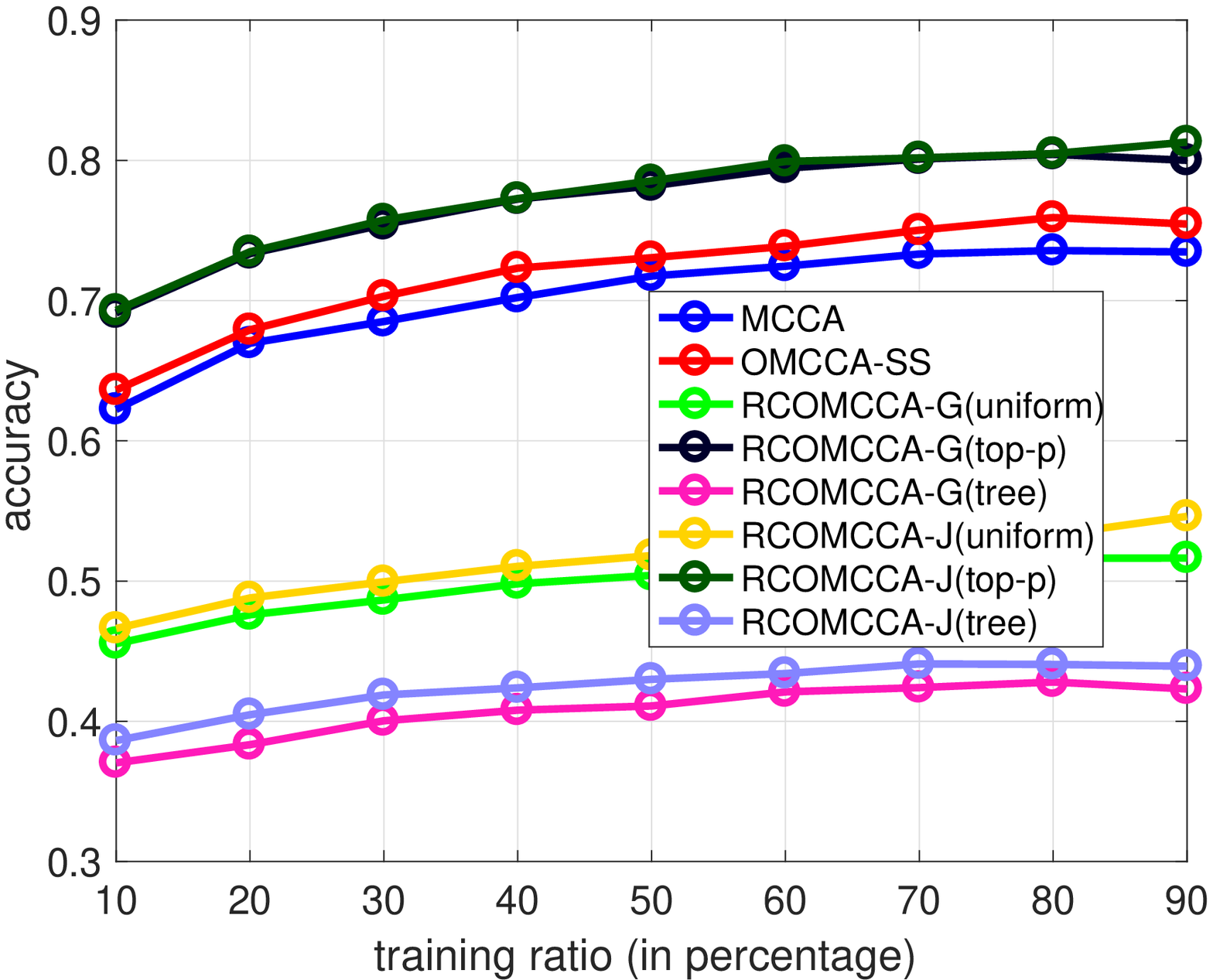}} &
		\raisebox{-.5\height}{\includegraphics[width=0.235\textwidth]{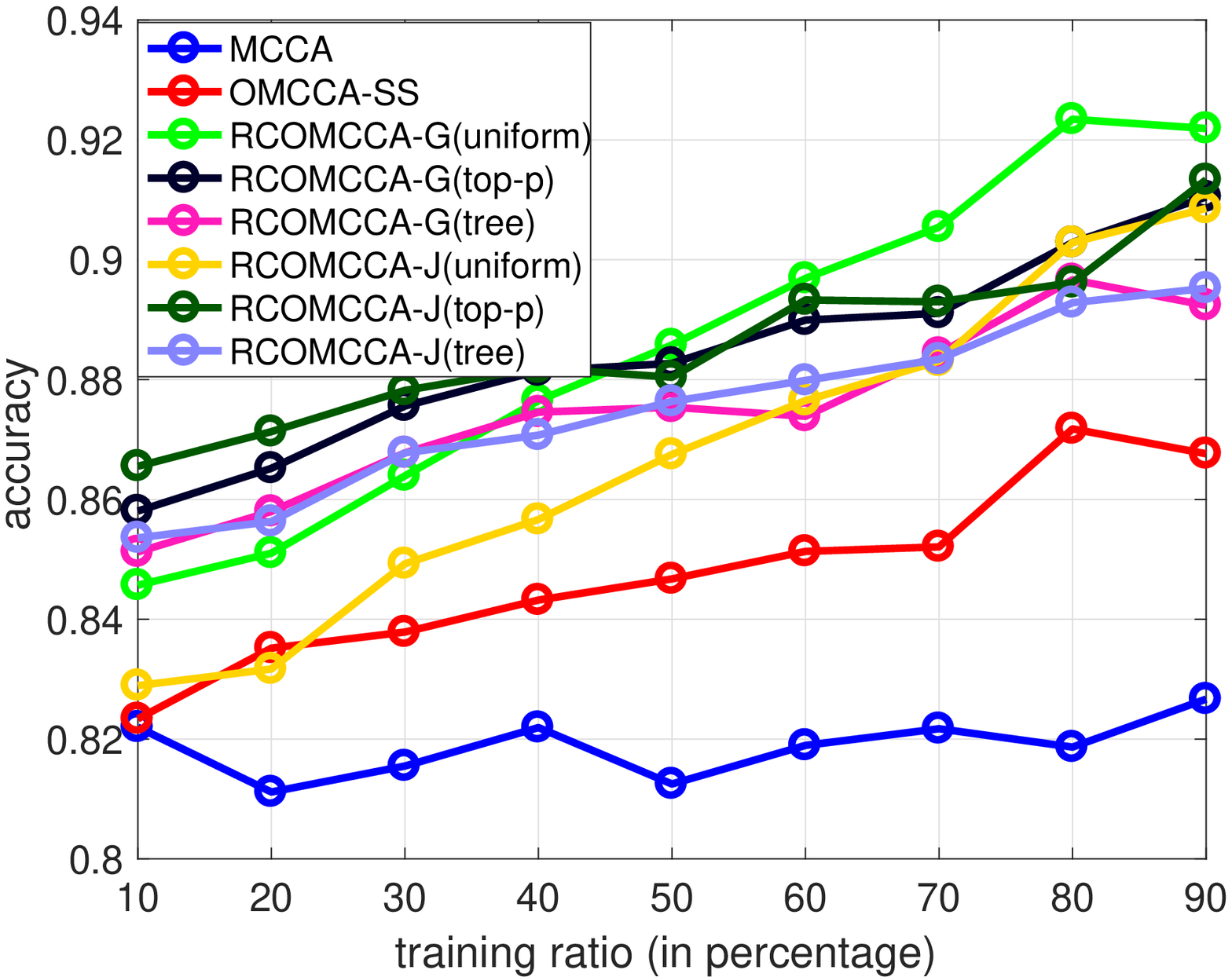}} \\
	\end{tabular}
	\caption{Accuracy and CPU time of three MCCA methods on four datasets by varying the reduced dimension $k$ and the training ratio.} \label{fig:mv_viz}
\end{figure*}

We  use 1-nearest neighbor classifier as the base classifier for evaluating the performance of multi-view feature extraction.
We run CCA methods to generate embeddings by varying $k \in \{3, 4, 5, 6\}$ for mfeat, and $k \in \{3, 5, 10, 15, 20, 25, 30, 35, 40, 45, 50 \}$ for other datasets. We split the data into training and testing with the ratio 30/70. Classification accuracy is used as the performance evaluation criterion. The experimental results are reported based on the average of $10$ randomly drawn splits.

We first compare eight variants of CCA methods and the classifiers based on each single view using all  input features.
Table \ref{fig:mv_viz} shows the results over five multi-view datasets with the best $k$ shown in bracket
for each method. From Table \ref{fig:mv_viz}, we have the following observations:
\begin{enumerate}
	\item CCA-based methods can achieve competitive or better results using a small set of features comparing with the best single view of the input features.
	
	\item OCCA methods including RCOMCCA (top-$p$) and OMCCA-SS generally show  better results than classical MCCA.
	This implies that orthogonality constraints added to MCCA can improve the learning performance.
	
	\item Our proposed RCOMCCA methods with top-$p$ weighting scheme demonstrates much better results
	than MCCA and OMCCA-SS can with large margins. Except for yeast\_ribosomal, RCOMCCA-G (top-$p$) and RCOMCCA-J (top-$p$)
	outperform the classifier of the best single view on the other four datasets.
	
	\item RCOMCCA with top-$p$ weighting scheme outperforms RCOMCCA with other two scheme. This implies that pairs of views could contribute differently to the downstreaming classification problem.
	
	\item For the same weighting schemes, our proposed RCOMCCA methods with Gauss-Seidel-style and Jacobi-style show almost similar results. It is recommended to take the problem structure into account for selecting the proper solver for efficiency as discussed in Section \ref{sec:MOCCA}.
\end{enumerate}
We also compare eight variants of CCA methods in terms of three other measurements including the sensitivity of parameter $k$,
the CPU time, and the sampling ratio of training and testing data. The results are shown in Fig. \ref{fig:mv_viz}. It is clear to see that
\begin{enumerate}
	\item Accuracies of all CCA methods are increasing with $k$. However, MCCA on Caltech101-7 and OMCCA-SS on yeast\_ribosomal
	behaves abnormally since the performance degrades significantly after a small $k$.
	
	\item RCOMCCA generally is the most efficient method than OMCCA-SS and MCCA. Due to the incremental optimization scheme,
	OMCCA-SS takes linear computational complexity with $k$, and so its CPU time increases with $k$. MCCA becomes less efficient
	if the total number of features in all views are large, for example, on yeast\_ribosomal because the analytic solution  is
	dependent of the generalized eigenvalue problem. As shown in Fig. \ref{fig:mv_viz}, on yeast\_ribosomal, MCCA also takes more than 10 times longer than RCOMCCA.
	
	\item All methods demonstrate  better performance when the number of training data increases.
	One notable exception is MCCA on yeast\_ribosomal, which does not show much gain as 
	training data ratio increases significantly.
	All orthogonally constrained CCAs do not show this issue.
\end{enumerate}
These observations imply that our proposed RCOMCCA not only can achieve noticeably better performance
but also much faster than OMCCA-SS and MCCA for multi-view feature extraction.

\section{Conclusion}\label{sec:conclusion}
In this paper, we start by proposing an efficient algorithm for solving CCA with orthogonality constraints, often called orthogonal CCA.
Then to model the data with more than two views, we presented a novel weighted multiset CCA again with orthogonality constraints.
Both algorithms rely on the solution of a subproblem with trace-fractional structure, which is solved by a newly proposed SCF iteration.
Theoretically, we performed a global and  local convergence analysis.
Extensive experiments are conducted for evaluating the proposed algorithms against existing methods in terms of various measurements,
including sensitivity analysis, correlation analysis, computation analysis, and data visualization.
We further apply our methods to two real world applications: multi-label classification and multi-view feature extraction.
Experimental results show that our methods not only perform competitively to or better than baselines in terms of accuracy
but also are more efficient. This work focuses on the linear orthogonal projection. In the future, we would like to
explore similar ideas for nonlinear CCA and other variants of CCAs with orthogonality constraints.

\bibliographystyle{plain}

\begin{thebibliography}{10}

\bibitem{abbg:2007}
P.-A. Absil, C.~G. Baker, and K.~A. Gallivan.
\newblock Trust-region methods on {R}iemannian manifolds.
\newblock {\em Found. Comput. Math.}, 7(3):303--330, 2007.

\bibitem{abms:2008}
P.-A. Absil, R.~Mahony, and R.~Sepulchre.
\newblock {\em Optimization Algorithms On Matrix Manifolds}.
\newblock Princeton University Press, 2008.

\bibitem{akaho2006kernel}
Shotaro Akaho.
\newblock A kernel method for canonical correlation analysis.
\newblock {\em International Meeting on Psychometric Society}, 2001.

\bibitem{abbd:1999}
E.~Anderson, Z.~Bai, C.~Bischof, J.~W. Demmel, J.~Dongarra, J.~Du Croz,
  A.~Greenbaum, S.~Hammarling, A.~McKenney, S.~Ostrouchov, and D.~Sorensen.
\newblock {\em {LAPACK} Users' Guide}.
\newblock SIAM, Philadelphia, 3rd edition, 1999.

\bibitem{andrew2013deep}
Galen Andrew, Raman Arora, Jeff Bilmes, and Karen Livescu.
\newblock Deep canonical correlation analysis.
\newblock In {\em International conference on machine learning}, pages
  1247--1255, 2013.

\bibitem{balv:2018}
Z.~Bai, D.~Lu, and B.~Vandereycken.
\newblock Robust rayleigh quotient minimization and nonlinear eigenvalue
  problems.
\newblock {\em SIAM J. Sci. Comput.}, 40:A3495--A3522, 2018.

\bibitem{bddrv:2000}
Zhaojun Bai, J.~Demmel, J.~Dongarra, A.~Ruhe, and H.~van~der Vorst~{\rm
  (editors)}.
\newblock {\em Templates for the solution of Algebraic Eigenvalue Problems: A
  Practical Guide}.
\newblock SIAM, Philadelphia, 2000.

\bibitem{boutell2004learning}
Matthew~R Boutell, Jiebo Luo, Xipeng Shen, and Christopher~M Brown.
\newblock Learning multi-label scene classification.
\newblock {\em Pattern recognition}, 37(9):1757--1771, 2004.

\bibitem{cai2005orthogonal}
Deng Cai and Xiaofei He.
\newblock Orthogonal locality preserving indexing.
\newblock In {\em Proceedings of the 28th annual international ACM SIGIR
  conference on Research and development in information retrieval}, pages
  3--10. ACM, 2005.

\bibitem{cazb:2018}
Y.~Cai, L.-H. Zhang, Z.~Bai, and R.-C. Li.
\newblock On an eigenvector-dependent nonlinear eigenvalue problem.
\newblock {\em SIAM J. Matrix Anal. Appl.}, 39(3):1360--1382, 2018.

\bibitem{chlg:2013a}
D.~Chu, L.~Liao, M.~K. Ng, and X.~Zhang.
\newblock Sparse kernel canonical correlation analysis.
\newblock In {\em Proceedings of the International MultiConference of Engineers
  and Computer Scientists 2013}, volume~I of {\em IMECS 2013}, Hong Kong, March
  2013.

\bibitem{cugh:2015}
J.~P. Cunningham and Z.~Ghahramani.
\newblock Linear dimensionality reduction: Survey, insights, and
  generalizations.
\newblock {\em J. Mach. Learning Res.}, 16:2859--2900, 2015.

\bibitem{demm:1997}
J.~Demmel.
\newblock {\em Applied Numerical Linear Algebra}.
\newblock SIAM, Philadelphia, PA, 1997.

\bibitem{dhillon2011multi}
Paramveer Dhillon, Dean~P Foster, and Lyle~H Ungar.
\newblock Multi-view learning of word embeddings via cca.
\newblock In {\em Advances in neural information processing systems}, pages
  199--207, 2011.

\bibitem{Dua:2019}
Dheeru Dua and Casey Graff.
\newblock {UCI} machine learning repository, 2017.

\bibitem{govl:2013}
G.~H. Golub and C.~F. {Van Loan}.
\newblock {\em Matrix Computations}.
\newblock Johns Hopkins University Press, Baltimore, Maryland, 4th edition,
  2013.

\bibitem{hardoon2007unsupervised}
David~R Hardoon, Janaina Mourao-Miranda, Michael Brammer, and John
  Shawe-Taylor.
\newblock Unsupervised analysis of fmri data using kernel canonical
  correlation.
\newblock {\em NeuroImage}, 37(4):1250--1259, 2007.

\bibitem{hardoon2004canonical}
David~R Hardoon, Sandor Szedmak, and John Shawe-Taylor.
\newblock Canonical correlation analysis: An overview with application to
  learning methods.
\newblock {\em Neural computation}, 16(12):2639--2664, 2004.

\bibitem{hojo:1991}
R.~A. Horn and C.~R. Johnson.
\newblock {\em Topics in Matrix Analysis}.
\newblock Cambridge University Press, Cambridge, 1991.

\bibitem{hote:1936}
H.~Hotelling.
\newblock Relations between two sets of variates.
\newblock {\em Biometrika}, 28(3-4):321--377, 1936.

\bibitem{kakade2007multi}
Sham~M Kakade and Dean~P Foster.
\newblock Multi-view regression via canonical correlation analysis.
\newblock In {\em International Conference on Computational Learning Theory},
  pages 82--96. Springer, 2007.

\bibitem{kettenring1971canonical}
Jon~R Kettenring.
\newblock Canonical analysis of several sets of variables.
\newblock {\em Biometrika}, 58(3):433--451, 1971.

\bibitem{klami2013bayesian}
Arto Klami, Seppo Virtanen, and Samuel Kaski.
\newblock Bayesian canonical correlation analysis.
\newblock {\em Journal of Machine Learning Research}, 14(Apr):965--1003, 2013.

\bibitem{knya:2001}
A.~V. Knyazev.
\newblock Toward the optimal preconditioned eigensolver: Locally optimal block
  preconditioned conjugate gradient method.
\newblock {\em SIAM J. Sci. Comput.}, 23(2):517--541, 2001.

\bibitem{knne:2003}
A.~V. Knyazev and K.~Neymeyr.
\newblock Efficient solution of symmetric eigenvalue problems using multigrid
  preconditioners in the locally optimal block conjugate gradient method.
\newblock {\em Electron. Trans. Numer. Anal.}, 15:38--55, 2003.

\bibitem{kokiopoulou2007orthogonal}
Effrosyni Kokiopoulou and Yousef Saad.
\newblock Orthogonal neighborhood preserving projections: A projection-based
  dimensionality reduction technique.
\newblock {\em IEEE Transactions on Pattern Analysis and Machine Intelligence},
  29(12):2143--2156, 2007.

\bibitem{lazebnik2006beyond}
Svetlana Lazebnik, Cordelia Schmid, and Jean Ponce.
\newblock Beyond bags of features: Spatial pyramid matching for recognizing
  natural scene categories.
\newblock In {\em 2006 IEEE Computer Society Conference on Computer Vision and
  Pattern Recognition (CVPR'06)}, volume~2, pages 2169--2178. IEEE, 2006.

\bibitem{fei2007learning}
Fei-Fei Li, Rob Fergus, and Pietro Perona.
\newblock Learning generative visual models from few training examples: An
  incremental bayesian approach tested on 101 object categories.
\newblock {\em Computer vision and Image understanding}, 106(1):59--70, 2007.

\bibitem{lizh:2019}
L.~Li and Z.~Zhang.
\newblock Semi-supervised domain adaptation by covariance matching.
\newblock {\em IEEE Trans. Pattern Anal. Mach. Intell.}, 2019.
\newblock to appear.

\bibitem{mart:2004}
R.~M. Martin.
\newblock {\em Electronic Structure: Basic Theory and Practical Methods}.
\newblock Cambridge University Press, Cambridge, UK, 2004.

\bibitem{nielsen2002multiset}
Allan~Aasbjerg Nielsen.
\newblock Multiset canonical correlations analysis and multispectral, truly
  multitemporal remote sensing data.
\newblock {\em IEEE transactions on image processing}, 11(3):293--305, 2002.

\bibitem{nowr:2006}
Jorge Nocedal and Stephen Wright.
\newblock {\em Numerical Optimization}.
\newblock Springer, 2nd edition, 2006.

\bibitem{ojala2002multiresolution}
Timo Ojala, Matti Pietik{\"a}inen, and Topi M{\"a}enp{\"a}{\"a}.
\newblock Multiresolution gray-scale and rotation invariant texture
  classification with local binary patterns.
\newblock {\em IEEE Transactions on Pattern Analysis \& Machine Intelligence},
  (7):971--987, 2002.

\bibitem{oliva2001modeling}
Aude Oliva and Antonio Torralba.
\newblock Modeling the shape of the scene: A holistic representation of the
  spatial envelope.
\newblock {\em International journal of computer vision}, 42(3):145--175, 2001.

\bibitem{rai2009multi}
Piyush Rai and Hal Daume.
\newblock Multi-label prediction via sparse infinite cca.
\newblock In {\em Advances in Neural Information Processing Systems}, pages
  1518--1526, 2009.

\bibitem{sacs:2010}
Y.~Saad, J.~R. Chelikowsky, and S.~M. Shontz.
\newblock Numerical methods for electronic structure calculations of materials.
\newblock {\em SIAM Rev.}, 52(1):3--54, 2010.

\bibitem{sargin2007audiovisual}
Mehmet~Emre Sargin, Y{\"u}cel Yemez, Engin Erzin, and A~Murat Tekalp.
\newblock Audiovisual synchronization and fusion using canonical correlation
  analysis.
\newblock {\em IEEE Transactions on Multimedia}, 9(7):1396--1403, 2007.

\bibitem{shen2013orthogonal}
Xiao-Bo Shen, Quan-Sen Sun, and Yun-Hao Yuan.
\newblock Orthogonal canonical correlation analysis and its application in
  feature fusion.
\newblock In {\em Proceedings of the 16th International Conference on
  Information Fusion}, pages 151--157. IEEE, 2013.

\bibitem{shen2015orthogonal}
Xiaobo Shen and Quansen Sun.
\newblock Orthogonal multiset canonical correlation analysis based on
  fractional-order and its application in multiple feature extraction and
  recognition.
\newblock {\em Neural Processing Letters}, 42(2):301--316, 2015.

\bibitem{sun:1987}
Ji-Guang Sun.
\newblock {\em Matrix Perturbation Analysis}.
\newblock Academic Press, Beijing, 1987.
\newblock In Chinese.

\bibitem{sun2010canonical}
Liang Sun, Shuiwang Ji, and Jieping Ye.
\newblock Canonical correlation analysis for multilabel classification: A
  least-squares formulation, extensions, and analysis.
\newblock {\em IEEE Transactions on Pattern Analysis and Machine Intelligence},
  33(1):194--200, 2010.

\bibitem{tsoumakas2007multi}
Grigorios Tsoumakas and Ioannis Katakis.
\newblock Multi-label classification: An overview.
\newblock {\em International Journal of Data Warehousing and Mining (IJDWM)},
  3(3):1--13, 2007.

\bibitem{uurtio2018tutorial}
Viivi Uurtio, Jo{\~a}o~M Monteiro, Jaz Kandola, John Shawe-Taylor, Delmiro
  Fernandez-Reyes, and Juho Rousu.
\newblock A tutorial on canonical correlation methods.
\newblock {\em ACM Computing Surveys (CSUR)}, 50(6):95, 2018.

\bibitem{wara:2015}
Z.~Wang, Q.~Ruan, and G.~An.
\newblock Projection-optimal local {F}isher discriminant analysis for feature
  extraction.
\newblock {\em Neural Comput {\&} Applic}, 26:589--601, 2015.

\bibitem{witten2009penalized}
Daniela~M Witten, Robert Tibshirani, and Trevor Hastie.
\newblock A penalized matrix decomposition, with applications to sparse
  principal components and canonical correlation analysis.
\newblock {\em Biostatistics}, 10(3):515--534, 2009.

\bibitem{wu2008place}
Jianixn Wu and James~M Rehg.
\newblock Where am i: Place instance and category recognition using spatial
  pact.
\newblock In {\em 2008 Ieee Conference on Computer Vision and Pattern
  Recognition}, pages 1--8. IEEE, 2008.

\bibitem{yazs:2014}
W.~H. Yang, L.-H. Zhang, and R.~Y. Song.
\newblock Optimality conditions of the nonlinear programming on {Riemannian}
  manifolds.
\newblock {\em Pacific J. Optim.}, 10:415--434, 2014.

\bibitem{ye2005characterization}
Jieping Ye.
\newblock Characterization of a family of algorithms for generalized
  discriminant analysis on undersampled problems.
\newblock {\em Journal of Machine Learning Research}, 6(Apr):483--502, 2005.

\bibitem{zhang2014semi}
Bo~Zhang, Jie Hao, Gang Ma, Jinpeng Yue, and Zhongzhi Shi.
\newblock Semi-paired probabilistic canonical correlation analysis.
\newblock In {\em International Conference on Intelligent Information
  Processing}, pages 1--10. Springer, 2014.

\bibitem{zhan:2011}
L.-H. Zhang.
\newblock Uncorrelated trace ratio {LDA} for undersampled problems.
\newblock {\em Patt. Recog. Lett.}, 32:476--484, 2011.

\bibitem{zhln:2010}
L.-H. Zhang, L.-Z. Liao, and M.~K. Ng.
\newblock Fast algorithms for the generalized {Foley-Sammon} discriminant
  analysis.
\newblock {\em SIAM J. Matrix Anal. Appl.}, 31(4):1584--1605, 2010.

\bibitem{zhli:2014a}
Lei-Hong Zhang and Ren-Cang Li.
\newblock Maximization of the sum of the trace ratio on the {Stiefel} manifold,
  {I}: Theory.
\newblock {\em SCIENCE CHINA Math.}, 57(12):2495--2508, 2014.

\bibitem{zhli:2014b}
Lei-Hong Zhang and Ren-Cang Li.
\newblock Maximization of the sum of the trace ratio on the {Stiefel} manifold,
  {II}: Computation.
\newblock {\em SCIENCE CHINA Math.}, 58(7):1549--1566, 2015.

\bibitem{zhang2007ml}
Min-Ling Zhang and Zhi-Hua Zhou.
\newblock Ml-knn: A lazy learning approach to multi-label learning.
\newblock {\em Pattern recognition}, 40(7):2038--2048, 2007.

\bibitem{zhang2011multi}
Yi~Zhang and Jeff Schneider.
\newblock Multi-label output codes using canonical correlation analysis.
\newblock In {\em Proceedings of the fourteenth international conference on
  artificial intelligence and statistics}, pages 873--882, 2011.

\end{thebibliography}

\end{document}